\documentclass{article}
\usepackage[T1]{fontenc}
\usepackage{microtype}
\usepackage{caption}
\usepackage{amsmath}
\usepackage{amssymb}
\usepackage{amsthm}
\usepackage{mathtools}
\usepackage{thmtools}
\usepackage{mathrsfs}
\usepackage{libertine}
\usepackage[varl,varqu]{zi4}
\usepackage{array}
\usepackage{booktabs}
\usepackage[inline,shortlabels]{enumitem}
\usepackage{graphicx}
\usepackage[dvipsnames]{xcolor}
\usepackage{hyperref}
\usepackage{cleveref}
\definecolor{alizarin}{RGB}{227,38,54}
\definecolor{ultramarine}{RGB}{18,10,143}
\definecolor{Amaranth}{rgb}{0.9, 0.17, 0.31}
\definecolor{tumgreen}{RGB}{162,173,0}
\definecolor{tumblue}{RGB}{0, 101, 189}

\hypersetup{
  linkcolor  = tumblue,
  citecolor  = ultramarine,
  urlcolor   = ultramarine,
  colorlinks = true
}
\usepackage[margin=1in]{geometry}
\usepackage[outline]{contour}
\contourlength{1.2pt}

\def\mA{{\bm{A}}}
\def\mB{{\bm{B}}}
\def\mC{{\bm{C}}}
\def\mD{{\bm{D}}}
\def\mE{{\bm{E}}}
\def\mF{{\bm{F}}}
\def\mG{{\bm{G}}}
\def\mH{{\bm{H}}}
\def\mI{{\bm{I}}}
\def\mJ{{\bm{J}}}
\def\mK{{\bm{K}}}
\def\mL{{\bm{L}}}
\def\mM{{\bm{M}}}
\def\mN{{\bm{N}}}

\def\mP{{\bm{P}}}
\def\mQ{{\bm{Q}}}
\def\mR{{\bm{R}}}
\def\mS{{\bm{S}}}
\def\mT{{\bm{T}}}
\def\mU{{\bm{U}}}
\def\mV{{\bm{V}}}
\def\mW{{\bm{W}}}
\def\mX{{\bm{X}}}
\def\mY{{\bm{Y}}}
\def\mZ{{\bm{Z}}}

\def\va{{\bm{a}}}
\def\vb{{\bm{b}}}

\def\vd{{\bm{d}}}
\def\ve{{\bm{e}}}

\def\vg{{\bm{g}}}
\def\vh{{\bm{h}}}

\def\vm{{\bm{m}}}

\def\vp{{\bm{p}}}

\def\vr{{\bm{r}}}
\def\vs{{\bm{s}}}

\def\vu{{\bm{u}}}
\def\vv{{\bm{v}}}
\def\vw{{\bm{w}}}
\def\vx{{\bm{x}}}
\def\vy{{\bm{y}}}
\def\vz{{\bm{z}}}

\usepackage{amsmath,amsfonts,bm}
\usepackage{tcolorbox}

\def\eqref#1{equation~\ref{#1}}

\def\1{\bm{1}}

\def\va{{\bm{a}}}
\def\vb{{\bm{b}}}

\def\vd{{\bm{d}}}
\def\ve{{\bm{e}}}

\def\vg{{\bm{g}}}
\def\vh{{\bm{h}}}

\def\vm{{\bm{m}}}

\def\vp{{\bm{p}}}

\def\vr{{\bm{r}}}
\def\vs{{\bm{s}}}

\def\vu{{\bm{u}}}
\def\vv{{\bm{v}}}
\def\vw{{\bm{w}}}
\def\vx{{\bm{x}}}
\def\vy{{\bm{y}}}
\def\vz{{\bm{z}}}

\def\mA{{\bm{A}}}
\def\mB{{\bm{B}}}
\def\mC{{\bm{C}}}
\def\mD{{\bm{D}}}
\def\mE{{\bm{E}}}
\def\mF{{\bm{F}}}
\def\mG{{\bm{G}}}
\def\mH{{\bm{H}}}
\def\mI{{\bm{I}}}
\def\mJ{{\bm{J}}}
\def\mK{{\bm{K}}}
\def\mL{{\bm{L}}}
\def\mM{{\bm{M}}}
\def\mN{{\bm{N}}}

\def\mP{{\bm{P}}}
\def\mQ{{\bm{Q}}}
\def\mR{{\bm{R}}}
\def\mS{{\bm{S}}}
\def\mT{{\bm{T}}}
\def\mU{{\bm{U}}}
\def\mV{{\bm{V}}}
\def\mW{{\bm{W}}}
\def\mX{{\bm{X}}}
\def\mY{{\bm{Y}}}
\def\mZ{{\bm{Z}}}

\DeclareMathAlphabet{\mathsfit}{\encodingdefault}{\sfdefault}{m}{sl}
\SetMathAlphabet{\mathsfit}{bold}{\encodingdefault}{\sfdefault}{bx}{n}

\newcommand{\E}{\mathbb{E}}

\newcommand{\R}{\mathbb{R}}

\definecolor{first}{HTML}{2CA02C}  
\definecolor{second}{HTML}{FF7F0E} 
\definecolor{third}{HTML}{FFD700}  

\definecolor{alizarin}{RGB}{227,38,54}
\definecolor{ultramarine}{RGB}{18,10,143}
\definecolor{Amaranth}{rgb}{0.9, 0.17, 0.31}
\definecolor{tumgreen}{RGB}{162,173,0}
\definecolor{tumblue}{RGB}{0, 101, 189}

\hypersetup{
  linkcolor  = tumblue,
  citecolor  = ultramarine,
  urlcolor   = ultramarine,
  colorlinks = true
}

\newcommand{\norm}[1]{\left\|#1\right\|}

\usepackage{microtype}
\usepackage{titletoc}
\usepackage{subfigure}
\usepackage{natbib}
\usepackage{hyperref}
\usepackage{algorithm}
\usepackage{algorithmic}
\usepackage{hyperref}
\usepackage{url}
\usepackage{array,multirow,graphicx}
\usepackage{comment}
\usepackage{booktabs}
\usepackage{xcolor}
\usepackage{enumitem}
\usepackage{color}
\usepackage{amsmath,amsfonts,bm}
\usepackage{diagbox}
\usepackage{relsize}
\usepackage{wrapfig,lipsum,booktabs}
\usepackage{bbm}

\usepackage[table]{xcolor}
\usepackage[dvipsnames]{xcolor}

\newtheorem{theorem}{Theorem}

\newtheorem{proposition}[theorem]{Proposition}

\newtheorem{corollary}[theorem]{Corollary}

\makeatletter
\def\moverlay{\mathpalette\mov@rlay}
\def\mov@rlay#1#2{\leavevmode\vtop{%
   \baselineskip\z@skip \lineskiplimit-\maxdimen
   \ialign{\hfil$\m@th#1##$\hfil\cr#2\crcr}}}
\newcommand{\charfusion}[3][\mathord]{
    #1{\ifx#1\mathop\vphantom{#2}\fi
        \mathpalette\mov@rlay{#2\cr#3}
      }
    \ifx#1\mathop\expandafter\displaylimits\fi}
\makeatother

\usepackage[textsize=tiny]{todonotes}
\usepackage{parskip}
\usepackage{etoolbox}
\makeatother

\title{Learning to Compare Neural Representations   from \\ Parameterized Computation Graphs}
\author{
Ya-Wei Eileen Lin$^{1}$, Daniel Herbst$^{1}$, and Stefanie Jegelka$^{1,2}$
\\ 
\\
\normalsize{$^1$Technical University of Munich; School of CIT, MCML, MDSI} \\
\normalsize{
$^2$MIT; Department of EECS, CSAIL} 
}
\date{}

\begin{document}
\maketitle

\begin{abstract}
Comparing internal representations across trained models is important for identifying corresponding layers and selecting components for model reuse.   Yet, direct comparisons at inference require evaluating candidate models on known reference inputs and collecting their internal activations. In this paper, we study whether internal representation similarity measured across independently trained models can instead be learned to predict in unseen models from their \textit{parameterized computations}. We introduce a framework that applies a graph metanetwork for mapping each internal site to an embedding derived from its upstream computations while accounting for hidden-coordinate permutation and internal attention-basis transformations. We analyze the structural properties of our estimators and derive conditional guarantees for layer retrieval and stitching candidate selection. Empirically,  we find that the learned predictors support relation estimation and low-regret layer retrieval. They also select useful stitching connections while not being trained on stitching outcomes. This opens up a new way to study representation similarity directly from trained neural networks: representational relations learned across a collection of trained models can generalize to new computations,  and that the predicted relations can guide candidate selection before direct measurement or adapter fitting for functional evaluations.
\end{abstract}

\section{Introduction}

\begin{wrapfigure}[21]{r}{0.48\textwidth}  
\vspace{-4 mm}
     \includegraphics[width=0.48\textwidth]{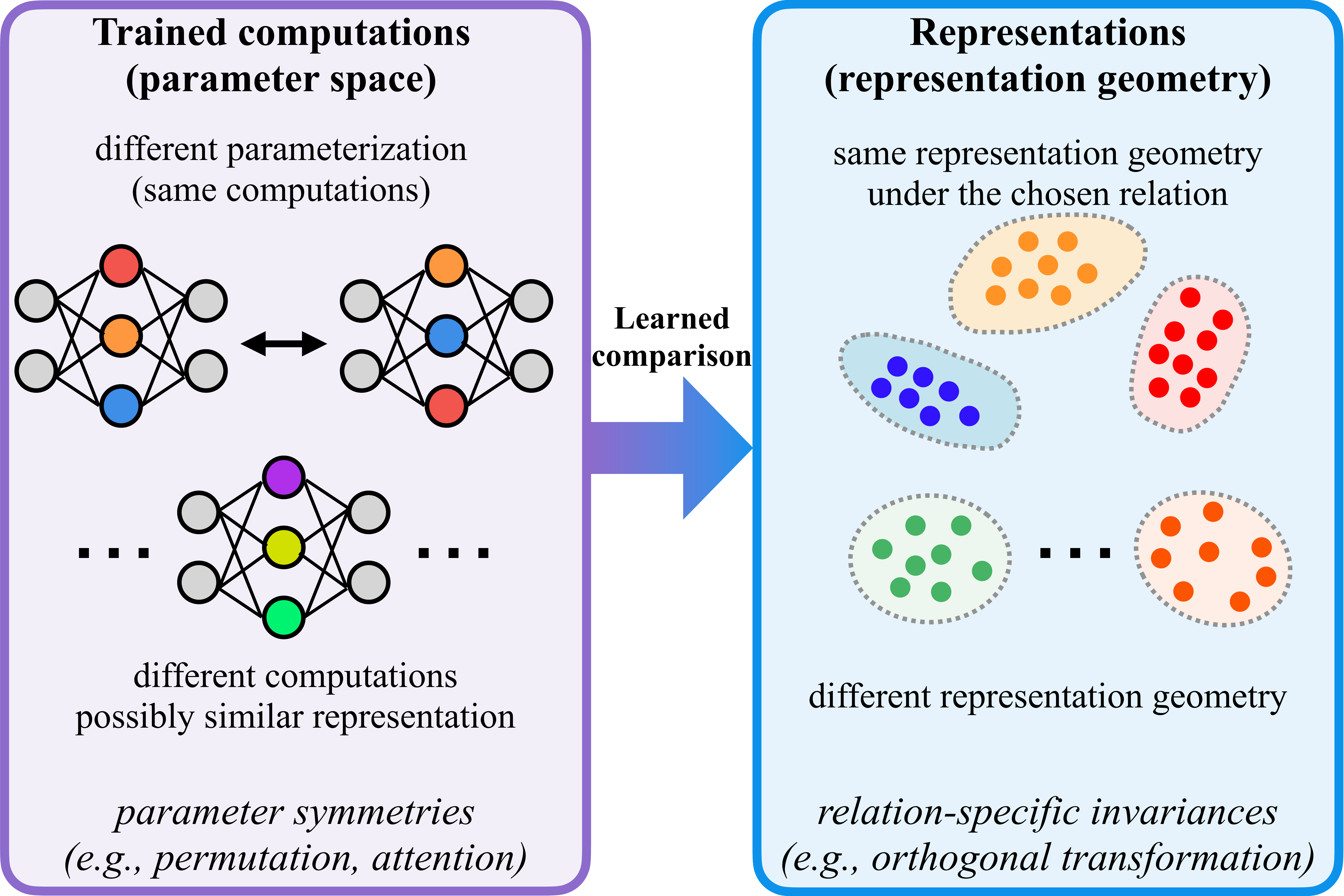}
     \vspace{-6 mm}
    \caption{
    Learning a connection between trained computations and representation geometry. 
    A representation site is determined by its upstream computation, but symmetry-related parameterizations can induce equivalent computations.
    Representational correspondence can also extend beyond these parameter symmetries: different computations may produce the same representation geometry under a chosen relation.
    }
    \label{fig:param_to_repre}
\end{wrapfigure} 
Finding corresponding representations across trained neural networks \citep{bau2018identifying, huh2024position, gurnee2024universal} is an important scientific task for understanding shared internal structure across independently trained models \citep{pmlr-v44-li15convergent, morcos2018insights, raghu2017svcca, kornblith2019similarity} and for identifying components that can be reused across networks \citep{yosinski2014transferable, bansal2021revisiting, csiszarik2021similarity, moschella2023relative}.
Both goals motivate searching within and across a collection of trained models: given an internal representation in one model, which sites provide useful counterparts for analysis or potential connection points for model composition \citep{maiorca2023latent,cannistraci2024from,pan2023stitchable, yang2022deep, hu2024model, wang2023towards}? 
These correspondences cannot be determined by layer indices or tensor shapes \citep{nguyen2021do, khosla2024soft, singh2020model}, since networks may differ in depth and width, and each representation depends on the trained computation that produces it \citep{yurochkin2019bayesian, tatro2020optimizing}.

\begin{figure*}[t]
    \centering
     \includegraphics[width=0.98\textwidth]{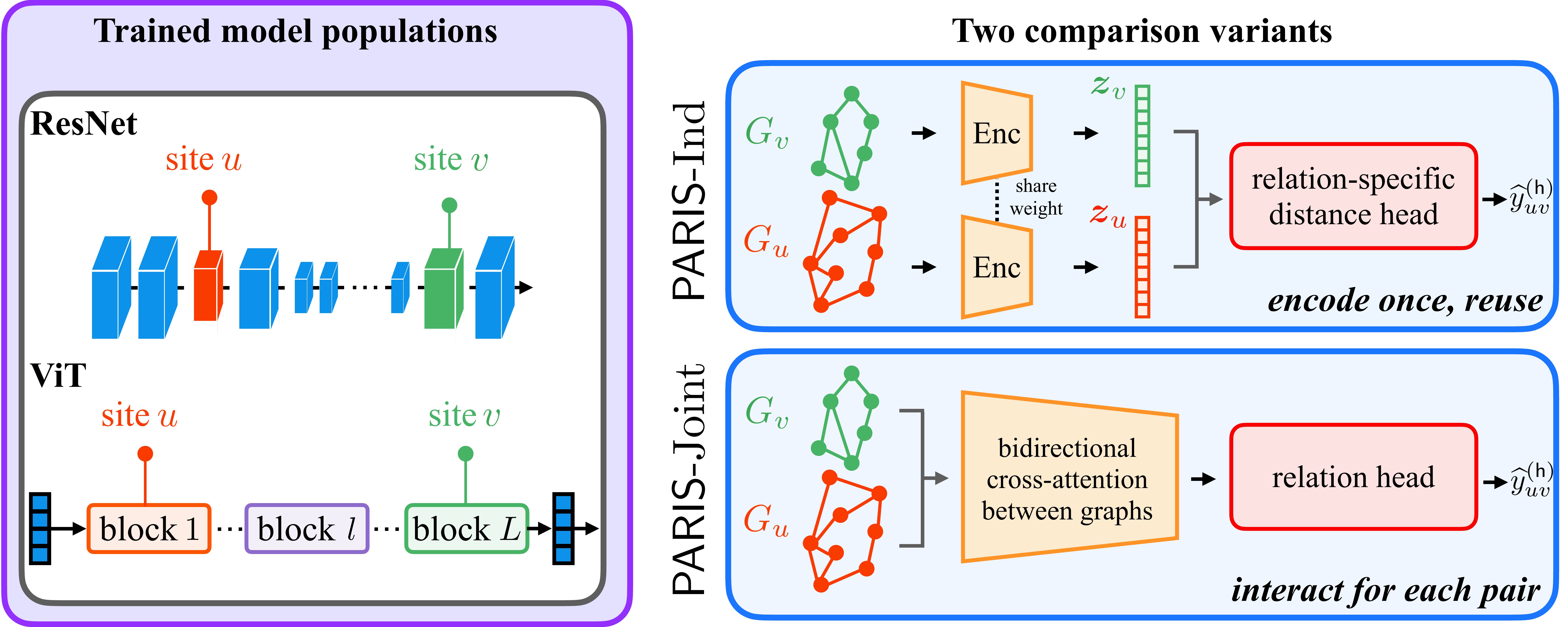}

\caption{
Overview of $\mathsf{PARIS}$.
We consider a population of trained models. Each site is described by a graph of its upstream computation. $\mathsf{PARIS}$-Ind encodes each site independently and compares reusable embeddings using relation-specific distance heads. $\mathsf{PARIS}$-Joint allows the two site graphs to interact before predicting their relation.
Both are trained with internal representation similarity and rank candidate sites in unseen models without collecting new base-model activations. Our metanetworks can guide the selection of sites and be used for adapter fitting.
}
\label{fig:illstration}
\end{figure*}

To compare internal representations at different sites across models, i.e., layers or other locations at which activations are measured, multiple criteria exist  \citep{klabunde2025resi, kriegeskorte2008representational, diedrichsen2017representational}.
These methods emphasize different aspects of representational similarity, including response geometry, neighborhood structure, and predictive behavior \citep{morcos2018insights, williams2021generalized, davari2023reliability, pmlr-v162-barannikov22a}. 
For instance, CKA \citep{kornblith2019similarity} compares response geometry through the alignment of centered kernel matrices, CKNNA \citep{huh2024position} focuses on local structure by restricting kernel alignment to shared nearest neighbors, and  readout-based methods compare representations through the disagreement of fitted predictors under a specified prediction protocol
\citep{alain2016understanding, harvey2024what}. 
Direct evaluation of these activation-based relations relies on responses to shared reference inputs. Responses from multiple sites can be collected during the same model traversal, and the responses or suitable comparison descriptors can be cached across comparisons. For previously unmeasured models, however, the corresponding responses still need to be acquired. We investigate a different use of these measurements: as supervision for learning how parameterized computations relate to the representations they produce. This leads to our central question:

\begin{tcolorbox}
   \textit{Can we learn to predict representation similarity scores in unseen models across different sites using only their architecture and trained weights?}
\end{tcolorbox}

In this paper, we demonstrate that this is indeed possible.
We study this question as a metanetwork learning task \citep{unterthiner2020predicting,schurholt2021self, zhou2023neural, zhou2023permutation, navon2023equivariant}:
given two internal sites from different trained models, we aim to predict how their representations are related. 
To do so, we need to know both which sites we are comparing and which relation we want to predict. The weights of the layer immediately before a site may miss earlier computations that shape its representation. In contrast, a single description of the whole model cannot distinguish between sites without additional information. 
As such, we represent each site using the computations that lead to it and train the metanetwork to learn relations measured from activations. This allows us to study how well trained computations can predict different properties of representations, including their geometry \citep{kriegeskorte2008representational, raghu2017svcca}, neighborhood structure \citep{gwilliam2022beyond}, and predictive behavior \citep{alain2016understanding, ding2021grounding}.
However, model parameters do not always make these relations easy to recognize.
Parameter symmetries \citep{hecht1990algebraic, godfrey2022symmetries, zhao2026symmetry} pose one source of challenge for this learning problem (see Fig.~\ref{fig:param_to_repre}).
Even when tensor shapes match, relabelings of hidden units can yield different parameter configurations that produce the same network function \citep{sussmann1992uniqueness, chen1993geometry, ainsworth2023git}. 
Distances computed directly in parameter space can conflate differences in learned computation with differences in coordinate choice \citep{phuong2020functional, vlavcic2021affine, stock2023embedding, bona2023parameter, simsek2021geometry}.
Representational correspondence can also extend beyond these parameter symmetries.
For example, orthogonally rotating features preserves their inner-product geometry while changing the individual feature values
\citep{kornblith2019similarity}.
Different upstream computations may produce similar representation geometry without
being connected by these transformations.
The learning problem is thus both to recognize equivalent parameterizations and to characterize which differences in trained computations matter for representational correspondence. 

To this end, we introduce \emph{Parameter-based Analysis for Relational Inference of Sites} ($\mathsf{PARIS}$), a framework for learning internal representation relations from the computations that produce them (see Fig.~\ref{fig:illstration}).
We build on graph metanetworks that encode network parameters with their connectivity \citep{lim2024empirical, kofinas2024graph}.
To accommodate variation in architecture,  we represent each site's upstream computation as a graph encoding its architecture, trained parameters, and inference state.
To account for permutation symmetries (e.g., hidden-coordinate permutation and head permutation), the encoding maps relabelings of hidden units and channels to graph relabelings while preserving parameter correspondences across successive layers.
For attention, we encode the composed query-key and value-output operators, which are invariant to the changes of basis within their internal spaces.
To learn representational correspondences beyond those induced by parameter symmetries, we train the graph metanetwork using representation similarity 
between sites across independently trained models.
At search time, it predicts these relations for unseen models from their site graphs, without requiring new base-model forward passes.

We consider two variants that differ in when information from the two sites is combined.
\textit{(i)} $\mathsf{PARIS}$-Ind maps each site graph to a fixed embedding and compares embeddings using relation-specific distance heads. 
These embeddings can be computed once and reused across comparisons. 
\textit{(ii)} $\mathsf{PARIS}$-Joint uses cross-attention between the two site graphs before predicting their relation, allowing intermediate features of each site to interact with the other.
Comparing the two variants tests whether such interactions improve relation prediction beyond what can be achieved with
site embeddings.

\textbf{Our contributions} are fourfold: 
\textit{(i)} 
we formulate and study the prediction of representation similarity
between internal representations in independently trained deep networks, using their parameterized upstream computations as inputs, 
\textit{(ii)} we develop a graph-metanetwork framework for this task and compare reusable site embeddings with joint processing of site pairs, 
\textit{(iii)}
we analyze the structural properties of both estimators and derive conditional bounds connecting prediction error to retrieval regret and stitching-candidate selection, and
\textit{(iv)}
we evaluate how well our models predict comparison scores, retrieve layers, and select stitching candidates across model populations of varying widths and depths, showing that training on the representation similarity on parameterized computations can support candidate selection without using stitching outcomes as supervision.

\section{Related work}

Our work connects the comparison of internal representations across models with learning from neural network artifacts. We discuss the two topics here;  App.~\ref{app:extend_related_work} discusses extended related work.

\noindent\textbf{Comparing and matching neural representations.}
Comparisons across independently trained networks have been used to study recurring features and shared representational structure \citep{pmlr-v44-li15convergent,gurnee2024universal,huh2024position}.
Activation-based approaches compare shared subspaces using SVCCA and PWCCA \citep{raghu2017svcca,morcos2018insights}, response geometry using CKA \citep{kornblith2019similarity}, or regularized linear prediction using GULP \citep{boix2022gulp}. Model stitching evaluates compatibility by connecting parts of trained networks \citep{bansal2021revisiting,csiszarik2021similarity}, while component reassembly methods select fragments from model collections \citep{yang2022deep}. Direct evaluation of activation-based relations depends on responses to reference inputs.
In this paper, $\mathsf{PARIS}$ addresses the complementary problem of predicting these relations from the \textit{parameterized computations} that produce the representations.

\noindent\textbf{Learning from neural network parameters.}
Metanetworks learn from trained networks to predict model properties, construct representations of their weights, or produce parameter updates.
DWSNets and neural functional networks account for hidden-unit permutation symmetries \citep{navon2023equivariant,zhou2023permutation}, while graph-based metanetworks encode parameters with connectivity across diverse architectures \citep{lim2024graph,kofinas2024graph}.
Related work learns parameter alignment \citep{navon2024equivariant} and predicts functional similarity from 
synaptic connectivity in simulated spiking networks \citep{haber2022learning}.
Building on this perspective, our work draws on graph-based methods for processing trained networks. We use these methods to study a different prediction problem: learning activation-derived relations between internal sites from their parameterized upstream computations, and used for retrieval and stitching candidate selection.

\section{Parameter-based Analysis for Relational Inference of Sites}

\label{sec:proposed_method}

Our goal is to
predict representation similarity between internal sites from architecture and trained weights alone 
without evaluating the compared models on probe inputs at inference. 
Motivated by the geometry and symmetries of weight space, we represent each site by a graph of its upstream computation, and train a metanetwork to recover similarity scores computed from reference responses.

\noindent\textbf{Notation.} Let $\mathcal{F}_{\rm tr} = \{f_{\bm{\theta}_m}\}_{m=1}^M$ denote the population of models available for meta-training, where each base model $f_{\theta_m}:\mathcal{X}\rightarrow\mathcal{Y}_{\mathrm{task}}$ is parameterized by $\bm{\theta}_m\in \Theta$.
Each index $m$ refers to one independently initialized and trained model.
Model $m$ contains a set of internal representation sites $\mathcal{S}_m$.
We write $u=(m,s)$, with $s\in\mathcal{S}_m$, for an internal representation site in model $m$, 
where a \emph{representation site} is a tensor-valued location in the forward computation, e.g., the output of a residual block or the class-token activation after a Transformer block. We write $\mathcal{C}_u$ for the architecture and trained parameters of the computation upstream of $u$, which allows us to distinguish two sites with identical local layer parameters whose preceding parameterized computations produce different features.

\subsection{Problem formulation}
\label{sec:problem_formulation}

Let $\mX = (\vx_i)_{i=1}^n$ be a shared
probe dataset from a fixed reference distribution on $\mathcal{X}$.
For a site $u$, let $\bm{r}_u(\vx)\in\mathbb{R}^{p_u}$ denote its
representation of input $\vx$.
We apply a fixed extraction rule at each site, such as selecting a token, pooling spatial features, or vectorizing the output.
Its reference responses are $\mH_u = (\vr_u(\vx_i)^\top)_{i=1}^n\in\mathbb{R}^{n\times p_u}$, where rows correspond to the same ordered inputs across sites while feature dimensions may differ. 
We keep the reference distribution, preprocessing, and extraction rules fixed across all comparisons.
Each relation type $\mathsf{h}$ defines a target distance based on activations $y_{uv}^{(\mathsf{h})}=\delta_{\mathsf{h}}(\mH_u,\mH_v)$ for comparing the activations at two sites. 
In this work, we consider CKA-based dissimilarities
\citep{kornblith2019similarity,brockmeier2017quantifying}, GULP distance \citep{boix2022gulp}, MutualNN dissimilarity \citep{huh2024position}, or held-out linear-readout disagreement (see App.~\ref{app:relations} for further details).
For measures naturally defined as similarities, such as CKA or MutualNN, we use the corresponding dissimilarity so that all targets have the same orientation.

We split the independently trained models into disjoint meta-training, validation, and test populations before constructing site pairs.
For every pair of sites $(u,v)$ and every relation type $\mathsf{h}$, we construct the example $(\mathcal{C}_u,\mathcal{C}_v,y_{uv}^{(\mathsf{h})})$.
Let $\mathcal{T}_{\mathrm{tr}}$ denote the collection of all such examples constructed from the meta-training population.
A metanetwork $g_{\bm{\omega}}$ parameterized by $\bm{\omega}$  is trained by minimizing 
\begin{equation}
\underset{\bm{\omega}}{\min} \;
\frac{1}{|\mathcal{T}_{\mathrm{tr}}|}
\sum_{(\mathcal{C}_u,\mathcal{C}_v,y_{uv}^{(\mathsf{h})})\in\mathcal{T}_{\mathrm{tr}}}
\ell_{\mathsf{h}}(g_{\bm{\omega}}(\mathcal{C}_u,\mathcal{C}_v,\mathsf{h}),y_{uv}^{(\mathsf{h})}).
\end{equation}
We define the loss $\ell_h$ in Sec.~\ref{sec: Graph metanetwork estimators}. At inference, we evaluate whether $g_\omega$ can predict the representation similarity score for sites from independently trained
models that were not seen during meta-training.

\subsection{Encoding the computation that produces a representation}

We represent each internal representation site $u$ by the parameterized computation upstream of that site.
To construct its graph, we build on the parameter-graph representations of neural networks by
\cite{lim2024graph,kofinas2024graph}.
The graph contains the operations and trained parameters that produce the representation at $u$, with $u$ marked as the target site, given by
\begin{equation}
    G_u = \Psi(\mathcal{C}_u) = (V_u,E_u,\mX_u,\mA_u,t_u),
\end{equation}
where 
$\Psi$ is deterministic, 
$V_u$ and $E_u$ are the nodes and directed edges, $\mX_u$ and $\mA_u$ contain node and edge attributes, and $t_u$ identifies the target site.
The graph encoding preserves the structure of the upstream computation and the correspondences between its parameter tensors.
For each site, we build a graph that captures the operations, parameters, and inference state needed to produce its representation. In a fixed network, changing the parameters \begin{wrapfigure}[21]{r}{0.48\textwidth}

    \centering
    \includegraphics[width=\linewidth]{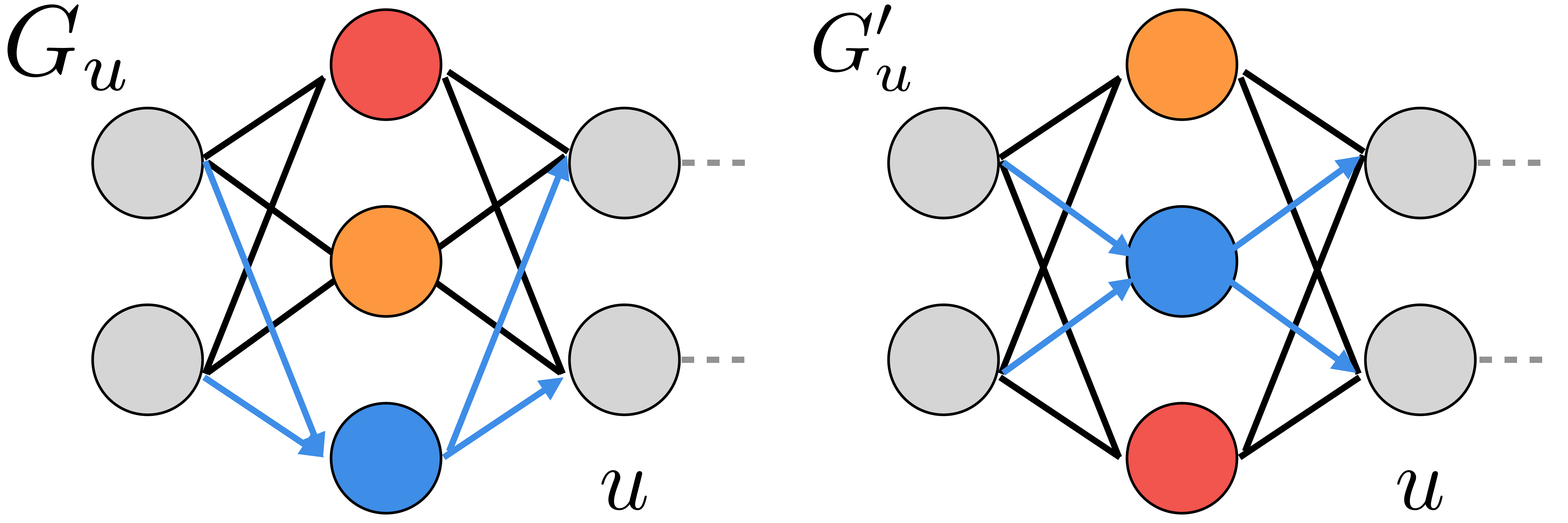}
    \vspace{-5mm}
    \caption{%
        Same function, different parameterization.
        The graph only changes by a relabeling.
    }
    \label{fig:site_graph_perm}

    \vspace{3mm}

    \includegraphics[width=\linewidth]{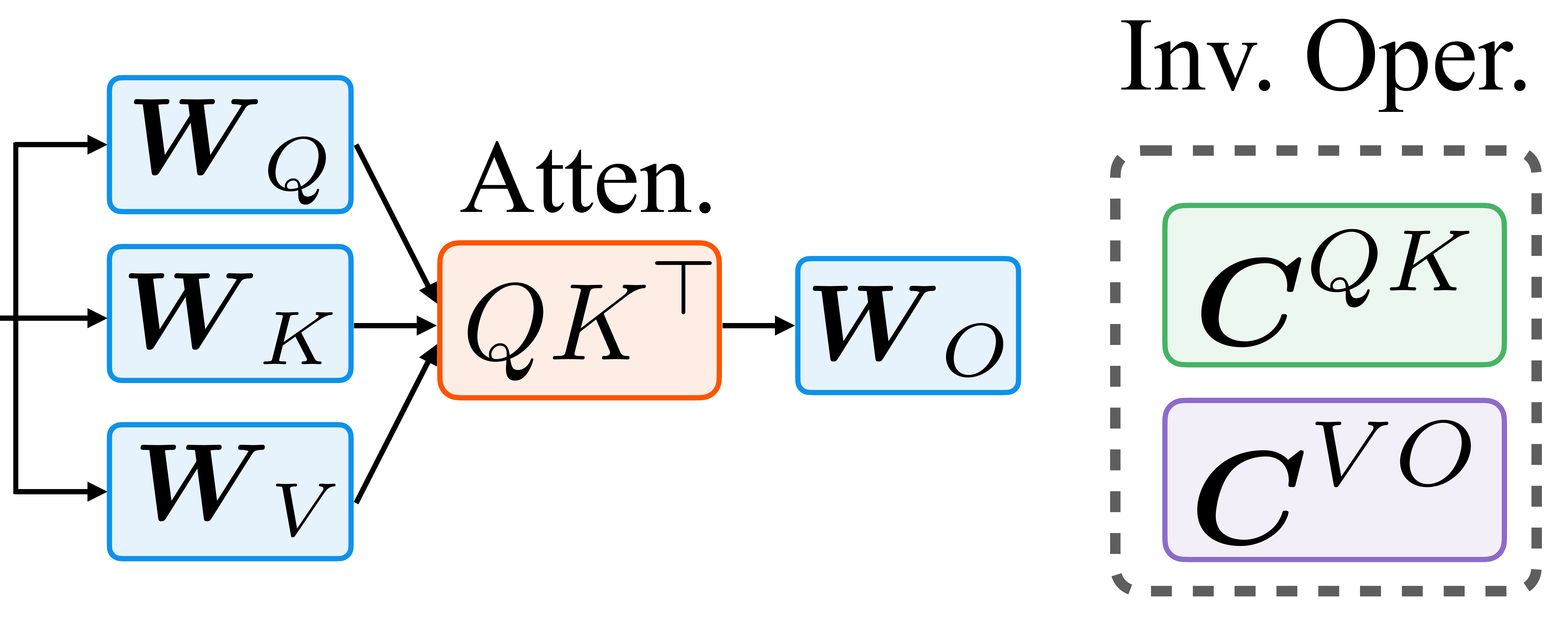}
    \vspace{-5mm}
    \caption{%
        For attention, we encode the composed operator. 
        The graph is invariant to basis changes in query-key and value-output spaces. 
    }
    \label{fig:site_graph_atten}

    \vspace{-4mm}
\end{wrapfigure} in later layers does not affect the representation at that site. Thus, we include only the computations that come before the selected site. A whole-network encoder could also produce site-specific predictions, however, it would also process parameters that do not affect the selected representations (see App.~\ref{app:whole_network_encoder} for details). Our approach uses only the relevant computations, which can also reduce overfitting when only a small number of training models are available (as shown in App.~\ref{app:whole_network_encoder}).

These correspondences are also important since neural parameters have permutation symmetries
(see Fig.~\ref{fig:site_graph_perm}).
For example, consider two linear maps, 
hidden coordinate $i$ corresponds to row $i$ of the first weight matrix and column $i$ of the second.
Simultaneously permuting these
indices gives a different parameterization of the same network
function \citep{hecht1990algebraic,ainsworth2023git, lim2024empirical, navon2023equivariant}.
In the graph, this transformation only relabels the corresponding nodes.
Our graph ties them to the same coordinate node.
The same construction is used for ResNet channels, Transformer feed-forward coordinates, and attention heads. 
Thus, the graph encoder can treat these discrete
reparameterizations as graph relabelings.

Attention additionally admits continuous changes of basis in the query-key and value-output spaces  \citep{vaswani2017attention, theus2025generalized}. 
Unlike permutation symmetries, these transformations cannot be represented
as graph relabelings, thus they are not handled directly by the graph construction.
In our work, we account for these symmetries before graph encoding by considering the corresponding composed query-key
and value-output operators.
The attention depends on its four factors through the attention circuits via $\mC^{\rm QK}=\mW_{Q}\mW_{K}^{\top}$ and $\mC^{\rm VO}=\mW_{V}\mW_{O}$, where  $\mC^{\rm QK}$ determines the bilinear interaction underlying the attention logits and $\mC^{\rm VO}$ determines how attended residual-stream features are mapped back into the residual stream (see Fig.~\ref{fig:site_graph_atten} for an illustration). 
We encode their entries as edges
associated with a shared head node, preserving the pairing
between the QK and VO operators, thus retaining the invariances. 
The construction for affine self-attention, including query, value, and output biases at residual-stream output sites, is in App.~\ref{app:exp_details}.

We remark that the transformations we consider, including hidden-coordinate permutations and attention-basis changes, depend on the network architecture.
In our work, we do not cover all reparameterizations that preserve the network's input-output function, because they may still change the internal representations we want to compare \citep{braun2025not, klabunde2025similarity}.
For example, rescaling intermediate features and compensating downstream can leave the network output unchanged while altering the selected representation and its similarity to other representations \citep{neyshabur2015path, grigsby2023hidden}.  Conversely, when the rescaling and compensation both occur upstream of the queried site, its representation and reference similarity scores remain unchanged. 
We empirically examine this latter case in App.~\ref{app:Sensitivity to positive-rescaling}, finding relatively stable prediction correlations in the  ResNet populations even though the  exact invariance is not guaranteed.

\subsection{Graph metanetwork estimators}
\label{sec: Graph metanetwork estimators}

Given two site graphs $G_u$ and $G_v$, we aim to predict their internal representation dissimilarities.
We study two variants that use the same supervision but differ in when
the two sites can interact. $\mathsf{PARIS}$-Ind encodes each site independently into a reusable embedding and compares the two embeddings only afterward. $\mathsf{PARIS}$-Joint lets the representations of the two site graphs
interact before it predicts their dissimilarities. This comparison tests whether a fixed representation of each site is sufficient or whether pair-dependent processing provides additional
information (see App.~\ref{app:relation_retrieval_setup} for the trade-off). 

\noindent\textbf{Independent site embeddings.} The independent estimator, $\mathsf{PARIS}$-Ind,  maps each site graph to a fixed embedding using a shared
graph metanetwork, $\vz_u = \mathrm{Enc}_\phi(G_u) \in \R^k$, where $\phi$ denotes the learnable encoder parameters.
The encoder first maps the attributes of each node $i$ to an initial hidden state $\vh_i^{(0)}\in\R^d$. During message-passing round $r$, an edge $i\rightarrow j$ with attributes $\va_{ij}\in\R^{d_e}$ produces a message $\vm_{ij}^{(r)}=\mW \vh_i^{(r)}\odot \psi_{\rm edge}(\va_{ij})$, and the receiving node is updated as 
\begin{equation}
    \vh_j^{(r+1)}=\operatorname{LN}
    \left(\vh_j^{(r)} + \psi
    \left(
    \left[
    \vh_j^{(r)}, 
    \operatorname{AGG}_{\rm s}
    \left(
    (\vm_{ij}^{(r)})_{i\in\mathcal{N}^-(j)}
    \right)
    ,
    \operatorname{AGG}_{\rm m}
    \left(
    (\vm_{ij}^{(r)})_{i\in\mathcal{N}^-(j)}
    \right)
    \right]
    \right)
    \right),
\end{equation}
where $\psi_{\rm edge}:\R^{d_e}\rightarrow\R^d$ and $\psi:\R^{3d} \rightarrow \R^d$ are multilayer perceptrons, $\operatorname{AGG}_{\rm s}$ and $\operatorname{AGG}_{\rm m}$ are aggregate functions of the incoming messages using their sum and mean, $\mW \in \R^{d\times d}$ is learned, $\mathcal{N}^-(j)$ is the incoming-neighbor set of node $j$, and $\operatorname{LN}$ denotes layer normalization.

After $R$ message-passing rounds, the encoder pools the final node
states both over the nodes associated with the target representation
site and over the complete graph.  A readout network maps the resulting
summary to $\vz_u\in\mathbb{R}^{k}$.  Given two sites
$u$ and $v$, relation $\mathsf{h}$ is learned from the distance
\begin{equation}
    d_{\phi,\mathsf{h}}(u,v)
    =
    \norm{\mW_\mathsf{h}(\vz_u-\vz_v)}_2,
    \label{eq:site-embedding-distance}
\end{equation}
that defines a pseudometric on the site-graph representation,  where $\mW_\mathsf{h}\in\R^{{k_{\mathsf{h}}} \times k}$ is a learned relation-specific metric head.
Thus, the graph metanetwork assigns each site an embedding that can be reused in comparisons with multiple other sites.
A monotone calibration then converts it to the scale of the target  $\widehat{y}_{uv}^{(\mathsf{h})} = q_{\mathsf{h}}(d_{\phi,\mathsf{h}}(u,v)) =  1-\exp(-\varepsilon_{\mathsf{h}}(d_{\phi,\mathsf{h}}(u,v)))$ for $\varepsilon_{\mathsf{h}}>0$. 
It is worth noting that Eq.~(\ref{eq:site-embedding-distance})  is related to multidimensional scaling through distance fitting,  which assigns coordinates to objects so that their pairwise distances approximate observed dissimilarities \citep{ cox2008multidimensional}. 
In our setting, the objects are internal sites, and the dissimilarities are activation-derived relations.

\noindent\textbf{Joint comparison metanetwork.}
$\mathsf{PARIS}$-Joint processes a pair of sites together.
Given $G_u$ and $G_v$, it first encodes the node tokens of both
graphs and then applies bidirectional cross-attention between them.
The model allows information from $v$ to affect how it
processes $u$, and vice versa, before it produces the final scores.
It predicts all supervised relation types for that site pair:
\begin{equation}
    \widehat{\vy}_{uv}
=
\operatorname{Comp}_{\vartheta}(G_u,G_v)
=
\left(
\widehat y_{uv}^{(1)},\ldots,\widehat y_{uv}^{(H)}
\right)
\in[0,1]^H,
\label{eq:dis_joint_comp}
\end{equation}
where $\vartheta$ is the learnable parameters of the comparator and $\widehat y_{uv}^{(\mathsf h)}$ are the learned dissimilarity associated with the relation $\mathsf h\in\mathcal{H}$.
We refer \emph{joint} to this pair-dependent processing. 
Unlike $\mathsf{PARIS}$-Ind, $\mathsf{PARIS}$-Joint does not assign a fixed representation to a site independently of its comparison partner.

\noindent\textbf{Comparing sites across two networks.}
Both variants predict one score for one pair of internal representation sites at a time.
To compare the sites of two networks, we apply the estimator to all cross-network pairs. 
Let models $A$ and $B$ contain the site sets
$\mathcal{S}_A$ and $\mathcal{S}_B$.
For relation type $h$, the target and predicted pairwise
dissimilarity matrices are respectively defined by
\begin{equation}
\mD_{\mathsf{h}}(A,B)_{st}
=
y_{(A,s),(B,t)}^{(\mathsf{h})},
\qquad
\widehat{\mD}_{\mathsf{h}}(A,B)_{st}
=
\widehat y_{(A,s),(B,t)}^{(\mathsf{h})},
\end{equation}
where $s\in\mathcal{S}_A$ and $t\in\mathcal{S}_B$.
For $\mathsf{PARIS}$-Ind, the entries of $\widehat{\mD}_{\mathsf{h}}$ come from the
distance in Eq.~(\ref{eq:site-embedding-distance}), and for $\mathsf{PARIS}$-Joint, they come directly from the joint comparator in Eq.~(\ref{eq:dis_joint_comp}).

Each row of $\mD_{\mathsf{h}}(A,B)$ ranks the sites of model $B$ relative to one query site in model $A$. Thus,  $\mD_{\mathsf{h}}(A,B)$ can be used for layer retrieval, stitching-candidate selection, and sequence alignment \citep{kornblith2019similarity, bansal2021revisiting, nguyen2023cross}. 
Transforming this matrix to one scalar ``network distance'' \citep{klabunde2025similarity} would introduce additional assumptions on how sites
should be aggregated or matched, 
and a separate set-level aggregation needs to be applied (e.g., see App.~\ref{app:exp_details} for  Hausdorff construction \citep{huttenlocher1993comparing}).
Thus, in this work, we keep site-pair prediction separate from any set-level aggregation, and leave it for future exploration.

\noindent\textbf{Shared supervision and training loss.}
We train  $\mathsf{PARIS}$-Ind and $\mathsf{PARIS}$-Joint on the same meta-training site pairs.
For a training pair $p=(u,v)$, let $y_p^{(\mathsf{h})}$ denote the target dissimilarity and $\widehat y_p^{(\mathsf{h})}$ the model prediction.
We use a smooth regression loss $\mathcal{L}_{\rm reg}^{(\mathsf{h})} = \frac{1}{|\mathcal{P}|}\sum_{p\in\mathcal{P}}
\operatorname{SmoothL1}(\widehat y_p^{(\mathsf{h})},y_p^{(\mathsf{h})})$ with an ordinal loss that encourages the predicted distances to preserve the ordering of the target relations 
\begin{equation}
    \mathcal{L}_{\rm rank}^{(\mathsf{h})} = \E_{a,b\sim\mathcal{P}}\log(1+\exp[-\tau\,\operatorname{sign}(y_a^{(\mathsf{h})}-y_b^{(\mathsf{h})})(\widehat y_a^{(\mathsf{h})}-\widehat y_b^{(\mathsf{h})})]).
\end{equation}
The ranking term has a pairwise logistic loss used in Bradley-Terry form \citep{bradley1952rank} and RankNet \citep{burges2005learning}, which penalizes the model when the predicted ordering disagrees with the target ordering.
For $\mathsf{PARIS}$-Ind, 
we additionally discourage embedding collapse using $\mathcal{L}_{\rm var}= \max(0,\eta-\frac{1}{k}\sum_{j=1}^{k}\operatorname{sd}_{u}(\vz_{u,j}))$, where $\operatorname{sd}_{u}$ is the empirical standard deviation.
We select models using the mean correlation between predicted and target dissimilarities.
While we train on pairs of representation sites, 
this objective is not standard binary contrastive learning \citep{hadsell2006dimensionality, khosla2020supervised} as there are no
positive-vs-negative pair labels. The regression and ranking objectives train the estimators to recover the magnitude and ordering of representation relations. A comparison with contrastive learning and its extension is provided in App.~\ref{app:exp_details}.

\subsection{Theoretical properties}

We summarize the theoretical properties of the proposed methods here and provide the formal statements, assumptions, detailed constructions,  and proofs in App.~\ref{app:Proof and further theoretical results}.
We study three main questions: 
\textit{(i)} Which properties follow from how we represent and compare sites?
\textit{(ii)} How do independent and joint estimators differ in the scores they can represent?
\textit{(iii)} Under what conditions do accurate relation predictions support layer retrieval and stitching-candidate selection?

Both $\mathsf{PARIS}$ variants respect the symmetries accounted for by the site-graph construction and are symmetric under exchange of the compared sites. These properties follow from the encoding and comparator design. We refer to App.~\ref{app:Symmetries of the site graphs and predictors}, where we provide the statement and proofs.
We next relate the geometric restrictions imposed by the independent and joint comparison architectures.

\begin{proposition}[Reference geometry; informal]
\label{prop:ref_geometry}
For $N\ge2$ fixed sites evaluated on a common
reference sample, we  assume nonzero centered Gram matrices for chordal linear CKA and fixed hard readouts with common class semantics for readout disagreement.
Chordal linear CKA and GULP can be expressed as Euclidean distances between reference descriptors assigned separately to each site. Hard-readout disagreement can be expressed as a
squared Euclidean distance.

\end{proposition}

\begin{proposition}[Structural properties; informal]
\label{prop:main-structure}
Eq.~(\ref{eq:site-embedding-distance}) is a pseudometric,
and  $q_{\mathsf h}\circ d_{\phi,\mathsf{h}}$ is also a pseudometric and preserves the candidate ranking.
$\mathsf{PARIS}$-Joint compares each site with its partner. This gives the estimator more flexibility, but it does not enforce the triangle inequality.
\end{proposition}

The formal statements and proof of Prop.~\ref{prop:ref_geometry} and Prop.~\ref{prop:main-structure} are in App.~\ref{app:Geometry of independent and joint scores}. 

We further analyze when accurate scores help us select useful candidates.
For layer retrieval, we bound how much worse the selected candidate can be than the best candidate in the list, measured by reference dissimilarity. 
The bound depends on the MSE of the predicted scores and on the number of candidates.
For stitching, we consider two sources of error separately. First, screening may remove useful candidates. Second, after screening, we may choose the wrong candidate because we evaluate the shortlist on a finite, independent validation sample. If the list keeps a candidate with the smallest predicted dissimilarity, we can bound the error introduced by screening.
We refer to App~\ref{app:Score error, layer retrieval and stitching} for further statements and details. 
These results show that accurate comparison scores can help with screening. 
While they characterize the consequences of accurate predictions, whether such accuracy transfers to unseen models is evaluated empirically in the next section.

\section{Experiments}
\label{sec:exp}

We begin with a qualitative analysis of representation-site correspondence across network depths, which examines the structure of the pairwise relations predicted by our method. We then quantitatively evaluate the performance of $\mathsf{PARIS}$  on two representation search tasks: \textit{(i)} representation relation prediction and \textit{(ii)} layer retrieval. We further assess its utility for stitching candidate selection. The experimental protocol and implementation details are provided in App.~\ref{app:relation_retrieval_setup}. Additional results, including an ablation study and meta-training across diverse architectures, are in App.~\ref{app:more_exp}.

\subsection{Representation-site correspondence}
\label{sec: Representation-site correspondence}

To illustrate whether our method recovers meaningful structure across networks,  we first use a qualitative example to inspect how it compares internal representations across networks of different depths.
We examine a cross-depth correspondence map between ResNet56 and ResNet20 \citep{he2016deep} to explore which sites in the shallower network are preferred counterparts of sites in the deeper network. We compare models trained on CIFAR-10 and the predictions are produced by $\mathsf{PARIS}$-Ind for  HSIC-based CKA \citep{kornblith2019similarity}.
We index the 27 selected \begin{wrapfigure}[14]{r}{0.6\textwidth}  
           \captionsetup{skip=3pt}
     \includegraphics[width=0.6\textwidth]{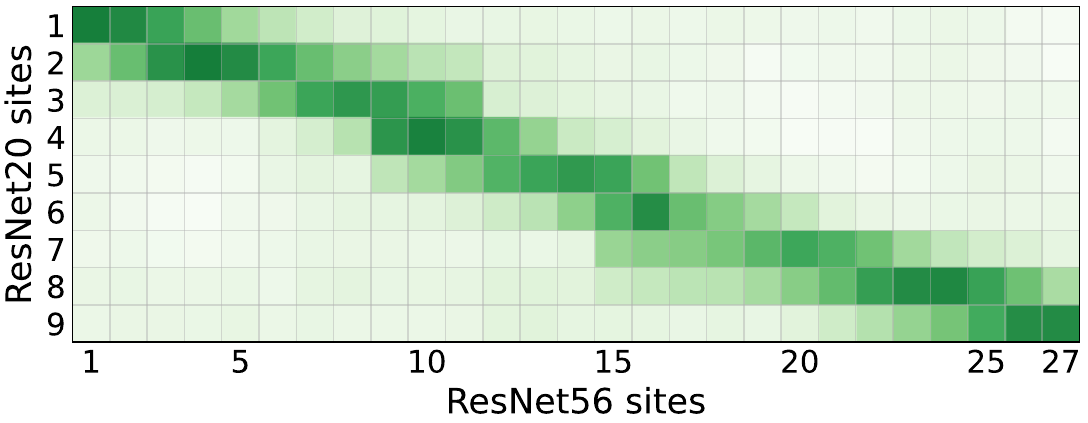}
    \caption{
    Predicted correspondence between networks of different depths: between 27 ResNet56 sites (columns) and nine ResNet20 sites (rows), ordered from earlier to later sites.
    Darker green indicates stronger correspondence.
    }
    \label{fig:site_correspondence}
\end{wrapfigure} ResNet-56 sites
and the nine selected ResNet-20 sites
in forward computational order.
Fig.~\ref{fig:site_correspondence} exhibits a depth-ordered band of stronger predicted correspondence.
As the query moves from early to later ResNet56 sites, the region of stronger correspondence moves predominantly toward later ResNet20 sites. 
It is qualitatively consistent with prior CKA analyses reporting stage-dependent similarity in residual networks and correspondence across network depths \citep{kornblith2019similarity,nguyen2021do}.
This pattern is approximately stagewise, \textit{many-to-one} correspondence under the 
relation, i.e., consecutive sites in the deeper network share similar or neighboring preferred counterparts in the shallower network. 
Directly computing the reference matrix requires running both base
models, storing their responses, and evaluating CKA
for every site pair.
In contrast, $\mathsf{PARIS}$-Ind encodes each site once and reuses the
resulting embedding across comparisons at inference. 
Their complexity analysis is in App.~\ref{app:relation_retrieval_setup}.

\subsection{Representation-score prediction}
\label{sec:Relation prediction}

\begin{wrapfigure}[17]{r}{0.65\textwidth}  
\vspace{-8 mm}
       \captionsetup{skip=3pt}
     \includegraphics[width=0.65\textwidth]{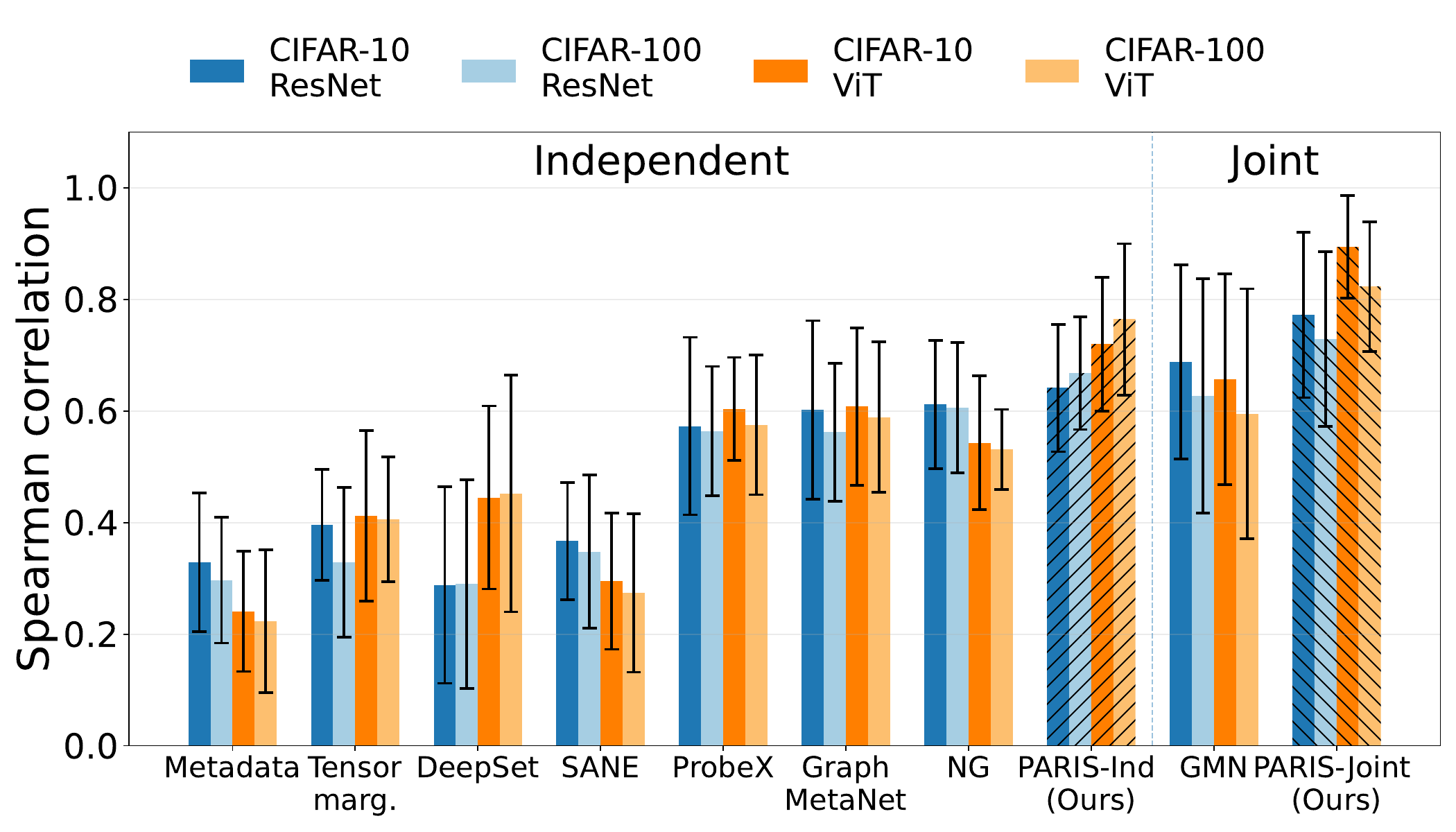}
     \vspace{-5 mm}
    \caption{Spearman correlation ($\uparrow$) between predicted and ground-truth representation relations across ResNet and ViT populations on CIFAR-10 and CIFAR-100.
    }
    \label{fig:relation_prediction}
\end{wrapfigure}We next evaluate the score prediction, i.e., whether a predictor can estimate the reference dissimilarity between two internal representations from the computations that produce them.
Each test example consists of one representation site of each of two distinct and independent models in the test partition and within-model site pairs are excluded.
The predictor receives the sites' upstream architectures and trained parameters, but not their representations on the reference probe inputs.
We use the corresponding reference score to evaluate the prediction.
We consider five comparison criteria: chordal linear CKA distance, HSIC-based CKA dissimilarity, GULP distance, MutualNN dissimilarity, and disagreement between fitted linear readouts on held-out data.
We evaluate ResNet and ViT \citep{dosovitskiy2021an} populations on CIFAR-10 and CIFAR-100.
For each architecture family and downstream task, we construct a population of models with varying widths and depths. Representation sites correspond to tensor-valued locations in each model's forward computation.
We compare $\mathsf{PARIS}$ against eight baselines:
\textit{(i)} architectural metadata, which use upstream computational depth, operator-type counts, target tensor shape, and upstream parameter count,
\textit{(ii)} tensor marginal (weight statistics), which summarize the marginal distribution of entries in each upstream parameter tensor \citep{unterthiner2020predicting}, 
\textit{(iii)} DeepSets \citep{zaheer2017deep}, 
\textit{(iv)} SANE \citep{schurholt2024towards}, 
\textit{(v)} ProbeX \citep{horwitz2025learning}, 
\textit{(vi)} Graph Metanetworks \citep{lim2024graph}, 
\textit{(vii)} Neural Graph (NG) \citep{kofinas2024graph} without probe feature, and \textit{(viii)} Graph Matching Network (GMN) \citep{li2019graph}.
We adapt the baselines to predict comparison scores from their respective descriptions (except GMN as a joint method, the other baselines are independent approaches).
These comparisons evaluate different ways of predicting the same site relations from the available upstream computations.
We measure prediction quality using Spearman correlation between predicted and reference dissimilarities across five scores and three seeds. Detailed description of  baselines  and their adaptations 
are in App.~\ref{app:baseline adaptations}.

Fig.~\ref{fig:relation_prediction} reports the Spearman correlation between predicted and reference dissimilarities.
Among methods that encode representation sites independently, $\mathsf{PARIS}$-Ind obtains the best Spearman correlation in three of the four settings, with noticeably large gains on ViTs. Jointly encoding a pair of representation sites further improves performance.
These results 
indicate that $\mathsf{PARIS}$ learns representations that are informative about functional relationships between model internal representations,
with the joint variant providing the strongest agreement with ground-truth relation rankings.
The two variants, however, have different computational costs at search
time (inference).
$\mathsf{PARIS}$-Ind computes each site embedding once and reuses it
across all comparisons.
After this encoding step, searching a large candidate bank only requires
evaluating distance heads between stored embeddings.
$\mathsf{PARIS}$-Joint instead applies cross-attention separately to
each query-candidate pair.
Thus, the joint model trades higher score accuracy for more
pair-dependent computation, whereas the independent model is better
suited to repeated comparisons and large candidate banks.

\begin{figure}[t]
    \centering
     \captionsetup{skip=3pt}
    \includegraphics[width=1\linewidth]
    {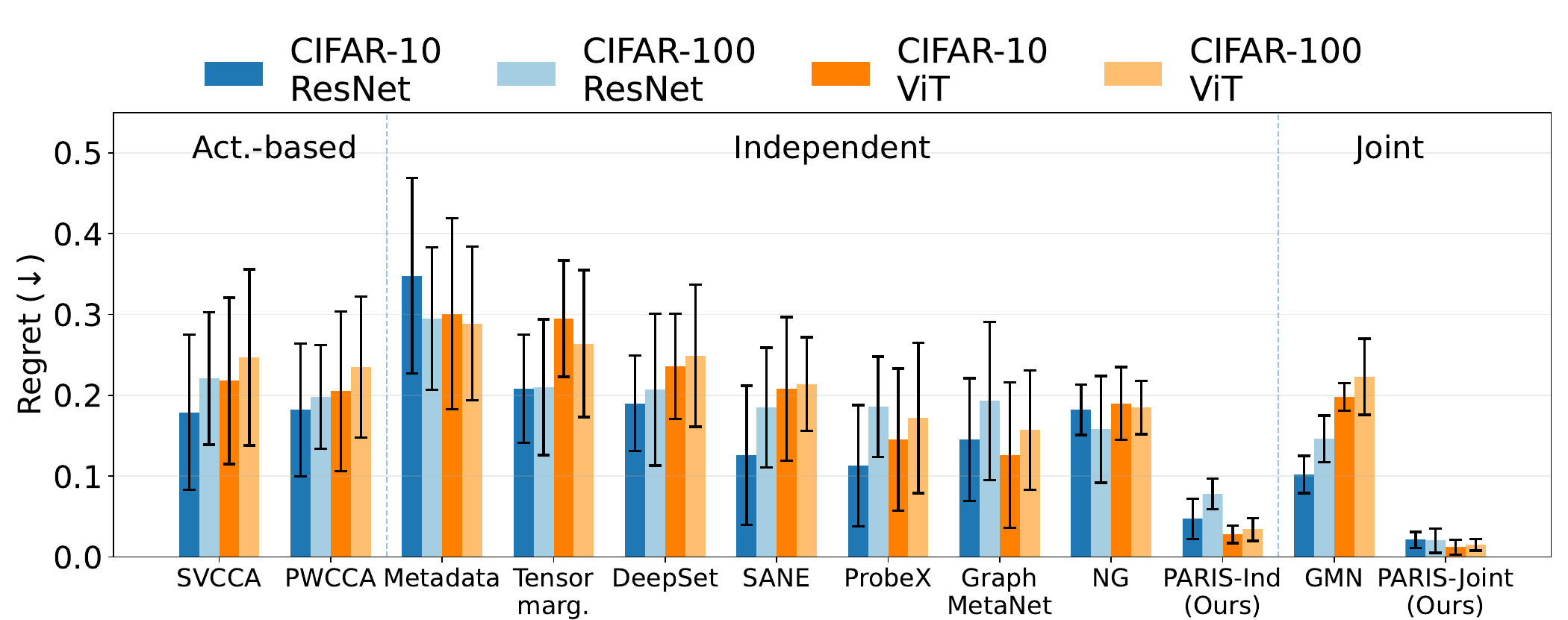}
    \caption{Layer retrieval across model populations, measured by Mean Regret@5 ($\downarrow$) for retrieving representation sites from a candidate bank spanning multiple test models.
    }
    \label{fig:retrieval}
\end{figure}

\subsection{Layer retrieval}

Next, we investigate the capabilities of $\mathsf{PARIS}$ on the layer retrieval task, where a query representation site is used to retrieve related sites from a bank spanning multiple models. This setting tests whether the learned relation predictors can identify close correspondences for individual queries. We use the same model populations, five reference relations, and model splits as in Sec.~\ref{sec:Relation prediction}. 
Both queries and candidate sites are drawn from test models.
For each reference relation, we rank the candidates by their predicted dissimilarity and return the top five.
We compare against the same competing baselines as in
Sec.~\ref{sec:Relation prediction}, and additionally include activation-based similarity methods SVCCA \citep{raghu2017svcca} and PWCCA \citep{morcos2018insights}.
Retrieval quality is measured using mean Regret@5, 
defined as the difference between the smallest reference dissimilarity among the five retrieved sites and the smallest reference dissimilarity in the full bank, averaged over queries. 
Lower Regret@5 indicates that the retrieved set contains a candidate closer to the best available match.

Fig.~\ref{fig:retrieval} presents the layer retrieval performance. 
Our method reduces retrieval regret across both architectures and datasets. Among methods that encode representation sites independently, ours achieves the lowest Regret@5. 
In addition, jointly modeling query-candidate pairs further improves retrieval quality and consistently outperforms GMN. These results show that the relations learned by our approaches transfer effectively from relation prediction to retrieval, enabling it to identify internal representation sites that closely match the best available candidates in a large cross-model bank.
Our metanetwork performs well on both relation prediction and layer retrieval, two complementary evaluations that assess global rank agreement across site pairs and the identification of close matches for individual queries, supporting amortized search over model collections.

\subsection{Stitching candidate selection}
\label{sec:Stitching candidate selection}

Finally, we evaluate the effectiveness of $\mathsf{PARIS}$ for stitching candidate selection, which identifies connections that preserve downstream performance under an adapter-fitting budget. We adopt the frozen-model stitching protocol of \citet{bansal2021revisiting} and evaluate ResNet-to-ResNet and ViT-to-ViT stitching on CIFAR-10 and CIFAR-100.
Each selection fixes a top model from the unseen trained models and provides every selector with the same pool of 20 eligible connections to bottom models from other unseen models. Each candidate connects the initial layers of a bottom model to the later layers of the top model through a trainable adapter (stitching layer). Each selector then chooses five candidates before their adapters are fitted.
We consider the same competing baselines as in Sec.~\ref{sec:Relation prediction} without training them on stitching outcomes, and additionally include SN-Net-H and KLAS-score, where SN-Net-H adapts SN-Net's heuristics for choosing model pairs and matching blocks \citep{pan2023stitchable}, and KLAS-score ranks candidates using normalized directional KL divergences between the class-probability distributions of fitted linear readouts \citep{sanyal2026klas}.
Both are evaluated as candidate selectors under the same fitting protocol (see App.~\ref{app:baseline adaptations} for their detailed adaptations). 
We report the stitching penalty \citep{bansal2021revisiting}, where the original top model's test accuracy minus the selected stitched model's test accuracy in percentage points.

\begin{figure}[t]
    \centering
     \captionsetup{skip=3pt}
    \includegraphics[width=1\linewidth]
    {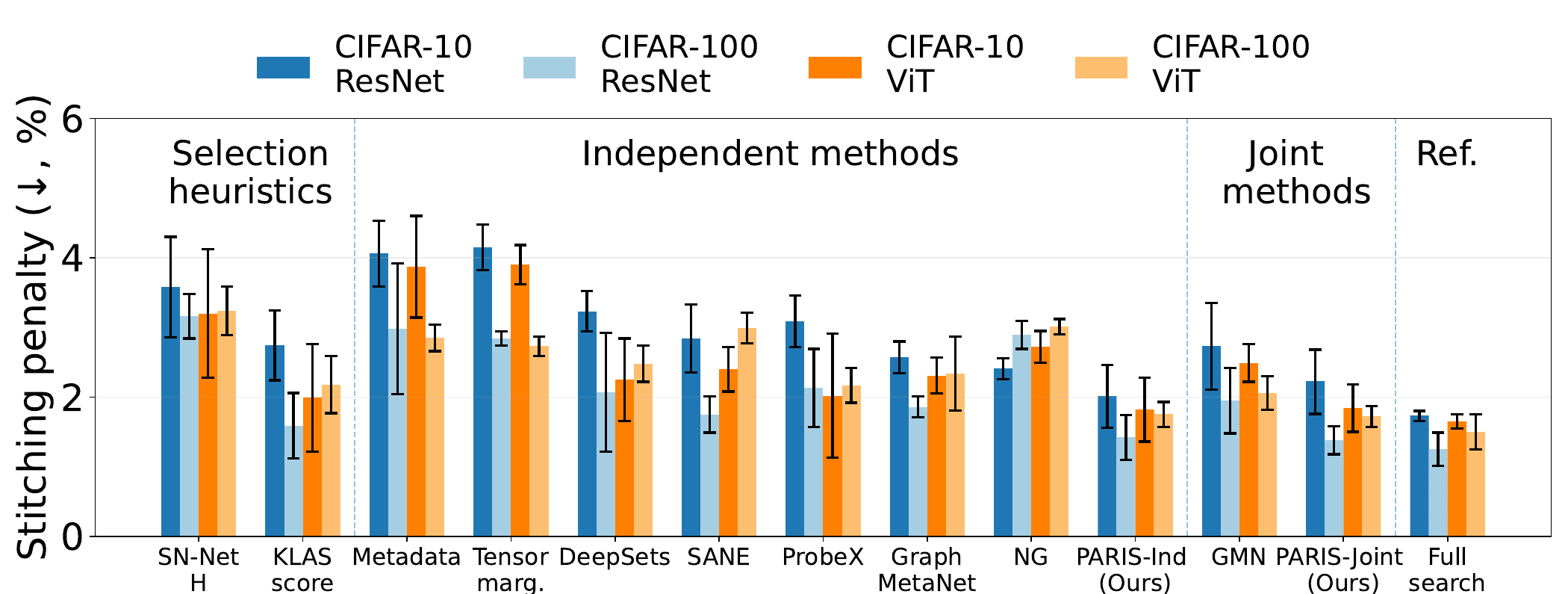}
    \caption{Stitching candidate selection for ResNet-to-ResNet and ViT-to-ViT connections on CIFAR-10 and CIFAR-100. 
    Performance is measured by the stitching penalty ($\downarrow$).
    }
    \label{fig:stitching_selection}
\end{figure}

Fig.~\ref{fig:stitching_selection} reports the performance of the stitching candidate selection methods.
We see that our approach achieves the lowest stitching penalty among budgeted selectors. Specifically, $\mathsf{PARIS}$-Ind performs best on CIFAR-10 while $\mathsf{PARIS}$-Joint performs best on CIFAR-100, approaching the performance of full search.
This suggests that the learned predictor provides a useful signal for identifying compatible connections before adapter fitting. 
Importantly, $\mathsf{PARIS}$ is not trained on stitching outcomes, suggesting that relational supervision derived from activation patterns transfers effectively to downstream model composition. 
At the same time, the consistent advantage of $\mathsf{PARIS}$-Joint in relation prediction does not translate into uniformly better stitching selection. The strong performance of $\mathsf{PARIS}$-Ind suggests that the learned site embeddings already capture much of the information needed for effective budgeted selection, without requiring pair-adaptive comparison.
In summary, these results support the usefulness of our methods for selecting stitching candidates.

\section{Discussion}

In this paper, we show that internal representational relations are \textit{learnable} from neural network artifacts across model populations.
Although activation-based similarity measures can be computed directly when suitable reference data are available, our method investigates whether these measurements can be learned from a population of trained models and subsequently predicted for unseen networks using only their trained computations. This enables representation comparison without requiring probe data or additional base-model inference at query time. 
Specifically, our approach produces compact site embeddings that can be reused for multiple comparisons and similarity measures, and it can compare each pair about 37 times faster and use about 11 times less peak memory. These benefits come at the cost of metanetwork training and initial graph encoding, making it most attractive for repeated comparisons or data-constrained settings and not one-off evaluations.
Our experiments show that these predictions support layer retrieval and the selection of stitching candidates before adapter fitting.
In addition, we provide initial evidence that our method can extend beyond unimodal models to image and text encoders from jointly pretrained CLIP-style models (see App.~\ref{app:more_exp}). 
One promising future direction is to extend the framework to multimodal model collections and investigate whether neural network artifacts can be utilized for cross-modal representational alignment (under the setting of independently trained unimodal vision and language models, which require defining cross-modal reference relations under (partially) paired or unpaired data).
This could provide a complementary parameter perspective (i.e., neural artifacts' perspective) on the Platonic Representation Hypothesis \citep{huh2024position} and on recent reassessments of
the extent and robustness of this convergence \citep{groger2026revisiting, koepke2026back, schnaus2026shared}.

\section*{Acknowledgments}
YEL acknowledges support from the Schmidt Futures Israeli Women’s Postdoctoral Award and the Viterbi Fellowship from the Faculty of Electrical and Computer Engineering at the Technion. 
DH acknowledges support from the Munich Data Science Institute (MDSI) via the MDSI Doctoral Fellowship. 
This research was supported by an Alexander von Humboldt Professorship.

\bibliography{bib}

\newpage

\appendix
\section*{Appendices}
\addcontentsline{toc}{section}{Appendices}

\startcontents[appendices]
\printcontents[appendices]{l}{1}{\setcounter{tocdepth}{2}}

\newpage

\section{Extended related work}
\label{app:extend_related_work}

\label{app:extended_related_work}

We expand on the two directions discussed in the main paper and their connections to model composition, parameter symmetries, graph comparison, and pretrained-model selection. Throughout, we distinguish defining a representation relation, learning to predict that relation, and using a predicted correspondence for a downstream intervention.

\noindent\textbf{Representation similarity.}
Comparing responses to shared inputs provides a way to study learned structure without assuming that individual coordinates correspond.
Representational similarity analysis compares patterns of dissimilarity among stimulus responses \citep{kriegeskorte2008representational}.
In artificial networks, convergent learning studies examine which features and subspaces recur across independent training runs
\citep{pmlr-v44-li15convergent}, while subsequent work investigates recurring neurons in language models \citep{gurnee2024universal} and broader representational convergence across models and modalities \citep{huh2024position}.
Analyses of width and depth further show that architectural changes affect the organization of learned representations \citep{nguyen2021do}.
These studies motivate searching for correspondences beyond identically indexed layers or neurons, while leaving the appropriate notion of correspondence dependent on the scientific question.

Different comparison methods emphasize different properties of the
responses. SVCCA and PWCCA characterize relationships between activation subspaces \citep{raghu2017svcca,morcos2018insights}, whereas CKA compares centered similarity matrices over examples
\citep{kornblith2019similarity}. Generalized shape metrics formalize representation comparison under specified transformation groups \citep{williams2021generalized}, and soft matching preserves information about individual-neuron tuning while accommodating different numbers of units \citep{khosla2024soft}. Neighborhood overlap supplies a more local criterion, as in the mutual-nearest-neighbor comparisons used by \citet{huh2024position}. These approaches do not need to induce the same ranking of candidate sites: agreement in global geometry, local neighborhoods, and individual-unit responses constitutes different evidence of shared structure.
\citet{balogh2023functional} show that robust and non-robust representations can be compatible in ordinary classification accuracy while differing in robustness compatibility, with the relationship changing across layers.  

Predictive criteria offer a complementary perspective. GULP defines a family of distances motivated by regularized linear prediction \citep{boix2022gulp}. Geometric and predictive interpretations are not mutually exclusive: \citet{harvey2024what} relate several similarity measures to the alignment of optimal linear readouts across a distribution of decoding tasks. Fitted probes instead evaluate information accessible under a particular readout family and training procedure \citep{alain2016understanding,hewitt2019designing}. Comparing the resulting predictions provides a task-specific behavioral relation, whose interpretation depends on the readout protocol and the chosen agreement statistic \citep{bo2024evaluating}.
Statistical evaluations, analyses of CKA reliability, and the ReSi benchmark demonstrates the importance of specifying what a similarity measure is intended to capture \citep{ding2021grounding,davari2023reliability,klabunde2025resi}.

Parameter-based representation comparison also has direct precedents. \citet{chen2023inner} introduce a filter-subspace similarity for convolutional networks whose filters are decomposed into model-specific atoms and coefficients shared across networks. This construction permits comparison from filter atoms without probing forward passes, with theoretical and empirical connections to activation-based similarities. Their applications also include retrieval of related models for knowledge transfer. Thus, neither avoiding probing activations nor searching a model collection alone distinguishes our approach. $\mathsf{PARIS}$ instead learns selected activation-defined relations between internal sites from their upstream computations, without imposing this shared-coefficient construction on the base models. It targets retrieval in held-out models from the studied populations.

A different analytical approach is developed by
\citet{gonzalez2026two}, who define tensor similarity for polynomial and multilinear network architectures. Their recursive tensor comparison accounts for cross-layer computation and parameter symmetries without sampling input responses. Localized variants compare output components and functional differences. Under their stated assumptions, a normalized similarity of one characterizes functional equivalence up to a positive scalar. This differs from our learning objective: $\mathsf{PARIS}$ predicts empirical activation relations for the base model architectures and we do not aim for providing an algebraic certificate of global functional equivalence.

$\mathsf{PARIS}$ is complementary to these comparison criteria. We use measured activation-derived relations as supervision and learn to predict them from parameterized upstream computations. Direct evaluation of these activation-defined targets requires obtaining the relevant responses or derived statistics, although
these can be cached. Our search-time predictor does not collect new base-model activations. Its predictions remain tied to the reference-input distribution and, where applicable, the readout protocol used to define the target. Thus, fidelity to the chosen relation and the usefulness of that relation for interpretation or reuse are separate questions.

\noindent\textbf{Model stitching, latent alignment, and component reuse.}
Feature-transfer studies show that reusing network components depends on both feature specialization and interactions between neighboring layers \citep{yosinski2014transferable}.
Representation-equivalence experiments study transformations that allow one representation to substitute for another
\citep{lenc2015understanding}. Model stitching develops this perspective by connecting the lower portion of one trained network to the upper portion of another and evaluating the resulting computation \citep{bansal2021revisiting,csiszarik2021similarity}. Stitching supplies a task-dependent test of compatibility, whereas an activation-based statistic compares the representations under a specified geometric or predictive criterion. These objectives can be related without being identical.

Latent-space communication methods address the coordinate mismatch between independently learned components. Relative representations describe examples through their relationships to reference samples \citep{moschella2023relative}, latent-space translation learns explicit alignments between separately trained spaces \citep{maiorca2023latent}, and products of invariances combine complementary descriptions to improve communication \citep{cannistraci2024from}. At the level of model collections, Deep Model Reassembly partitions pretrained models and selects blocks for a new network \citep{yang2022deep}. StitchNet uses CKA to guide the selection of pretrained fragments \citep{teerapittayanon2023stitchnet}, while Stitchable Neural Networks vary connections between pretrained anchors to obtain different accuracy--efficiency trade-offs \citep{pan2023stitchable}. These methods provide concrete motivations for locating useful components
within and across trained models.

More recently, KLAS uses KL divergence between the predictive distributions of linear probes attached to intermediate blocks, together with a normalized stitch score, to select candidate connections across pretrained models \citep{sanyal2026klas}. This is a direct precedent for using a readout-based relation to guide stitch selection. Its scoring relies on responses to data, whereas $\mathsf{PARIS}$ predicts site relations from parameterized computations at search time. Our meta-training targets are representation relations and \textit{not} the accuracy of assembled networks.  Any subsequent connector training and stitching validation remain separate
from relation prediction.

Our objective is to retrieve candidate sites under a selected representation relation, and we do not directly optimize an assembled network or fit a transformation between latent spaces. $\mathsf{PARIS}$ can provide a candidate-identification stage for a subsequent composition procedure, but a predicted match does not itself establish interchangeability. A retrieved component may still require an adapter, fine-tuning, and task-specific validation. Conversely, successful stitching does not establish that two original representations contain the same information: studies of stitching protocols demonstrate that functional alignment can be achieved despite substantial representational differences \citep{balogh2025not,smith2025functional}. We consequently distinguish representation retrieval from both the
construction and validation of a combined model.

\noindent\textbf{Connection to representation-space model merging.}
\citet{wang2026model} average sample-level representation kernels from independently trained models and optimize a student toward the averaged kernel using differentiable mutual-nearest-neighbor similarity. This suggests a possible extension in which predicted site relations help screen source models or select compatible sites before representation-space merging. Such screening would require validating which relations identify useful sources.  
Moreover, scalar pairwise relation predictions do not specify the sample-level Gram matrices needed for kernel averaging. The current $\mathsf{PARIS}$ objective is not an activation-free replacement for this merging procedure. A fuller integration could use $\mathsf{PARIS}$ for candidate selection while retaining response-based merging, or investigate a new prediction target carrying the additional kernel information. These are prospective extensions, not components of the present framework.

\noindent\textbf{Parameter symmetries and structured computation encodings.}
Neural-network parameters are not unique descriptions of network functions. Consistent hidden-unit relabelings produce equivalent parameterizations, a property studied in early analyses of weight spaces \citep{hecht1990algebraic, lim2024empirical, burgin2026beyond, lin2026spectral}. Permutation alignment is central to methods for comparing trained solutions and merging models, including neuron-aligned mode connectivity, Git Re-Basin, and matched averaging \citep{tatro2020optimizing,ainsworth2023git,wang2020federated}. Optimal-transport fusion further permits soft correspondences between models with different widths \citep{singh2020model}.
These methods distinguish parameter coordinates from learned computation.  Yet, Soft matching across different architectures does not have to be conflated with an exact function-preserving permutation.

An alternative to explicitly aligning each pair is to incorporate the symmetry into the parameter-processing architecture. DWSNets and neural functional networks respect the coupled action of hidden-unit permutations on neighboring weight tensors \citep{navon2023equivariant,zhou2023permutation}.
Neural Functional Transformers and equivariant polynomial architectures extend this approach with more expressive parameter interactions \citep{zhou2023neural,vo2025equivariant}.
Graph Metanetworks and neural-graph representations encode parameters with connectivity, supporting diverse input-network architectures \citep{lim2024graph,kofinas2024graph}.
We build on these representations and condition the prediction on a selected internal site. Its input contains the computation upstream of that site and the inference state needed to produce its representation, e.g., 
BatchNorm inference depends on population statistics in addition to trainable parameters \citep{ioffe2015batch}.

Attention introduces internal coordinate choices beyond hidden-unit permutations. The transformer circuits perspective separates the query-key
and output-value computations \citep{elhage2021mathematical}, while
Transformer-NFN formalizes internal basis symmetries and uses composed query-key and value-output operators \citep{tran2025equivariant}.
Related work exploits attention rotation symmetries for model fusion \citep{zhang2025beyond}. \citet{theus2025generalized} explicitly uses the composed QK and OV operators as canonical head representations in their Transformer-alignment framework, removing the corresponding internal invertible reparameterizations from the comparison.
$\mathsf{PARIS}$ builds on this existing reduction into its site-graph encoding. Thus, it avoids distinguishing factorizations that differ only by admissible internal basis changes.
This invariance concerns the supported attention computation, not
arbitrary transformations of every Transformer implementation.  Positional operations and other intervening transformations are respected when determining the admissible symmetry.

The invariances of a parameter encoder are also required to agree with the relation being predicted. Studies of network symmetries connect function-preserving parameter transformations to changes in internal representations \citep{godfrey2022symmetries}, and ScaleGMNs incorporate additional activation-dependent scaling symmetries \citep{kalogeropoulos2024scale}.
Nevertheless, preserving the end-to-end function does not guarantee that an internal representation retains its geometry. In particular, linear CKA is invariant to coordinate permutations and orthogonal transformations, but not arbitrary anisotropic rescaling \citep{kornblith2019similarity}.
For $\mathsf{PARIS}$, symmetry handling removes irrelevant coordinate choices while retaining distinctions that can affect the target site relation. It is not a claim of invariance to all functionally equivalent parameterizations.

\citet{theiss2026parameter} further sharpens this distinction: they characterize how parameter symmetries reshape representations through addition, duplication, and scaling, and derive conditions under which functionally equivalent nonlinear networks can have arbitrarily dissimilar representations. For $\mathsf{PARIS}$, this implies that invariance needs to be checked against the relation at the queried site, not inferred solely from equality of the network's final output. If an encoding identifies two upstream computations whose target relations to a third site differ, no deterministic predictor on that encoding can reproduce both targets. Accordingly, we seek invariance to target-preserving relabelings and admissible internal basis changes, while retaining reparameterizations that alter the selected representation geometry.

\noindent\textbf{Independent embeddings and interaction-based graph comparison.}
Learning to compare graph-structured objects provides a direct
architectural precedent for our two variants. Graph Matching Networks contrast independently computed graph embeddings with pair-conditioned encodings that exchange information through cross-graph attention \citep{li2019graph}.
SimGNN combines graph-level descriptors with fine-grained node comparisons \citep{bai2019simgnn}, while GraphSim compares node embeddings at multiple scales \citep{bai2020learning}.
Graph embeddings have also been used to retrieve related compiled
functions from control-flow graphs \citep{xu2017neural}.
These approaches establish independent encoding and cross-graph interaction as existing strategies for comparing structured objects. Our task instead uses parameterized neural computations as inputs and activation-derived site relations as supervision.

An analogous design space appears in information retrieval.
Sentence-BERT and dense passage retrieval use separately computed query and candidate representations \citep{reimers2019sentence,karpukhin2020dense}.
Poly-encoders study intermediate designs between independent encoding and full pairwise interaction \citep{humeau2019poly}, while ColBERT precomputes candidate representations and retains query-dependent late interaction \citep{khattab2020colbert}.
Thus, reusable candidate features and pair-adaptive scoring are not mutually exclusive. Learned relation modules also provide an alternative to prescribing a fixed distance between embeddings
\citep{sung2018learning}.

$\mathsf{PARIS}$-Ind and $\mathsf{PARIS}$-Joint instantiate two points in this design space. The independent variant computes a reusable descriptor for each upstream-computation graph and applies relation-specific heads.
The joint variant permits interaction between site graphs before producing a score, so its comparison can depend on the particular query-candidate pair. These variants adapt established graph-comparison designs to parameter-based site-relation prediction and allow us to study their trade-offs in this setting. Reusable embeddings avoid repeating candidate encoding across queries. Efficient nearest-neighbor indexing additionally depends on the form of the comparison head. Note that greater pairwise interaction does not by itself guarantee better retrieval: the relevant trade-off is empirical and includes prediction quality, encoding reuse, and per-comparison cost.

\noindent\textbf{Relational knowledge distillation.}
Our setup is also related to relational knowledge distillation \citep{park2019relational}, where a student learns relations from a teacher.
We adopt a related principle here, however, the targets compare internal sites from independently
trained models, while the predictor receives only the parameterized
computations that produce those sites.
The goal is to predict the same comparisons for previously unseen
models.

\noindent\textbf{Broader weight space learning and pretrained-model selection.}
Learning over populations of trained networks has been used to infer model characteristics directly from parameters. Early examples predict accuracy from weights \citep{unterthiner2020predicting} and identify properties of models and their training configurations \citep{eilertsen2020classifying}.
Hyper-representations learn reusable weight space embeddings for
model-characteristic prediction \citep{schurholt2021self}, and SANE extends weight space representation learning through sequential processing of weight subsets, including layer-wise embeddings \citep{schurholt2024towards}. Model-zoo datasets and augmentation strategies provide complementary
resources for learning and generalization across network populations \citep{schurholt2022model,shamsian2024improved}.
Accordingly, neither learning from trained parameters nor producing layer-level descriptors alone distinguishes our setting.

\citet{schurholt2025model} provide model zoos organized by loss-landscape phases, with controlled variation in architecture, size, and data.
This is a complementary resource for the population-level learning setting of $\mathsf{PARIS}$.
A prospective evaluation could use these populations to distinguish within-phase interpolation from cross-phase generalization, and assess whether relation prediction transfers beyond changes of random seed. Such an evaluation needs to separate complete training runs across meta-training and testing, keeping related checkpoints and their sites in the same split. Phase information can support stratified evaluation without becoming an activation-dependent input to the search-time predictor. Using these datasets and performing these stronger generalization tests are possible extensions, not established results of the present study.

Relational precedents require a more specific comparison. Deep-Align learns a symmetry-aware mapping from network pairs to parameter alignments \citep{navon2024equivariant}, where its objective is to recover permutations.
\citet{haber2022learning} learn a metric from synaptic differences that predicts response similarity in novel simulated spiking networks, and identify a sparse set of architectural features that supports this prediction. This is an especially close precedent for learning functional relations from network parameters. 
Yet, their comparisons assume known neuron correspondence across networks. $\mathsf{PARIS}$ compares selected sites whose feature coordinates need not be aligned and whose upstream computations may differ in size and structure.

Pretrained-model selection shares the goal of reducing
candidate evaluation, but typically ranks models for a target task rather than for a query representation. Task2Vec uses task embeddings to support feature-extractor selection \citep{achille2019task2vec}.
LEEP and LogME assess transferability using model predictions or extracted features on target data \citep{nguyen2020leep,you2021logme}, whereas
Model Spider learns model-task compatibility from approximate performance measurements \citep{zhang2023modelspider}.
These methods motivate learned surrogates and efficient search over model collections, but model-task suitability is distinct from a relation between two internal representation sites.

$\mathsf{PARIS}$ brings these directions together by treating
activation-derived site-pair measurements as supervision for a graph metanetwork. The query is a representation site, the candidates are sites in other trained models, and the predicted quantity is the selected representation relation. At inference, the base-model computation is supplied through its architecture, trained parameters, and fixed inference state.
The framework amortizes activation-based relation estimation across a model population, while relation-prediction accuracy and generalization to unseen models remain empirical questions.

\section{Representation (dis)similarity measures}
\label{app:relations} 

This section defines the activation-derived representation relations used in our paper.

\noindent\textbf{Reference observations and notation.} Let $ \mX=(\vx_i)_{i=1}^{n}$  denote the ordered reference sample, where $n=1000$ in our experiments. As defined in App.~\ref{app:representation_sites}, a representation site $u$ produces the feature vector $r_u(\vx_i)\in\mathbb{R}^{p_u}$ for reference input $\vx_i$. The corresponding activation matrix is 
\begin{equation} \mH_u(\mX) = \begin{bmatrix} r_u(\vx_1)^\top\\ \vdots\\ r_u(\vx_n)^\top \end{bmatrix} \in \mathbb{R}^{n\times p_u}. \label{eq:app_relation_activation_matrix} \end{equation} 
Rows correspond to the same reference examples in the same order for every site, whereas the feature dimension $p_u$ may differ across sites. For a relation type $\mathsf h$, the reference target is written as \begin{equation} y_{uv}^{(\mathsf h)} = \delta_{\mathsf h} \left( \mH_u(\mX), \mH_v(\mX) \right), \label{eq:app_reference_relation} \end{equation} where $\delta_{\mathsf h}$ denotes the relation-specific dissimilarity defined below. 
Reference-example identities and their ordering are fixed before computing any relation. We do not match examples according to model predictions or reorder the observations separately for different sites or site pairs. 

\noindent\textbf{Centering.} For relations that require centering over reference examples, define the centering matrix $ \mJ_n = \mI_n - \frac{1}{n} \mathbf{1}_n\mathbf{1}_n^\top$, and the centered activation matrix $ \overline{\mH}_u = \mJ_n\mH_u$.  Left multiplication by $\mJ_n$ subtracts the empirical mean of each feature over the reference examples. It does not center the feature coordinates within an individual example.

\noindent\textbf{Chordal linear-CKA distance.}
Linear centered kernel alignment (CKA) compares the centered inner-product geometry of two representations \citep{kornblith2019similarity,brockmeier2017quantifying}. For brevity, write $\mH_u=\mH_u(\mX)$ for the activation matrix of site $u$. We define its linear Gram matrix and centered Gram matrix as \begin{equation} \mK_u = \mH_u\mH_u^\top, \qquad \overline{\mK}_u = \mJ_n\mK_u\mJ_n = \overline{\mH}_u\overline{\mH}_u^\top, \label{eq:app_linear_gram} \end{equation} where $\overline{\mH}_u=\mJ_n\mH_u$ and $\mJ_n$ is the centering matrix defined above. For sites $u$ and $v$ with nonzero centered Gram matrices, the linear CKA similarity is \begin{equation} s_{\mathrm{CKA}}^{\mathrm{lin}}(u,v) = \frac{ \left\langle \overline{\mK}_u, \overline{\mK}_v \right\rangle_F }{ \|\overline{\mK}_u\|_F \|\overline{\mK}_v\|_F }, \label{eq:reference_linear_cka} \end{equation} where $\langle\mA,\mB\rangle_F =\operatorname{tr}(\mA^\top\mB)$ denotes the Frobenius inner product. Equivalently, the same quantity can be computed directly in feature space as \begin{equation} s_{\mathrm{CKA}}^{\mathrm{lin}}(u,v) = \frac{ \left\| \overline{\mH}_u^\top \overline{\mH}_v \right\|_F^2 }{ \left\| \overline{\mH}_u^\top \overline{\mH}_u \right\|_F \left\| \overline{\mH}_v^\top \overline{\mH}_v \right\|_F }. \label{eq:reference_linear_cka_feature} \end{equation}
To obtain the dissimilarity, consider the normalized centered Gram descriptor \begin{equation} \mB_u = \frac{ \overline{\mK}_u }{ \|\overline{\mK}_u\|_F }. \end{equation} We then use the unit-range chordal linear-CKA distance \begin{equation} \delta_{\mathrm{C}}(u,v) = \frac{1}{\sqrt{2}} \left\| \mB_u-\mB_v \right\|_F = \sqrt{ 1-s_{\mathrm{CKA}}^{\mathrm{lin}}(u,v) }. \label{eq:reference_chordal_cka} \end{equation} Because the normalized centered linear Gram matrices have unit Frobenius norm and $s_{\mathrm{CKA}}^{\mathrm{lin}}(u,v)\in[0,1]$ in exact arithmetic, $\delta_{\mathrm{C}}(u,v)\in[0,1]$. The factor $1/\sqrt{2}$ then converts the usual chordal distance $\sqrt{2-2s_{\mathrm{CKA}}^{\mathrm{lin}}(u,v)}$ to the unit interval. We retain the square root in Eq.~(\ref{eq:reference_chordal_cka}).  Hence, $1-s_{\mathrm{CKA}}^{\mathrm{lin}}(u,v)$ is the squared unit-range chordal distance and is not used as an interchangeable target. No additional normalization or rescaling is applied to $\delta_{\mathrm{C}}$.

\noindent\textbf{HSIC-based CKA dissimilarity.}
CKA normalizes the Hilbert-Schmidt
independence criterion (HSIC) \citep{gretton2005measuring} to compare the kernel-induced similarity structure of matched reference observations
\citep{kornblith2019similarity}. 

For site $u$, the RBF kernel is given by 
\begin{equation}
    \left(\mK_u^{\mathrm{rbf}}\right)_{ij}
    =
    \exp\left(
        -
        \frac{
            \|r_u(\vx_i)-r_u(\vx_j)\|_2^2
        }{
            2\sigma_u^2
        }
    \right),
    \label{eq:hsic_rbf_kernel}
\end{equation}
where $r_u(\vx_i)$ is the extracted representation vector, and $\sigma_u>0$ is the bandwidth. The bandwidth determines the scale at which pairwise distances between reference representations are compared.

Let $\mJ_n=\mI_n-n^{-1}\mathbf{1}_n\mathbf{1}_n^\top$ denote the centering matrix. For symmetric kernel matrices $\mK$ and $\mL$, the biased empirical HSIC estimator is
\begin{equation}
    \mathcal{H}_{\mathrm{b}}(\mK,\mL)
    =
    \frac{1}{(n-1)^2}
    \operatorname{tr}
    \left(
        \mK\mJ_n\mL\mJ_n
    \right).
    \label{eq:hsic_biased}
\end{equation}
Equivalently, $\mathcal{H}_{\mathrm{b}}(\mK,\mL)$ is proportional to the
Frobenius inner product between the centered kernel matrices $\mJ_n\mK\mJ_n$ and $\mJ_n\mL\mJ_n$.
The HSIC-based RBF-CKA similarity is defined as
\begin{equation}
    s_{\mathrm{CKA}}^{\mathrm{rbf},\nu}(u,v)
    =
    \frac{
        \mathcal{H}_{\mathrm{b}}
        \left(
            \mK_u^{\mathrm{rbf}},
            \mK_v^{\mathrm{rbf}}
        \right)
    }{
        \sqrt{
            \mathcal{H}_{\mathrm{b}}
            \left(
                \mK_u^{\mathrm{rbf}},
                \mK_u^{\mathrm{rbf}}
            \right)
            \mathcal{H}_{\mathrm{b}}
            \left(
                \mK_v^{\mathrm{rbf}},
                \mK_v^{\mathrm{rbf}}
            \right)
        }
    }.
    \label{eq:hsic_cka_similarity}
\end{equation}
The supervised dissimilarity is then
$
    \delta_{\mathrm{HSIC,rbf}}(u,v)
    =
    1-s_{\mathrm{CKA}}^{\mathrm{rbf},\nu}(u,v)$.

\noindent\textbf{GULP distance} \label{app:gulp} GULP compares representations through the behavior of regularized linear predictors fitted to their features \citep{boix2022gulp}. 
We first apply a sitewise scalar normalization, \begin{equation} a_u = \left( \frac{1}{n} \|\overline{\mH}_u\|_F^2 \right)^{1/2}, \qquad \mZ_u = \frac{\overline{\mH}_u}{a_u}. \label{eq:gulp_normalization} \end{equation} 
Define the empirical covariance and cross-covariance matrices \begin{equation} \mS_u = \frac{1}{n}\mZ_u^\top\mZ_u, \qquad \mS_{uv} = \frac{1}{n}\mZ_u^\top\mZ_v. \label{eq:gulp_covariances} \end{equation} For regularization parameter $\lambda_{\mathrm G}>0$, let $ \mA_u = \left( \mS_u+\lambda_{\mathrm G}\mI_{p_u} \right)^{-1}$.  The squared GULP distance between sites $u$ and $v$ is then give by \begin{equation}
d_{\mathrm G,\lambda_{\mathrm G}}(u,v)^2 ={}\operatorname{tr} \left( \mA_u\mS_u \mA_u\mS_u \right) + \operatorname{tr} \left( \mA_v\mS_v \mA_v\mS_v \right) - 2\operatorname{tr} \left( \mA_u\mS_{uv} \mA_v\mS_{uv}^{\top} \right).
\label{eq:gulp_covariance}
\end{equation}
We use $\lambda_{\mathrm G}=1$ with covariance matrices normalized by $1/n$ as in Eq.~(\ref{eq:gulp_covariances}). Under the equivalent formulation using the unnormalized matrix $\mZ_u^\top\mZ_u$, the corresponding regularizer is $n\lambda_{\mathrm G}\mI_{p_u}$. Thus, \begin{equation} \mS_u+\lambda_{\mathrm G}\mI_{p_u} = \frac{1}{n} \left( \mZ_u^\top\mZ_u + n\lambda_{\mathrm G}\mI_{p_u} \right). \end{equation}
The same distance can be expressed using the regularized sample-space operator \begin{equation} \mP_u = \frac{1}{n} \mZ_u\mA_u\mZ_u^\top = \mZ_u \left( \mZ_u^\top\mZ_u + n\lambda_{\mathrm G}\mI_{p_u} \right)^{-1} \mZ_u^\top. \label{eq:gulp_sample_operator} \end{equation} With this notation, we have \begin{equation} d_{\mathrm G,\lambda_{\mathrm G}}(u,v) = \|\mP_u-\mP_v\|_F. \label{eq:gulp_sample_distance} \end{equation} 
For completeness, consider a thin singular-value decomposition \begin{equation} \mZ_u = \mU_u\mD_u\mV_u^\top, \end{equation} where the diagonal entries $s_{u,j}$ of $\mD_u$ are the singular values. Then \begin{equation} \mP_u = \mU_u \operatorname{diag} \left( \frac{ s_{u,j}^2 }{ s_{u,j}^2+n\lambda_{\mathrm G} } \right) \mU_u^\top. \label{eq:gulp_svd} \end{equation}

\noindent\textbf{Mutual nearest-neighbor dissimilarity.} \label{app:mutual_nn} MutualNN compares the local neighborhood structure induced by two representations on the same reference examples \citep{huh2024position}. For each site $u$, let $\widetilde r_u(\vx_i)\in\mathbb{R}^{p_u}$ denote the feature vector.

For reference example $i$, let \begin{equation} \mathcal{N}_u^{k_{\mathrm{NN}}}(i) = \operatorname*{arg\,top}_{\substack{ k_{\mathrm{NN}}\\ j\in\{1,\ldots,n\}\setminus\{i\} }} \left\{ -d_{\mathrm{NN}} \left( \widetilde r_u(\vx_i), \widetilde r_u(\vx_j) \right) \right\} \label{eq:mutual_nn_neighborhood} \end{equation} denote the indices of the $k_{\mathrm{NN}}$ nearest reference examples to example $i$ at site $u$. 
For a pair of sites $u$ and $v$, MutualNN measures the fraction of neighbors shared by the two representation spaces for each corresponding reference example. We define \begin{equation} s_{\mathrm{NN}}(u,v) = \frac{1}{n k_{\mathrm{NN}}} \sum_{i=1}^{n} \left| \mathcal{N}_u^{k_{\mathrm{NN}}}(i) \cap \mathcal{N}_v^{k_{\mathrm{NN}}}(i) \right|. \label{eq:mutual_nn_similarity} \end{equation} The intersection is taken over reference-example indices.
The corresponding dissimilarity target is $ \delta_{\mathrm{NN}}(u,v) = 1-s_{\mathrm{NN}}(u,v)$.

\noindent\textbf{Centered kernel nearest-neighbor alignment.} \label{app:cknna} Centered kernel nearest-neighbor alignment (CKNNA) combines kernel-based representation comparison with a restriction to neighborhoods shared between the two representation spaces \citep{huh2024position}. 
For each site, neighborhoods are constructed as in MutualNN. For a pair of sites $u$ and $v$, consider the pair-specific shared-neighborhood mask \begin{equation} (\mM_{uv})_{ij} = \mathbb{I} \left[ j\in \mathcal{N}_u^{k_{\mathrm{CKNNA}}}(i) \cap \mathcal{N}_v^{k_{\mathrm{CKNNA}}}(i) \right]. \label{eq:cknna_shared_mask_definition} \end{equation} Thus, $(\mM_{uv})_{ij}=1$ only when reference example $j$ is among the $k_{\mathrm{CKNNA}}$ nearest neighbors of reference example $i$ in both representation spaces and $(\mM_{uv})_{ii}=0$. 
Let $\mK_u$ and $\mK_v$ denote the kernel matrices associated with the two sites, 
define the masked alignment statistic \begin{equation} \mathcal{A}_{\mathrm{b}} (\mK,\mL;\mM) = \mathcal{H}_{\mathrm{b}} \left( \mK\odot\mM,\, \mL\odot\mM \right), \label{eq:cknna_masked_hsic} \end{equation} where $\odot$ denotes the Hadamard product. 
The same pair-specific mask $\mM_{uv}$ is used in the cross term and in both self-normalization terms. The resulting similarity is then given by  \begin{equation} s_{\mathrm{CKNNA}}(u,v) = \frac{ \mathcal{A}_{\mathrm{b}} (\mK_u,\mK_v;\mM_{uv}) }{ \sqrt{ \mathcal{A}_{\mathrm{b}} (\mK_u,\mK_u;\mM_{uv}) \, \mathcal{A}_{\mathrm{b}} (\mK_v,\mK_v;\mM_{uv}) } }. \label{eq:cknna_shared_mask} \end{equation} 
The CKNNA dissimilarity is then given by $ \delta_{\mathrm{CKNNA}}(u,v) = 1-s_{\mathrm{CKNNA}}(u,v)$. 

\noindent\textbf{Held-out linear-readout disagreement.} \label{app:readout_disagreement} Held-out linear-readout disagreement compares two representation sites through the predictions of linear classifiers fitted independently to their features \citep{alain2016understanding,harvey2024what}.

Let $\mathcal{I}_{\mathrm{fit}}$, $\mathcal{I}_{\mathrm{tune}}$, and $\mathcal{I}_{\mathrm{eval}}$ denote disjoint sets of reference-example indices used for readout fitting, hyperparameter selection, and relation evaluation, respectively. 
Assume a classification task with $C$ classes having the same semantics and class ordering across all models. At site $u$, the linear readout has weight matrix $\mW_u\in\mathbb{R}^{C\times p_u}$ and intercept $\vb_u\in\mathbb{R}^{C}$. Let $\widetilde r_u(\vx)\in\mathbb{R}^{p_u}$ denote the representation.
The predicted class-probability vector is \begin{equation} \vp_u(\vx) = \operatorname{softmax} \left( \mW_u\widetilde r_u(\vx) + \vb_u \right) \in[0,1]^C. \label{eq:readout_probability} \end{equation} Let $c_i\in\{1,\ldots,C\}$ denote the class label of reference example $\vx_i$. For regularization parameter $\lambda_{\mathrm{ro}}>0$, we fit the readout by minimizing 
\begin{equation} 
\mathcal{L}_{\mathrm{ro}} (\mW_u,\vb_u) = -\frac{1}{|\mathcal{I}_{\mathrm{fit}}|} \sum_{i\in\mathcal{I}_{\mathrm{fit}}} \log \left[ \operatorname{softmax} \left( \mW_u\widetilde r_u(\vx_i) + \vb_u \right) \right]_{c_i} + \frac{\lambda_{\mathrm{ro}}}{2} \|\mW_u\|_F^2.
\label{eq:readout_fit} 
\end{equation} The cross-entropy term is averaged over the fitting examples.

After fitting, the hard prediction of the readout at site $u$ is given by \begin{equation} \widehat c_u(\vx) = \operatorname*{arg\,max}_{c\in\{1,\ldots,C\}} [\vp_u(\vx)]_c. \label{eq:readout_hard_prediction} \end{equation}
We define the held-out readout disagreement between sites $u$ and $v$ as \begin{equation} \delta_{\mathrm{ro}}(u,v) = \frac{1}{|\mathcal{I}_{\mathrm{eval}}|} \sum_{i\in\mathcal{I}_{\mathrm{eval}}} \mathbb{I} \left[ \widehat c_u(\vx_i) \neq \widehat c_v(\vx_i) \right]. \label{eq:readout_disagreement} \end{equation} Thus, $\delta_{\mathrm{ro}}(u,v)$ is the fraction of held-out examples on which the two independently fitted readouts predict different classes.

For fixed fitted readouts, let $\mF_u\in\{0,1\}^{|\mathcal{I}_{\mathrm{eval}}|\times C}$ denote the matrix of one-hot hard predictions at site $u$, with row $i$ corresponding to the prediction on the $i$th held-out evaluation example. Then \begin{equation} \delta_{\mathrm{ro}}(u,v) = \frac{ \|\mF_u-\mF_v\|_F^2 }{ 2|\mathcal{I}_{\mathrm{eval}}| }. \label{eq:readout_disagreement_euclidean} \end{equation} Hence, for fixed readouts and a fixed evaluation set, hard-readout disagreement is a squared Euclidean distance.

\section{Theoretical properties}
\label{app:Proof and further theoretical results}

\subsection{Symmetries of the site graphs and predictors}
\label{app:Symmetries of the site graphs and predictors}
We show which changes to a model leave the scores predicted by
$\mathsf{PARIS}$ unchanged.
The proof uses the shared graph updates \citep{lim2024graph} and our
paired attention encoding \citep{theus2025generalized, maurya2026training}.
We also show that swapping the two compared sites does not change the
predicted score.

Let
$G_u=(V_u,E_u,X_u,A_u,T_u)$
denote the graph for site $u$, where $X_u=(\bm x_i)_{i\in V_u}$ contains the node attributes,
$A_u=(\bm a_e)_{e\in E_u}$ contains the edge attributes, and
$T_u\subseteq V_u$ marks the nodes associated with the target site.
The graph may contain parallel edges.
A graph relabeling changes only the node and edge labels.
It moves each attribute, endpoint, and target marker together with the
corresponding node or edge: $    s'(\pi_E(e))=\pi_V(s(e)),
    t'(\pi_E(e))=\pi_V(t(e)),
    \bm x'_{\pi_V(i)}=\bm x_i,
    \bm a'_{\pi_E(e)}=\bm a_e,
    t'_u=\pi_V(t_u)$, where $s(e)$ and $t(e)$ denote the source and target of edge $e$, and
$\pi_V$ and $\pi_E$ are bijections.
The relabeling does not change node or edge types, operand roles, or
spatial attributes.
We write $G\cong G'$ when two graphs differ only by such labels.

We call a hidden-coordinate permutation \emph{compatible} if it
reorders every affected weight axis, bias, and connection in
the same way.
The permutation is required to also respect residual branches, normalization
groups, shared parameters, and the rule used to extract the target
representation.
Coordinates with a fixed external meaning remain fixed.
Permuting complete attention heads moves each head's QK and
VO operators with its dimensions, scale, and mask.
The graph does not encode arbitrary coordinate or head indices.

For attention head $a$, we use row-vector notation and denote
\begin{equation}
 \mW_{Q,a},\mW_{K,a}\in\mathbb R^{D\times d_{q,a}},\qquad
 \mW_{V,a}\in\mathbb R^{D\times d_{v,a}},\qquad
 \mW_{O,a}\in\mathbb R^{d_{v,a}\times D}.
\end{equation}
The biases $\vb_{Q,a}$, $\vb_{K,a}$, and $\vb_{V,a}$ are row vectors, and
$\vb_O\in\mathbb R^{1\times D}$.
The graph encodes
\begin{equation}
\begin{aligned}
 \mC_a^{\mathrm{QK}}&=\mW_{Q,a}\mW_{K,a}^{\top},
 & \mC_a^{\mathrm{VO}}&=\mW_{V,a}\mW_{O,a},\\
 \bar \mC_a^{\mathrm{QK}}
 &=\alpha_a
 \begin{bmatrix}
     \mC_a^{\mathrm{QK}}\\ \vb_{Q,a}\mW_{K,a}^{\top}
 \end{bmatrix},
 & \vb_{\mathrm{eff}}&=\vb_O+\sum_a \vb_{V,a}\mW_{O,a},
\end{aligned}
\label{eq:paris_invariance_encoded_circuits}
\end{equation}
where $\alpha_a$ denotes the logit scale.
The graph keeps each QK/VO pair in the same head subgraph and records
fixed settings such as its scale and mask.
We use this encoding for ordinary affine scaled dot-product attention
and for target sites at or after the combined projected attention
output.

\begin{proposition}
 
\label{prop:paris_invariance}
 
Let $\widehat{\bm y}_{\mathsf A}(G_u,G_v)$ denote the score vector
predicted by
$\mathsf A\in\{\mathrm{Ind},\mathrm{Joint}\}$.
For fixed metanetwork parameters, the following properties hold.

\begin{enumerate}[leftmargin=1.75em,itemsep=2pt,topsep=0pt, itemsep=0pt, parsep=0pt, partopsep=0pt]
\renewcommand{\labelenumi}{(\roman{enumi})}

\item \textbf{Graph relabeling.}
Relabeling either site graph leaves the independent site embedding and
the predicted scores unchanged.
For any relabelings $\pi_u$ and $\pi_v$,
\begin{equation}
    \operatorname{Enc}_{\phi}(\pi_uG_u)
    =\operatorname{Enc}_{\phi}(G_u),
    \label{eq:paris_theory_encoder_invariance}
\end{equation}
and
\begin{equation}
    \widehat{\bm y}_{\mathsf A}(\pi_uG_u,\pi_vG_v)
    =\widehat{\bm y}_{\mathsf A}(G_u,G_v).
    \label{eq:paris_theory_joint_invariance}
\end{equation}

\item \textbf{Changes of basis within attention heads.}
For invertible matrices
$\mA_a\in\mathbb R^{d_{q,a}\times d_{q,a}}$ and
$\mB_a\in\mathbb R^{d_{v,a}\times d_{v,a}}$, let
\begin{equation}
\begin{aligned}
 \mW'_{Q,a}&=\mW_{Q,a}\mA_a,
 & \mW'_{K,a}&=\mW_{K,a}\mA_a^{-\top},\\
 \mW'_{V,a}&=\mW_{V,a}\mB_a,
 & \mW'_{O,a}&=\mB_a^{-1}\mW_{O,a},\\
 \vb'_{Q,a}&=\vb_{Q,a}\mA_a,
 & \vb'_{K,a}&=\vb_{K,a}\mA_a^{-\top},\\
 \vb'_{V,a}&=\vb_{V,a}\mB_a,
 & \vb'_O&=\vb_O.
\end{aligned}
\label{eq:paris_supported_basis_action}
\end{equation}
Keep all other parameters, settings, and the target extraction rule
fixed.
These transformations leave the encoded attention operators unchanged.
Applying them to either site then leaves the independent embedding
and the scores from both estimators unchanged.

\item \textbf{Swapping the sites.}
Both estimators are symmetric:
\begin{equation}
    \widehat{\bm y}_{\mathsf A}(G_u,G_v)
    =\widehat{\bm y}_{\mathsf A}(G_v,G_u).
    \label{eq:paris_pair_exchange_invariance}
\end{equation}
\end{enumerate}

\end{proposition}

\begin{proof}
\textbf{Graph relabeling.}
Let $G'=\pi G$.
Because the shared node encoder receives the same attributes at
corresponding nodes,
$\bm h_{\pi_V(i)}^{\prime(0)}=\bm h_i^{(0)}$.
Assume that this equality holds after message-passing round $r$.
For an edge $e$, the message rule gives
\begin{equation*}
\begin{aligned}
 \bm m_{\pi_E(e)}^{\prime(r)}
 &=
 \left(\mW^{(r)}\bm h_{\pi_V(s(e))}^{\prime(r)}\right)
 \odot\psi_{\mathrm{edge},r}(\bm a'_{\pi_E(e)})\\
 &=
 \left(\mW^{(r)}\bm h_{s(e)}^{(r)}\right)
 \odot\psi_{\mathrm{edge},r}(\bm a_e)
 =
 \bm m_e^{(r)}.
\end{aligned}
\end{equation*}
Every incoming edge of node $j$ has a corresponding edge at
$\pi_V(j)$, including parallel and reverse edges.
The same holds when a node has no incoming edges.
Because the two graphs use the same update MLP, residual step, and
nodewise LayerNorm, we obtain
$\bm h_{\pi_V(j)}^{\prime(r+1)}=\bm h_j^{(r+1)}$.
Induction gives then this equality after every message-passing round.

Relabeling also leaves the two averages used by
$\mathsf{PARIS}$-Ind unchanged:
\begin{equation*}
\begin{aligned}
 \bm p'_u
 &=
 \frac{1}{|\pi_V(t_u)|}
 \sum_{i'\in\pi_V(t_u)}\bm h_{i'}^{\prime(R)}
 =
 \frac{1}{|t_u|}
 \sum_{i\in t_u}\bm h_i^{(R)}
 =
 \bm p_u,\\
 \bm g'_u
 &=
 \frac{1}{|V'_u|}
 \sum_{i'\in V'_u}\bm h_{i'}^{\prime(R)}
 =
 \frac{1}{|V_u|}
 \sum_{i\in V_u}\bm h_i^{(R)}
 =
 \bm g_u.
\end{aligned}
\label{eq:paris_pool_relabeling_proof}
\end{equation*}
Thus, the readout
$\bm z_u=\rho_{\phi}([\bm p_u\Vert\bm g_u])$
receives the same input and returns the same site embedding.
Each comparison head in $\mathsf{PARIS}$-Ind returns the same
score.

For $\mathsf{PARIS}$-Joint, relabeling only reorders the operator and
head tokens.
It leaves the summary tokens and token roles unchanged.
For one cross-attention head, we have
\begin{equation*}
 \operatorname{CA}_{\mM}(\mU,\mV)
 =
 \operatorname{softmax}
 \left(
 \frac{\mathcal Q(\mU)\mathcal K(\mV)^\top}{\sqrt{d_a}}+\mM
 \right)
 \mathcal V(\mV),
\label{eq:paris_crossattention_for_proof}
\end{equation*}
where $\mathcal Q$, $\mathcal K$, and $\mathcal V$ are affine maps
applied to each row.
These maps commute with row permutations.
For permutation matrices $\mP_1,\mP_2$ and masked logits $\mL$,
\begin{equation*}
 \operatorname{softmax}(\mP_1\mL\mP_2^\top)
 =
 \mP_1\operatorname{softmax}(\mL)\mP_2^\top.
\end{equation*}
For masks containing $-\infty$, a permutation only reindexes the
entries.
Using $\mP_2^\top \mP_2=\mI$ gives
\begin{equation*}
 \operatorname{CA}_{\mP_1\mM\mP_2^\top}(\mP_1\mU,\mP_2\mV)
 =
 \mP_1\operatorname{CA}_{M}(\mU,\mV).
 \label{eq:paris_attention_permutation_proof}
\end{equation*}
Thus, permuting the query and key/value tokens only permutes the output
query tokens in the same way.
The output projection, residual update, feed-forward network, and
LayerNorm preserve this property because they apply the same function
to corresponding rows.
Masks remove padded keys and pooling ignores padded tokens.
The target-summary token keeps its role, and averaging valid tokens is
independent of their order.
The two contextual readouts then return the same vectors, so the
joint estimator also returns the same score.

We now connect graph relabeling to hidden-coordinate permutations.
For this example only, use column-vector notation.
For two linear maps separated by a coordinatewise nonlinearity, a
permutation $\mP$ of the hidden coordinates gives
\begin{equation*}
 (\mW_1,\vb_1,\mW_2)
 \longmapsto
 (\mP\mW_1,\mP\vb_1,\mW_2\mP^\top).
\end{equation*}
Thus, the corresponding rows of $\mW_1$, entries of $\vb_1$, and columns
of $\mW_2$ move together.
This operation only relabels the intermediate coordinate nodes.
The same consistency holds across residual branches, normalization
groups, and the target extraction rule.

The attention encoding behaves in the same way under coordinate
permutations.
If $\mP$ and $\mR$ permute the input and output residual coordinates, then we have 
\begin{equation*}
 \mC_a^{\mathrm{QK}\prime}
 =
 \mP \mC_a^{\mathrm{QK}}\mP^\top,
 \qquad
 \mC_a^{\mathrm{VO}\prime}
 =
 \mP \mC_a^{\mathrm{VO}}\mR^\top.
\end{equation*}
The query-bias term becomes
$\vb_{Q,a}\mW_{K,a}^\top \mP^\top$, and the effective output bias becomes
$\vb_{\mathrm{eff}}\mR^\top$.
Permuting complete attention heads also moves each paired head
subgraph as one unit.
All these transformations reduce to graph relabelings, so
the result above applies.

\medskip
\textbf{Changes of basis within attention heads.}
We omit the head index.
Substituting the transformed parameters from
Eq.~(\ref{eq:paris_supported_basis_action}) gives
\begin{equation*}
\begin{aligned}
 \mW'_Q(\mW'_K)^\top
 &=\mW_Q\mA\mA^{-1}\mW_K^\top=\mC^{\mathrm{QK}},\\
 \mW'_V\mW'_O
 &=\mW_V\mB\mB^{-1}\mW_O=\mC^{\mathrm{VO}},\\
 \vb'_Q(\mW'_K)^\top
 &=\vb_Q\mA\mA^{-1}\mW_K^\top=\vb_Q\mW_K^\top,\\
 \vb'_V\mW'_O
 &=\vb_V\mB\mB^{-1}W_O=\vb_V\mW_O.
\end{aligned}
\end{equation*}
Hence
$\bar \mC_a^{\mathrm{QK}}$,
$\mC_a^{\mathrm{VO}}$, and
$\vb_{\mathrm{eff}}$
remain unchanged.
The head pairing, dimensions, scales, masks, and structural edges also
remain fixed.
Because the graph does not contain the raw projection factors or any
basis-dependent statistics, the encoded attention subgraph does not
change.

These encoded quantities also determine the combined attention output.
For $\mX\in\mathbb R^{T\times D}$, write
\begin{equation*}
 \mQ=\mX\mW_Q+\bm 1\vb_Q,\qquad
 \mK=\mX\mW_K+\bm 1\vb_K,\qquad
 \mV=\mX\mW_V+\bm 1\vb_V.
\end{equation*}
Expanding the logits gives
\begin{equation*}
 \mQ\mK^\top
 =
 \mX\mC^{\mathrm{QK}}\mX^\top
 +\bm 1(\vb_Q\mW_K^\top)\mX^\top
 +
 \left(
 \mX\mW_\mQ\vb_K^\top+\bm 1\vb_Q\vb_K^\top
 \right)\bm 1^\top.
\label{eq:paris_affine_logit_expansion}
\end{equation*}
The final term is constant across each query row, so row-wise softmax
removes it, including in the presence of a fixed mask.
Each row of the resulting attention matrix $P_a(\mX)$ sums to one because
every query has at least one allowed key.
Thus, $P_a(\mX)\bm 1=\bm 1$.
The supported changes of basis leave this tensor unchanged for every
input $\mX$.

\medskip
\textbf{Swapping the sites.}
For $\mathsf{PARIS}$-Ind, each comparison head computes
\begin{equation*}
 \widehat y_{\mathrm{Ind}}^{(h)}(G_u,G_v)
 =
 q_h\left(
   \|\mW_h(\bm z_u-\bm z_v)\|_2
 \right).
\end{equation*}
Swapping $u$ and $v$ negates the projected difference but leaves its
Euclidean norm unchanged.
The calibration then returns the same score.

For $\mathsf{PARIS}$-Joint, let $F_\ell(\mU,\mV)$ denote the full update of
the first stream in cross-attention block $\ell$.
Both directions use the same function and read the states from the
start of the block:
\begin{equation*}
\begin{aligned}
 \mT_u^{(\ell+1)}
 &=F_\ell(\mT_u^{(\ell)},\mT_v^{(\ell)}),\\
 \mT_v^{(\ell+1)}
 &=F_\ell(\mT_v^{(\ell)},\mT_u^{(\ell)}).
\end{aligned}
\end{equation*}
Swapping the input graphs swaps the initial streams because both sides
use the same encoder and tokenization.
The masks move with their streams.
By induction
that every cross-attention block swaps its two outputs.
The shared readout  swaps
$\bm c_{u\mid v}$ and $\bm c_{v\mid u}$.
The final comparison head uses
\begin{equation*}
 [
 \bm c_{u\mid v}+\bm c_{v\mid u}
 \;\Vert\;
 |\bm c_{u\mid v}-\bm c_{v\mid u}|
 \;\Vert\;
 \bm c_{u\mid v}\odot\bm c_{v\mid u}
 ],
\end{equation*}
which is unchanged when the two contextual vectors are swapped.
The final MLP and sigmoid return the same score.
This argument requires the two cross-attention directions to share
their parameters, but it does not require the query and key
projections within a direction to be equal.

\end{proof}

\subsection{Geometry of independent and joint scores}
\label{app:Geometry of independent and joint scores}

For a fixed common evaluation sample, the relations considered below admit sitewise Euclidean or squared-Euclidean representations. These constructions motivate embeddings that are computed without reference to the comparison partner. Our graph metanetwork aims to learn embeddings with corresponding pairwise geometry from the site-specific computation graphs.

\begin{proposition}
Given a common sample of $n$ inputs and let
$\mC_n=\mI_n-\mathbf{1}\mathbf{1}^{\top}/n$.
Let $\mK_u=\mC_n\mH_u\mH_u^{\top}\mC_n$ and assume $\mK_u\ne0$ for every site under consideration.
For the empirical centered linear CKA defined by normalized Frobenius inner products of centered Gram matrices,
set
$
 \widetilde{\mK}_u=\frac{\mK_u}{\|\mK_u\|_F}$ and $\delta_{\rm C}(u,v)=
 \sqrt{1-\langle\widetilde{\mK}_u,\widetilde{\mK}_v\rangle_F}$.
Then, we have $
 \delta_{\rm C}(u,v)=\|\Phi_{\rm C}(u)-\Phi_{\rm C}(v)\|_2$ and  $
 \Phi_{\rm C}(u)=2^{-1/2}\operatorname{vec}(\widetilde{\mK}_u)$.
In addition, let $\overline{\mH}_u$ denote the activation matrix obtained using
the fixed sitewise centering and scaling convention used to define the GULP targets.
Define
$
 \mA_u=\frac{\overline{\mH}_u}{\sqrt n}$, 
 $\mS_u=\mA_u^{\top}\mA_u$,  $\mS_{uv}=\mA_u^{\top}\mA_v$, and 
$\mR_u=(\mS_u+\lambda \mI)^{-1}$ for $\lambda>0$.
For the covariance form of GULP, we have 
$
 \delta_{\rm G,\lambda}(u,v)^2
 =\operatorname{tr}(\mR_u\mS_u\mR_u\mS_u)
  +\operatorname{tr}(\mR_v\mS_v\mR_v\mS_v)
  -2\operatorname{tr}(\mR_u\mS_{uv}\mR_v\mS_{uv}^{\top})$,
and
$
 \delta_{\rm G,\lambda}(u,v)
 =\|\Phi_{\rm G,\lambda}(u)-\Phi_{\rm G,\lambda}(v)\|_2$, where
$
 \Phi_{\rm G,\lambda}(u)=\operatorname{vec}(\mA_u\mR_u\mA_u^{\top})$.
 
\end{proposition}

\begin{proof}
We treat the CKA and GULP cases separately.

First, for CKA, observe that $
    \mK_u
    =
    \mC_n\mH_u\mH_u^{\top}\mC_n
    =
    (\mC_n\mH_u)(\mC_n\mH_u)^{\top}$,
because $\mC_n$ is symmetric. Thus, $\mK_u$ is symmetric positive semidefinite. Since $\mK_u\neq 0$, its Frobenius normalization is
well defined and satisfies $\|\widetilde{\mK}_u\|_F=1$.
The same statements hold for the site $v$.
Moreover, since both normalized Gram matrices are positive semidefinite, we have $
    \left\langle
        \widetilde{\mK}_u,
        \widetilde{\mK}_v
    \right\rangle_F
    =
    \operatorname{tr}
    \left(
        \widetilde{\mK}_u
        \widetilde{\mK}_v
    \right)
    \ge 0$.
Note that 
$
    \operatorname{tr}
    \left(
        \widetilde{\mK}_u
        \widetilde{\mK}_v
    \right)
    =
    \operatorname{tr}
    \left(
        \widetilde{\mK}_u^{1/2}
        \widetilde{\mK}_v
        \widetilde{\mK}_u^{1/2}
    \right)
    \ge 0,
$
because the matrix inside the last trace is positive semidefinite.
In addition, by the Frobenius Cauchy-Schwarz inequality, we have 
$
    \left\langle
        \widetilde{\mK}_u,
        \widetilde{\mK}_v
    \right\rangle_F
    \le
    \|\widetilde{\mK}_u\|_F
    \|\widetilde{\mK}_v\|_F
    =1$.
Thus, the normalized Frobenius inner product lies in $[0,1]$, so $\delta_{\rm C}$ is well defined.
We now expand the squared Frobenius distance between the two normalized Gram matrices:
$$
\begin{aligned}
    \frac12
    \|
        \widetilde{\mK}_u
        -
        \widetilde{\mK}_v
    \|_F^2
    &=
    \frac12
    \left(
        \|\widetilde{\mK}_u\|_F^2
        +
        \|\widetilde{\mK}_v\|_F^2
        -
        2
        \left\langle
            \widetilde{\mK}_u,
            \widetilde{\mK}_v
        \right\rangle_F
    \right)
    \\
    &=
    1-
    \left\langle
        \widetilde{\mK}_u,
        \widetilde{\mK}_v
    \right\rangle_F
    \\
    &=
    \delta_{\rm C}(u,v)^2.
\end{aligned}
$$
Therefore, we obtain
$$
    \delta_{\rm C}(u,v)
    =
    \frac{1}{\sqrt2}
    \|
        \widetilde{\mK}_u
        -
        \widetilde{\mK}_v
    \|_F.
$$

Finally, vectorization is an isometry between the Frobenius norm on
matrices and the Euclidean norm on their vectorizations, i.e.,  $
    \|\operatorname{vec}(\mX)\|_2=\|\mX\|_F$.
Thus, we have 
$$
    \delta_{\rm C}(u,v)
    =
    \left\|
        \frac{1}{\sqrt2}
        \operatorname{vec}(\widetilde{\mK}_u)
        -
        \frac{1}{\sqrt2}
        \operatorname{vec}(\widetilde{\mK}_v)
    \right\|_2 
    =
    \|\Phi_{\rm C}(u)-\Phi_{\rm C}(v)\|_2.
$$

We next show that the covariance expression for GULP is the squared Frobenius distance between two $n\times n$ matrices.
Since $
    \mS_u=\mA_u^{\top}\mA_u
$
is symmetric positive semidefinite and $\lambda>0$, we have, for
every nonzero $\vx\in\mathbb{R}^{p_u}$ that 
$
    \vx^{\top}(\mS_u+\lambda\mI_{p_u})\vx
    =
    \|\mA_u \vx\|_2^2
    +
    \lambda\|\vx\|_2^2
    >0.
$
Thus, $\mS_u+\lambda\mI_{p_u}$ is positive definite and invertible. In addition, $\mR_u$ is well defined and symmetric. 
Next, define the sample-space matrix
$
    \mT_u
    :=
    \mA_u\mR_u\mA_u^{\top}
    \in\mathbb{R}^{n\times n}.
$
Since $\mR_u$ is symmetric, we have 
$
    \mT_u^{\top}
    =
    \mA_u\mR_u^{\top}\mA_u^{\top}
    =
    \mT_u,
$
and thus $\mT_u$ is symmetric. Note that since $\mT_u$ is always an
$n\times n$ matrix, irrespective of the representation dimension $p_u$. Thus $\mT_u$ and $\mT_v$ belong to the same sample space
even when $p_u\neq p_v$.

We then compute the squared Frobenius norm of $\mT_u$. Since $\mT_u$ is symmetric,
we have
$$
    \|\mT_u\|_F^2
    =
    \operatorname{tr}(\mT_u^{\top}\mT_u)
     =
    \operatorname{tr}(\mT_u^2)
    =
    \operatorname{tr}
    \left(
        \mA_u\mR_u\mA_u^{\top}
        \mA_u\mR_u\mA_u^{\top}
    \right).
$$
Using cyclicity of the trace and
$\mA_u^{\top}\mA_u=\mS_u$, this becomes
$$
    \|\mT_u\|_F^2
    =
    \operatorname{tr}
    \left(
        \mR_u
        \mA_u^{\top}\mA_u
        \mR_u
        \mA_u^{\top}\mA_u
    \right)
    =
    \operatorname{tr}
    (\mR_u\mS_u\mR_u\mS_u).
$$

Similarly, we have 
$    \|\mT_v\|_F^2
    =
    \operatorname{tr}
    (\mR_v\mS_v\mR_v\mS_v).
$
For the cross term, by using symmetry and
cyclicity of the trace, we have 
$$
\begin{aligned}
    \langle \mT_u,\mT_v\rangle_F
    &=
    \operatorname{tr}(\mT_u^{\top}\mT_v)
     =
    \operatorname{tr}(\mT_u\mT_v)
    \\
    &=
    \operatorname{tr}
    \left(
        \mA_u\mR_u\mA_u^{\top}
        \mA_v\mR_v\mA_v^{\top}
    \right)
    \\
    &=
    \operatorname{tr}
    \left(
        \mR_u
        \mA_u^{\top}\mA_v
        \mR_v
        \mA_v^{\top}\mA_u
    \right).
\end{aligned}
$$
Since
$
    \mA_u^{\top}\mA_v=\mS_{uv}$ and $
    \mA_v^{\top}\mA_u=\mS_{uv}^{\top}$, we have 
$
    \langle \mT_u,\mT_v\rangle_F
    =
    \operatorname{tr}
    \left(
        \mR_u\mS_{uv}
        \mR_v\mS_{uv}^{\top}
    \right)$.
Thus, we can expand the squared Frobenius distance as
$$
\begin{aligned}
    \|\mT_u-\mT_v\|_F^2
    &=
    \|\mT_u\|_F^2
    +
    \|\mT_v\|_F^2
    -
    2\langle\mT_u,\mT_v\rangle_F
    \\
    &=
    \operatorname{tr}
    (\mR_u\mS_u\mR_u\mS_u)
    +
    \operatorname{tr}
    (\mR_v\mS_v\mR_v\mS_v)
    -
    2\operatorname{tr}
    \left(
        \mR_u\mS_{uv}
        \mR_v\mS_{uv}^{\top}
    \right)
    \\
    &=
    \delta_{\rm G,\lambda}(u,v)^2.
\end{aligned}
$$

Both sides are nonnegative, by taking square roots gives
$
    \delta_{\rm G,\lambda}(u,v)
    =
    \|\mT_u-\mT_v\|_F$.
Applying the isometry of vectorization, we have 
$$
    \delta_{\rm G,\lambda}(u,v)
    =
    \|
        \operatorname{vec}(\mT_u)
        -
        \operatorname{vec}(\mT_v)
    \|_2
=
    \|\Phi_{\rm G,\lambda}(u)
      -\Phi_{\rm G,\lambda}(v)\|_2,
$$
where
$
    \Phi_{\rm G,\lambda}(u)
    =
    \operatorname{vec}
    (\mA_u\mR_u\mA_u^{\top}).
$
\end{proof}

\begin{proposition}
Let $N\ge2$ and $k\ge1$. Let
$
    \mY\in\R^{N\times N}
$
be symmetric, entrywise nonnegative, and zero on the diagonal. Let
$
    q:[0,\infty)\to[0,\infty)
$
be strictly increasing with $q(0)=0$, and assume that
$
    \mY_{ij}\in q([0,\infty))$ for all $i,j.$
Since $q$ is strictly increasing, its inverse is well defined on its
range. Define
$
    \mD_{ij}=q^{-1}(\mY_{ij})$,
let
$
    \mJ
    =
    \mI_N-\frac1N\mathbf1\mathbf1^\top,
$
and define the double-centered matrix
$
    \mB_{\rm D}
    =
    -\frac12 \mJ \mD^{\circ2}\mJ,
$
where $\mD^{\circ2}$ denotes the entrywise square of $\mD$.
Then, there exist vectors
$
    \vv_1,\ldots,\vv_N\in\R^k
$
such that
$
    \mY_{ij}
    =
    q\left(\|\vv_i-\vv_j\|_2\right)$ for all  $i,j$
if and only if
$
    \mB_{\rm D}\succeq0$ and  $
    \operatorname{rank}(\mB_{\rm D})\le k$.
That is, the calibrated target matrix $\mY$ is realizable by Euclidean site vectors in dimension at most $k$ precisely when the
inverse-calibrated matrix $\mD=q^{-1}(\mY)$ is a Euclidean distance matrix of embedding dimension at most $k$.
\end{proposition}

\begin{proof}
Because $q$ is strictly increasing and $q(0)=0$, the inverse
$q^{-1}$ is well defined on $q([0,\infty))$. Note that 
$\mD$ is symmetric and entrywise nonnegative, thus we have 
$
    \mD_{ii}
    =
    q^{-1}(\mY_{ii})
    =
    q^{-1}(0)
    =
    0.
$

Suppose there exist $\vv_1,\ldots,\vv_N\in\R^k$ such that
$
    \mY_{ij}
    =
    q\left(\|\vv_i-\vv_j\|_2\right)
$
for every $i,j$. Applying $q^{-1}$ gives
$
   \mD_{ij}
    =
    \|\vv_i-\vv_j\|_2.
$
Let $
    \bar{\vv}=\frac1N\sum_{i=1}^N \vv_i$,
    $\widetilde{\vv}_i=\vv_i-\bar{\vv}$,
and let
$
    \mV
    =
    \begin{pmatrix}
        \widetilde{\vv}_1^\top\\
        \vdots\\
        \widetilde{\vv}_N^\top
    \end{pmatrix}
    \in\R^{N\times k}.
$
Since translation does not change pairwise distances, we have 
$
    \mD_{ij}
    =
    \|\widetilde{\vv}_i-\widetilde{\vv}_j\|_2,
$
in addition,
$
    \mathbf1^\top \mV=0.
$
Now set
$
    \mG=\mV\mV^\top,
$
we then have 
$
    \mD_{ij}^2
    =
    \mG_{ii}+\mG_{jj}-2\mG_{ij}$.
Denote $\vg=\operatorname{diag}(\mG)$ and denote in matrix form as
$
    \mD^{\circ2}
    =
    \vg\mathbf1^\top
    +
    \mathbf1\vg^\top
    -
    2\mG.
$
Since $\mJ\mathbf1=0$, double centering \citep{schoenberg1935remarks} eliminates the first two terms, i.e., 
$
    \mJ\mD^{\circ2}\mJ
    =
    -2\mJ\mG\mJ.
$
Since $\mV$ is centered, we have 
$
    \mJ\mV=\mV,
$
and
$
    \mJ\mG\mJ
    =
    \mJ\mV\mV^\top J
    =
    \mV\mV^\top
    =
    \mG.
$
Thus, we can obtain
$
    \mB_{\rm D}
    =
    -\frac12\mJ\mD^{\circ2}\mJ
    =
    \mV\mV^\top$, 
$
    \mB_{\rm D}\succeq0
$
and
$
    \operatorname{rank}(\mB_{\rm D})
    =
    \operatorname{rank}(\mV\mV^\top)
    \le
    \operatorname{rank}(\mV)
    \le k.
$

For another direction, now suppose
$
    \mB_{\rm D}\succeq0$ and $
    \operatorname{rank}(\mB_{\rm D})\le k$.
By positive semidefiniteness and the rank bound, there exists
$
    \mV\in\R^{N\times k}
$
such that
$
    \mB_{\rm D}=\mV\mV^\top,
$
where zero columns may be added if
$\operatorname{rank}(\mB_{\rm D})<k$.
Since
$
    \mB_{\rm D}
    =
    -\frac12\mJ\mD^{\circ2}\mJ
$
and $\mJ\mathbf1=0$, we have
$
    \mB_{\rm D}\mathbf1=0.
$
It also follows that
$
    0=
    \mathbf1^\top \mB_{\rm D}\mathbf1
    =
    \mathbf1^\top \mV\mV^\top\mathbf1
    =
    \|\mV^\top\mathbf1\|_2^2.
$
Therefore, we have 
$
    \mV^\top\mathbf1=0,
$ and the rows of $\mV$ are centered.
Let $\vv_1,\ldots,\vv_N$ denote the rows of $\mV$, and let $\mS'$ be their squared-distance matrix in the form of 
$
    \mS'_{ij}
    =
    \|\vv_i-\vv_j\|_2^2.
$
Let 
$
    \mS=\mD^{\circ2}.
$
Since $\mV\mV^\top=\mB_{\rm D}$ and the rows of $\mV$ are centered, by double-centering, we have 
$
    \mJ\mS'\mJ=-2\mB_{\rm D}.
$
By the definition of $\mB_{\rm D}$, we have 
$
    \mJ\mS\mJ=-2\mB_{\rm D}.
$, and thus 
$
    \mJ\mS'\mJ=\mJ\mS\mJ.
$
If
$
    \mA:=\mS-\mS',
$
then
$
    \mJ\mA\mJ=0.
$
Moreover, $A$ is symmetric,
$
    \mS_{ii}=\mD_{ii}^2=0$, and  $
    \mS'_{ii}=\|\vv_i-\vv_i\|_2^2=0$.
Next, let
$
    \mP=\frac1N\mathbf1\mathbf1^\top$ and 
$
    \mJ=\mI_N-\mP.
$
Since $\mJ\mA\mJ=0$, we have 
$$
\begin{aligned}
    \mA
    &=
    (\mP+\mJ)\mA(\mP+\mJ)
    \\
    &=
    \mP\mA\mP+\mP\mA\mJ+\mJ\mA\mP
    \\
    &=
    \mP\mA+\mA\mP-\mP\mA\mP.
\end{aligned}
$$
Next we define
$
    \vr=\frac1N \mA\mathbf1$ and $
    \alpha
    =
    \frac1{N^2}\mathbf1^\top \mA\mathbf1.
$
Using the symmetry of $\mA$, we have 
$
    \mA\mP=\vr\mathbf1^\top$, $
    \mP\mA=\mathbf1\vr^\top$, and 
$
    \mP\mA\mP=\alpha\mathbf1\mathbf1^\top.
$
Thus, we obtain
$
    \mA
    =
    \mathbf1\vr^\top
    +
    \vr\mathbf1^\top
    -
    \alpha\mathbf1\mathbf1^\top.
$
By setting
$
    \va=\vr-\frac{\alpha}{2}\mathbf1,
$
we obtain
$
    \mA
    =
    \mathbf1\va^\top+\va\mathbf1^\top.
$
Its diagonal entries satisfy
$
    \mA_{ii}=2\va_i.
$
But $\mA_{ii}=0$ for every $i$, thus 
    $\va_i=0$ for every $i$,
and 
$
    \mA=0.
$
Thus, we have 
$
    \mS=\mS',
$
that is,
$
    \mD_{ij}^2
    =
    \|\vv_i-\vv_j\|_2^2$ for all $i,j$.
Both sides are nonnegative, and $\mD_{ij}\ge0$, we obtain
$
    \mD_{ij}
    =
    \|\vv_i-\vv_j\|_2.
$
Finally, applying $q$ to both sides, we get
$
    \mY_{ij}
    =
    q(\mD_{ij})
    =
    q\left(\|\vv_i-\vv_j\|_2\right)
$
for all $i,j$.
\end{proof}

\begin{corollary}
There exist three valid centered activation matrices $\mH_1,\mH_2,\mH_3$ whose pairwise chordal-CKA distances cannot be represented in the form
$
    1-\exp\left(-\varepsilon\|\vv_i-\vv_j\|_2\right)$, for $i,j\in\{1,2,3\},
$
for any $\varepsilon>0$, any finite embedding dimension $k\ge1$, and any vectors $\vv_1,\vv_2,\vv_3\in\R^k$.
Specifically, let
$
    \mY_{ij}
    =
    \delta_{\rm C}(\mH_i,\mH_j)
$
denote their chordal-CKA distances, and define
\begin{equation}
    E_\star
    :=
    \inf_{\substack{\varepsilon>0,\;k\ge1\\
                    \vv_1,\vv_2,\vv_3\in\R^k}}
    \max_{1\le i<j\le3}
    \left|
        1-\exp\left(-\varepsilon\|\vv_i-\vv_j\|_2\right)
        -
        \mY_{ij}
    \right|.
\end{equation}
Then 
$
    E_\star
    =
    E_0
    :=
    1-\frac12\sqrt{7-2\sqrt3}
    >0.
$
The three chordal-CKA distances admit no exact
exponentially calibrated Euclidean realization, regardless of the embedding dimension.
\end{corollary}

\begin{proof}
We first construct the three centered activation matrices and compute
their chordal-CKA distances.
Choose orthonormal vectors
$\va,\vb\in\R^3$ that are orthogonal to $\mathbf1$. For example,
$
    \va=\frac{1}{\sqrt2}(1,-1,0)^\top$ and $
    \vb=\frac{1}{\sqrt6}(1,1,-2)^\top$.
For
$
    \theta\in
    \left\{
        0,\frac{\pi}{6},\frac{\pi}{3}
    \right\},
$
define the single-column activation matrix as
$
    \mH_\theta
    =
    \cos\theta\,\va+\sin\theta\,\vb
    \in\R^{3\times1}.
$
Since $\va$ and $\vb$ are orthogonal to $\mathbf1$, we have 
$
    \mathbf1^\top \mH_\theta=0,
$
and thus $\mH_\theta$ is centered. Moreover, since $\va$ and $\vb$ are orthonormal, we have 
$
    \|\mH_\theta\|_2=1,
$
and the centered Gram matrix is
$
    \mK_\theta=\mH_\theta \mH_\theta^\top,
$
and
$
    \|\mK_\theta\|_F
    =
    \|\mH_\theta\|_2^2
    =
    1.
$
For two angles $\theta,\theta'$, we consider 
$$
    \left\langle
        \mK_\theta,\mK_{\theta'}
    \right\rangle_F
    =
    \operatorname{tr}
    \left(
        \mH_\theta \mH_\theta^\top
        \mH_{\theta'}\mH_{\theta'}^\top
    \right)
    =
    \left(\mH_\theta^\top \mH_{\theta'}\right)^2
    =
    \cos^2(\theta-\theta').
$$
Then, we have 
$
    \operatorname{CKA}(\mH_\theta,\mH_{\theta'})
    =
    \cos^2(\theta-\theta'),
$
and the corresponding chordal-CKA distance is
$
    \delta_{\rm C}(\mH_\theta,\mH_{\theta'})
    =
    \sqrt{
        1-\cos^2(\theta-\theta')
    }
    =
    |\sin(\theta-\theta')|.
$
Now we index the three matrices by
$
    \mH_1=\mH_0$, 
    $\mH_2=\mH_{\pi/6}$ and 
    $\mH_3=\mH_{\pi/3}$.
Their pairwise target distances are respectively 
$
    \mY_{12}=\mY_{23}=\frac12$, and $
    \mY_{13}=c:=\frac{\sqrt3}{2}$.
We next derive a necessary constraint on any exponentially calibrated
Euclidean realization. Fix arbitrary
$\varepsilon>0$, $k\ge1$, and
$\vv_1,\vv_2,\vv_3\in\R^k$, and let 
$
    \widehat \mY_{ij}
    =
    1-\exp\left(
        -\varepsilon\|\vv_i-\vv_j\|_2
    \right)$.
Let 
$
    d_{ij}=\|\vv_i-\vv_j\|_2.
$
By the Euclidean triangle inequality, we have 
$
    d_{13}\le d_{12}+d_{23}.
$
Since
$
    q_\varepsilon(t)=1-e^{-\varepsilon t}
$
is strictly increasing, we have 
$$
    \widehat \mY_{13}
    \le
    q_\varepsilon(d_{12}+d_{23}).
$$
For $s,t\ge0$, we have 
$$
\begin{aligned}
    q_\varepsilon(s+t)
    &=
    1-e^{-\varepsilon(s+t)}
    \\
    &=
    1-
    (1-q_\varepsilon(s))
    (1-q_\varepsilon(t))
    \\
    &=
    q_\varepsilon(s)
    +
    q_\varepsilon(t)
    -
    q_\varepsilon(s)q_\varepsilon(t).
\end{aligned}
$$
Thus, every exponentially calibrated Euclidean realization need to satisfy
$
\widehat \mY_{13}
\le
\widehat \mY_{12}
+
\widehat \mY_{23}
-
\widehat \mY_{12}\widehat \mY_{23}.
$
At the target values in the CKA counterexample targets, the right-hand side would equal
$
    \frac12+\frac12-\frac14
    =
    \frac34,
$
whereas
$
    \mY_{13}
    =
    \frac{\sqrt3}{2}
    >
    \frac34.
$
Therefore, exact realization is impossible.
For an arbitrary candidate realization, set
$
    E
    =
    \underset{1\le i<j\le3}{\max}
    |\widehat \mY_{ij}-\mY_{ij}|.
$
We show that necessarily $E\ge E_0$.
If $E\ge1/2$, then the desired conclusion follows immediately because $E_0<1/2$. Suppose that
$
    E<\frac12.
$
The definition of $E$ implies that 
$
    \widehat \mY_{12}\le\frac12+E$, $
    \widehat \mY_{23}\le\frac12+E$, and 
$
    \widehat \mY_{13}\ge c-E.
$
Since $E<1/2$, both $\widehat \mY_{12}$ and $\widehat \mY_{23}$ lie in $[0,1)$.
The function
$
    f(x,y)=x+y-xy
$
is nondecreasing in each argument on $[0,1]^2$, since 
$
    \frac{\partial f}{\partial x}=1-y\ge0$ and 
$    \frac{\partial f}{\partial y}=1-x\ge0.
$
Therefore, combining all these, we have 
$$
\begin{aligned}
    c-E
    &\le
    \widehat \mY_{13}
    \\
    &\le
    \widehat \mY_{12}
    +
    \widehat \mY_{23}
    -
    \widehat \mY_{12}\widehat \mY_{23}
    \\
    &\le
    2\left(\frac12+E\right)
    -
    \left(\frac12+E\right)^2
    \\
    &=
    \frac34+E-E^2.
\end{aligned}
$$
Thus,  we obtain
$
    E^2-2E+c-\frac34\le0$ and 
$
    (1-E)^2
    \le
    \frac74-c.
$
Because $E<1/2$, we have $1-E>0$, and thus 
$
    E
    \ge
    1-\sqrt{\frac74-c}.
$
Substituting $c=\sqrt3/2$ gives 
$$
    E
    \ge
    1-\sqrt{
        \frac74-\frac{\sqrt3}{2}
    }
    =
    1-\frac12\sqrt{7-2\sqrt3}
    =
    E_0.
$$

Thus, every exponentially calibrated Euclidean realization has
maximum absolute error at least $E_0$.
Finally, we show that this lower bound is attainable. Fix any $\varepsilon>0$, and take three collinear points in $\R$:
$
    v_1=0$, 
$    v_2=t$, and $
    v_3=2t$,
where
$
    t
    =
    -\frac1\varepsilon
    \log\left(\frac12-E_0\right).
$
Since $0<E_0<1/2$, this choice is well defined and $t>0$. By construction, we have 
$
    e^{-\varepsilon t}
    =
    \frac12-E_0,
$
and thus 
$$
    \widehat \mY_{12}
    =
    \widehat \mY_{23}
    =
    1-e^{-\varepsilon t}
    =
    \frac12+E_0.
$$
For the remaining pair, we have 
$$
    \widehat \mY_{13}
    =
    1-e^{-2\varepsilon t}
    =
    1-\left(\frac12-E_0\right)^2
    =
    \frac34+E_0-E_0^2.
$$
We can easily see that the defining relation
$
    E_0
    =
    1-\sqrt{\frac74-c}
$
is equivalent to
$    \frac34+E_0-E_0^2
    =
    c-E_0$.
Thus, we get 
$
    \widehat \mY_{13}=c-E_0.
$
It then follows that
$$
    |\widehat \mY_{12}-\mY_{12}|
    =
    |\widehat \mY_{23}-\mY_{23}|
    =
    |\widehat \mY_{13}-\mY_{13}|
    =
    E_0.
$$
Thus, the lower bound is attained already in one dimension, and
$
    E_\star=E_0.
$
Since $E_0>0$, exact realization is impossible in every embedding dimension.
\end{proof}

The corollary shows that Euclidean representability of a target does not imply representability after imposing the exponential distance-to-target map.

We now examine whether the graph metanetwork can produce the required embeddings by separating the graph-summary map $\vs_\psi(G)$ from
the shared readout $\varphi_\theta$ and the relation-specific linear head. For a fixed relation, the predictor has the form $
    \vz(G)=\varphi_\theta(\vs_\psi(G))$ and 
    $ \widehat y(G,G')
    =
    q\left(
        \left\|
            \mW_{\mathsf h}
            \left(\vz(G)-\vz(G')\right)
        \right\|_2
    \right)$.
We study the expressivity of the graph embedding, in which they identify the restrictions and establish the existence of suitable parameters under explicit assumptions.
\begin{proposition}
Fix an integer $R\ge0$. For each site $s$, let $G_s=\Psi(\mathcal C_s)$ be its finite
attributed computation graph, and let $T_s\subseteq V_s$ be the nonempty set of nodes pooled for the target representation site.
Consider the $R$-round independent graph metanetwork in Sec.~\ref{sec:proposed_method}, with shared node initialization, gated messages
\begin{equation}
 \vm_{ij,s}^{(r)}
 =
 \mW\vh_{i,s}^{(r)}
 \odot\psi_{\mathrm{edge}}(\va_{ij,s}),
\end{equation}
and shared residual node updates using both the sum and the mean of incoming messages, followed by nodewise layer normalization. Assume deterministic evaluation and a fixed convention for the mean of an empty incoming neighborhood.
Assume also that the global and target pooling operations are shared functions only of their respective multisets of final node states, and that the readout uses only the resulting summary.

For every node $i\in V_s$, let $c_{i,s}^{(0)}$ be a symbolic record of all node attributes used by the encoder, including node type and any target-membership marker supplied as an input. Define
$$
 c_{j,s}^{(r+1)}
 =
 \left(
   c_{j,s}^{(r)},
   \{\{
     (\va_{ij,s},c_{i,s}^{(r)})
     :i\in\mathcal N_s^-(j)
   \}\}
 \right)
 \text{ for } r=0,\ldots,R-1,
$$
where $\va_{ij,s}$ contains all edge attributes used by
the message function, $\{\{\cdot\}\}$ denotes a multiset,
and symbolic labels are compared across graphs.
Suppose two sites $u,u'$ satisfy
$$
 \begin{aligned}
 \{\{c_{i,u}^{(R)}:i\in V_u\}\}
 &=
 \{\{c_{i,u'}^{(R)}:i\in V_{u'}\}\},\\
 \{\{c_{i,u}^{(R)}:i\in T_u\}\}
 &=
 \{\{c_{i,u'}^{(R)}:i\in T_{u'}\}\}.
 \end{aligned}
$$
Then, for every common choice of the encoder and readout parameters,
we have  $
 \vz_u=\vz_{u'}$.
For the independent estimator $
 d_{\phi,\mathsf h}(s,t)
 =
 \left\|
   \mW_{\mathsf h}(\vz_s-\vz_t)
 \right\|_2$ and  $
 \widehat y_{st}^{(\mathsf h)}
 =
 q_{\mathsf h}\left(d_{\phi,\mathsf h}(s,t)\right)$,
with shared relation-specific heads and calibrations, every relation $\mathsf h\in\mathcal H$ and comparison partner $v$ satisfy $
 d_{\phi,\mathsf h}(u,v) =
 d_{\phi,\mathsf h}(u',v)$ and $
 \widehat y_{uv}^{(\mathsf h)}
 =
 \widehat y_{u'v}^{(\mathsf h)}$.
In addition, we have $
 d_{\phi,\mathsf h}(u,u')=0$ and $
 \widehat y_{uu'}^{(\mathsf h)}=q_{\mathsf h}(0)$. If $
 \Delta
 =
 \left|y_{uv}^{(\mathsf h)}
       -y_{u'v}^{(\mathsf h)}\right|>0$,
then we also have
\begin{equation*}
 \max\left\{
   \left|\widehat y_{uv}^{(\mathsf h)}
         -y_{uv}^{(\mathsf h)}\right|,
   \left|\widehat y_{u'v}^{(\mathsf h)}
         -y_{u'v}^{(\mathsf h)}\right|
 \right\}
 \ge \frac{\Delta}{2}.
\end{equation*}
\end{proposition}

\begin{proof}
Fix the shared encoder and readout parameters.
We show by induction that, at every round $r$, the hidden state is a common function of the corresponding symbolic label, i.e.,  $
 \vh_{i,s}^{(r)}
 =
 f_r(c_{i,s}^{(r)})$ for $s\in\{u,u'\}$.
The statement holds at initialization because the initial symbolic label records all inputs to the shared node initializer.

Suppose it holds at round $r$. If two receiving nodes have equal round-$(r+1)$ symbolic labels, then they have equal round-$r$ labels and equal incoming multisets of edge attributes and predecessor labels. The induction hypothesis then gives equal receiving-node states and equal incoming multisets of edge attributes and predecessor states.

Since $\mW$ and $\psi_{\mathrm{edge}}$ are shared, the corresponding multisets of gated messages agree. Their sums agree, and their means agree because multiset equality also preserves cardinality. For empty incoming neighborhoods, equality follows from the fixed convention. Thus the inputs to the shared residual update agree, and the shared nodewise layer normalization produces equal next states. This establishes the common function $f_{r+1}$ and completes the induction.

Applying $f_R$ to the two assumed multiset equalities gives equal multisets of final hidden states over the complete graphs and
over the target pooling sets. The shared pooling operations then produce equal summaries, and the shared readout gives $
 \vz_u=\vz_{u'}$.
For every relation $\mathsf h$ and partner $v$, it follows that $
 \mW_{\mathsf h}(\vz_u-\vz_v)
 =
 \mW_{\mathsf h}(\vz_{u'}-\vz_v)$.
Taking Euclidean norms and applying the same calibration $q_{\mathsf h}$ proves equality of the latent distances and predicted targets.
Likewise, $\vz_u-\vz_{u'}=\mathbf 0$ gives
$
 d_{\phi,\mathsf h}(u,u')=0$ and  $
 \widehat y_{uu'}^{(\mathsf h)}=q_{\mathsf h}(0)$.
Finally, let $
 \alpha
 =
 \widehat y_{uv}^{(\mathsf h)}
 =
 \widehat y_{u'v}^{(\mathsf h)}
$
for the common prediction.
The triangle inequality gives
$$
 \Delta
 \le
 \left|y_{uv}^{(\mathsf h)}-\alpha\right|
 +
 \left|\alpha-y_{u'v}^{(\mathsf h)}\right|\le
 2\max\left\{
   \left|\widehat y_{uv}^{(\mathsf h)}
         -y_{uv}^{(\mathsf h)}\right|,
   \left|\widehat y_{u'v}^{(\mathsf h)}
         -y_{u'v}^{(\mathsf h)}\right|
 \right\}.
$$
\end{proof}

We next assume that one common attainable backbone separates a finite bank, allowing us to isolate the constraints imposed by the calibrated Euclidean head from those imposed by graphs.

\begin{theorem}[Finite-bank realizability and spectral approximation]
\label{thm:finite-bank-realizability}
Fix graphs $G_1,\ldots,G_N$, with $N\ge2$, and an attainable setting $\psi$ of the shared backbone and pooling maps such that 
\begin{equation}
 \vs_i=\vs_\psi(G_i)\in\R^p, \text{ for } i=1,\ldots,N,
\end{equation}
are pairwise distinct. Keep this backbone setting fixed.
Consider one relation, with embedding dimension $k\ge1$ and head dimension $k_{\mathsf h}\ge1$, and predictors:
\begin{equation}
 \vz_i=\rho(\vs_i)\in\R^k, \text{ and } 
 \widehat{\mY}_{ij}
 =
 q\left(
   \left\|
     \mW_{\mathsf h}(\vz_i-\vz_j)
   \right\|_2
 \right),
\end{equation}
where $\mW_{\mathsf h}\in\R^{k_{\mathsf h}\times k}$ is unrestricted.
Allow the shared readout $\rho$ to range over finite-width ReLU MLPs with biases and an affine output, with unrestricted hidden widths and weights.
Let
$
 q:[0,\infty)\to[0,\infty)
$
be a fixed strictly increasing function satisfying $q(0)=0$.
Let $\mY=(\mY_{ij})\in\R^{N\times N}$ be symmetric,
entrywise nonnegative, and zero-diagonal, with every entry in
$q([0,\infty))$.
Define 
 $\mD_{ij}=q^{-1}(Y_{ij})$, $\mD=(\mD_{ij})$,
 $d_*=\min\{k,k_{\mathsf h}\}$, $
 \mJ_N
 =
 \mI_N-\frac1N\mathbf 1_N\mathbf 1_N^\top$, and 
 $\mB_{\rm D}
 =
 -\frac12\mJ_N\mD^{\circ2}\mJ_N$,
where $\mD^{\circ2}$ denotes the entrywise square.
Then,  there exists a shared readout $\rho$ and a head $\mW_{\mathsf h}$ satisfying
$
 \widehat{\mY}=\mY
$
on the entire bank if and only if
$
 \mB_{\rm D}\succeq0$ and  $
 \operatorname{rank}(\mB_{\rm D})\le d_*$.
When these conditions hold, a readout with one hidden ReLU layer of width at most $N-1$ suffices.

In addition, suppose $\mB_{\rm D}\succeq0$, and denote its eigenvalues as $
 \lambda_1\ge\cdots\ge\lambda_N\ge0$.
Let $d_{\max}=\max_{i,j}D_{ij}$.
If $q$ is $L_q$-Lipschitz on $[0,d_{\max}]$, then for every integer $1\le r\le\min\{d_*,N-1\},$
there is an attainable predictor $\widehat{\mY}^{(r)}$ such that $
 0\le\widehat{\mY}_{ij}^{(r)}\le{\mY}_{ij}$ for all $i,j$,
and
\begin{equation}
 \max_{i,j}
 \left|\widehat{\mY}_{ij}^{(r)}-Y_{ij}\right|
 \le
 L_q\sqrt{2\lambda_{r+1}}.
 \label{eq:ex-spectral}
\end{equation}
The bound is zero when
$r\ge\operatorname{rank}(\mB_{\rm D})$.
\end{theorem}

\begin{proof}
Suppose the predictor realizes $\mY$.
Define
$
 \ve_i=\mW_{\mathsf h}\vz_i$ and  $
 \mE=
 \begin{bmatrix}
   \ve_1^\top\\
   \vdots\\
   \ve_N^\top
 \end{bmatrix}$.
Since $q$ is injective, we have $
 \mD_{ij}=\|\ve_i-\ve_j\|_2$.
Let $\bm{\nu}\in\R^N$ have entries
$\nu_i=\|\ve_i\|_2^2$.
Then, we have 
$$
 \mD^{\circ2}
 =
 \bm{\nu}\mathbf 1_N^\top
 +
 \mathbf 1_N\bm{\nu}^\top
 -
 2\mE\mE^\top.
$$
Using $\mJ_N\mathbf 1_N=\mathbf 0$, we can see that 
$$
 \mB_{\rm D}
 =
 (\mJ_N\mE)
 (\mJ_N\mE)^\top
 \succeq0.
$$
Moreover, if $\vz$ has rows $\vz_i^\top$, then
$\mE=\vz\mW_{\mathsf h}^\top$, and thus
$$
 \operatorname{rank}(\mB_{\rm D})
 \le
 \operatorname{rank}(\mE)
 \le
 \operatorname{rank}(\mW_{\mathsf h})
 \le d_*.
$$

We now prove in another direction. 
Suppose $\mB_{\rm D}\succeq0$ and
$r_0=\operatorname{rank}(\mB_{\rm D})\le d_*$.
Let $\vb_i$ denote the $i$-th standard basis vector
in $\R^N$.
Since
$\mJ_N(\vb_i-\vb_j)=\vb_i-\vb_j$
and $\mD$ is symmetric and zero-diagonal, we have 
$$
 (\vb_i-\vb_j)^\top
 \mB_{\rm D}
 (\vb_i-\vb_j) =
 -\frac12
 (\vb_i-\vb_j)^\top
 \mD^{\circ2}
 (\vb_i-\vb_j)=\mD_{ij}^2.
$$
If $r_0=0$, this identity gives $\mD=\mathbf 0$ and thus $\mY=\mathbf 0$.
A constant-zero readout realizes the target matrix.
Now suppose $r_0>0$ and factor $
 \mB_{\rm D}=\mX\mX^\top$ with $\mX\in\R^{N\times r_0}$.
Denote the rows of $\mX$ as $\vx_i^\top$.
Then, we have  $\|\vx_i-\vx_j\|_2^2=\mD_{ij}^2$ and $
 \|\vx_i-\vx_j\|_2=\mD_{ij}$.
Also, $\mB_{\rm D}\mathbf 1_N=\mathbf 0$ implies
$\mX^\top\mathbf 1_N=\mathbf 0$, so these coordinates
are centered.
Since $\vs_i$ are pairwise distinct, there exists
$\va\in\R^p$ such that $
 t_i=\va^\top\vs_i$ are pairwise distinct.
Note that the forbidden choices of $\va$ form a finite union of proper hyperplanes.
After relabeling the indices and assuming $
 t_1<\cdots<t_N$.
Define
$
 \vv_0=\mathbf 0$, 
 $\vv_j
 =
 \frac{\vx_{j+1}-\vx_j}{t_{j+1}-t_j}$ for 
 $j=1,\ldots,N-1$,
and the vector-valued piecewise-linear function
$$
 f(t)
 =
 \vx_1+
 \sum_{j=1}^{N-1}
   (\vv_j-\vv_{j-1})(t-t_j)_+,$$
 where $(t-t_j)_+=\max\{0,t-t_j\}.$
This function satisfies $f(t_1)=\vx_1$ and has
slope $\vv_j$ on $(t_j,t_{j+1})$.
Thus, we have $
 f(t_i)=\vx_i$ for every $i$.
The map $\vs\mapsto f(\va^\top\vs)$
is a ReLU network with one hidden layer of width $N-1$ and an
affine output.
Pad this output with zeros to obtain a readout
$\rho:\R^p\to\R^k$ satisfying
$
 \rho(\vs_i)
 =
 \begin{bmatrix}
   \vx_i\\
   \mathbf 0
 \end{bmatrix}.
$
Choose $\mW_{\mathsf h}$ to copy the first $r_0$
coordinates into $\R^{k_{\mathsf h}}$ and set all remaining
output coordinates to zero.
Then, we have 
$
 \left\|
   \mW_{\mathsf h}
   \left(\rho(\vs_i)-\rho(\vs_j)\right)
 \right\|_2
 =
 \|\vx_i-\vx_j\|_2
 =
 \mD_{ij}
$, and  applying $q$, we get $\widehat{\mY}=\mY$.

For the spectral approximation, let $\vu_1,\ldots,\vu_N$ be an orthonormal eigenbasis
of $\mB_{\rm D}$ corresponding to the ordered
eigenvalues, and define
$$
 \mB_r
 =
 \sum_{\ell=1}^r
 \lambda_\ell\vu_\ell\vu_\ell^\top,
 \qquad
 \mC_r=\mB_{\rm D}-\mB_r.
$$
Then, we can see that 
$
 \mC_r\succeq0$ and  $
 \|\mC_r\|_{\mathrm{op}}=\lambda_{r+1}$.
Factor $\mB_r=\mX_r\mX_r^\top$ with
$\mX_r\in\R^{N\times r}$, and let $\mD_{ij}^{(r)}$ be the Euclidean distances between its rows.
For every $i,j$, we have
$$
 \begin{aligned}
 0
 &\le
 \mD_{ij}^2-\left(\mD_{ij}^{(r)}\right)^2\\
 &=
 (\vb_i-\vb_j)^\top
 \mC_r
 (\vb_i-\vb_j)\\
 &\le
 \|\mC_r\|_{\mathrm{op}}
 \|\vb_i-\vb_j\|_2^2\\
 &\le 2\lambda_{r+1}.
 \end{aligned}
$$
Thus $0\le \mD_{ij}^{(r)}\le \mD_{ij}$.
Using $(x-y)^2\le x^2-y^2$ for $x\ge y\ge0$, we obtain
$$
 0\le \mD_{ij}-\mD_{ij}^{(r)}
 \le\sqrt{2\lambda_{r+1}}.
$$

The interpolation construction above realizes the rows of $\mX_r$  through the readout and head, since
$r\le d_*$.
For the resulting predictor, we have 
$
 \widehat{\mY}_{ij}^{(r)}=q(\mD_{ij}^{(r)}).
$
Both $\mD_{ij}^{(r)}$ and $\mD_{ij}$ lie in $[0,d_{\max}]$.
Monotonicity and the Lipschitz property of $q$ thus give
$$
 \begin{aligned}
 0
 &\le{\mY}_{ij}-\widehat{\mY}_{ij}^{(r)}\\
 &=
 q(\mD_{ij})-q(\mD_{ij}^{(r)})\\
 &\le
 L_q\left(\mD_{ij}-\mD_{ij}^{(r)}\right)\\
 &\le L_q\sqrt{2\lambda_{r+1}}.
 \end{aligned}
$$

\end{proof}

To study approximation throughout a domain, we next assume compactness and continuity.

\begin{theorem}[Uniform approximation]
\label{thm:ex-uniform-robust}
Let $\mathcal K$ be a nonempty compact metric space, and fix a continuous attainable $
 \vs_\psi:\mathcal K\to\R^p$.
Let $\Phi:\mathcal K\to\mathcal H$ be continuous, where $\mathcal H$ is a real Hilbert space.
Define $
 S=\vs_\psi(\mathcal K)$,
 $M_\Phi=\sup_{G\in\mathcal K}\|\Phi(G)\|_{\mathcal H}$,
and $
 \rho(G,G')
 =
 \|\Phi(G)-\Phi(G')\|_{\mathcal H}$.
Let $q:[0,\infty)\to\R$ be fixed and
$L_q$-Lipschitz on $[0,2M_\Phi+2]$.
For a target function
$y:\mathcal K\times\mathcal K\to\R$, assume
\begin{equation}
 \kappa
 =
 \sup_{G,G'\in\mathcal K}
 |q(\rho(G,G'))-y(G,G')|
 <\infty.
\end{equation}

Consider independent predictors
\begin{equation}
 \vz_\theta(G)
 =
 \varphi_\theta(\vs_\psi(G))\in\R^k \text{ and }
 \widehat y(G,G')
 =
 q\left(
   \left\|
     \mW_{\mathsf h}
     \left(\vz_\theta(G)-\vz_\theta(G')\right)
   \right\|_2
 \right),
\end{equation}
where $\mW_{\mathsf h}\in\R^{k_{\mathsf h}\times k}$
is unrestricted. Assume that the allowed readouts uniformly approximate arbitrary continuous finite-dimensional maps on compact subsets of $\R^p$, with error measured in the Euclidean norm.

For a finite-dimensional subspace $V\subseteq\mathcal H$, we define
\begin{equation}
 \alpha_{\psi,V}
 =
 \sup_{\vs\in S}
 \inf_{\mathbf c\in V}
 \sup_{\substack{G\in\mathcal K\\
                  \vs_\psi(G)=\vs}}
 \|P_V\Phi(G)-\mathbf c\|_{\mathcal H},
\end{equation}
and
\begin{equation}
 \vr_V(G)=\Phi(G)-P_V\Phi(G),
 \qquad
 \beta_V
 =
 \sup_{G,G'\in\mathcal K}
 \|\vr_V(G)-\vr_V(G')\|_{\mathcal H}.
\end{equation}
Then, for every $0<a\le1$ and positive integers $k,k_{\mathsf h}$
satisfying $
 k\ge\dim V$ and $
 k_{\mathsf h}\ge\dim V$, there exists a shared readout and a linear head such that
\begin{equation}
 \sup_{G,G'\in\mathcal K}
 |\widehat y(G,G')-y(G,G')|
 \le
 L_q\sqrt{4(\alpha_{\psi,V}+a)^2+\beta_V^2}
 +\kappa.
\end{equation}
\end{theorem}

\begin{proof}
Fix $V$ and $a$, and abbreviate
$\alpha=\alpha_{\psi,V}$.
For each $\vs\in S$, the set
\[
 \mathcal A_{\vs,V}
 =
 \left\{
   P_V\Phi(G):
   G\in\mathcal K,\ \vs_\psi(G)=\vs
 \right\}
\]
is nonempty and compact.
Since $V$ is finite-dimensional, there is a center
$\mathbf c_{\vs}\in V$ minimizing
\[
 \sup_{\mathbf v\in\mathcal A_{\vs,V}}
 \|\mathbf v-\mathbf c_{\vs}\|_{\mathcal H}.
\]
A minimizing center can be chosen in the closed convex hull of $\mathcal A_{\vs,V}$, i.e.,  projecting any center onto that convex hull cannot increase its distance to any point of $\mathcal A_{\vs,V}$.
Therefore, we have $
 \|\mathbf c_{\vs}\|_{\mathcal H}\le M_\Phi$ and 
and
\[
 \sup_{\substack{G\in\mathcal K\\
                  \vs_\psi(G)=\vs}}
 \|P_V\Phi(G)-\mathbf c_{\vs}\|_{\mathcal H}
 \le\alpha.
\]
Now, we define the compact set
\[
 \mathcal{C}_{\vs}
 =
 \left\{
   G\in\mathcal K:
   \|P_V\Phi(G)-\mathbf c_{\vs}\|_{\mathcal H}
   \ge\alpha+\frac a2
 \right\}.
\]
The set $\vs_\psi(\mathcal{C}_{\vs})$ is compact and does not contain $\vs$. Thus,  $
 \mathcal{U}_{\vs}
 =
 S\setminus\vs_\psi(\mathcal{C}_{\vs})$
is an open neighborhood of $\vs$.
For every $G$ whose summary lies in $\mathcal{U}_{\vs}$, we have $
 \|P_V\Phi(G)-\mathbf c_{\vs}\|_{\mathcal H}
 <\alpha+\frac a2$.
Choose a finite subcover $U_{\vs_1},\ldots,U_{\vs_m}$ of $S$ and continuous
nonnegative functions $\chi_1,\ldots,\chi_m$ such that
\[
 \sum_{j=1}^m\chi_j(\vs)=1,
 \qquad
 \chi_j(\vs)>0
 \ \Longrightarrow\
 \vs\in U_{\vs_j}.
\]
Such functions can be constructed by normalizing $
 w_j(\vs)
 =
 \operatorname{dist}
 \left(\vs,S\setminus U_{\vs_j}\right)$,
with $w_j\equiv1$ when $U_{\vs_j}=S$.
Their sum is positive everywhere because the sets form an open cover.
The continuous map $
 F_V(\vs)
 =
 \sum_{j=1}^m
 \chi_j(\vs)\mathbf c_{\vs_j}
 \in V
$ then  satisfies $
 \sup_{\vs\in S}
 \|F_V(\vs)\|_{\mathcal H}\le M_\Phi$
and $
 \sup_{G\in\mathcal K}
 \|F_V(\vs_\psi(G))-P_V\Phi(G)\|_{\mathcal H}
 \le\alpha+\frac a2$.

Choose an orthonormal basis of $V$, identify $F_V$ with its coordinate map into $\R^{\dim V}$, and pad with zeros to dimension $k$. The readout approximation property gives a shared readout with uniform Euclidean error at most $a/2$ from this padded map.
Choose $\mW_{\mathsf h}$ to select these coordinates
and pad with zeros in the head output.
Identifying the selected coordinates with a vector
$\vw(G)\in V$, we obtain
\[
 \sup_{G\in\mathcal K}
 \|\vw(G)-F_V(\vs_\psi(G))\|_{\mathcal H}
 \le\frac a2.
\]
Therefore, we have 
\[
 \sup_{G\in\mathcal K}
 \|\vw(G)-P_V\Phi(G)\|_{\mathcal H}
 \le\alpha+a,
\text{and}
 \sup_{G\in\mathcal K}
 \|\vw(G)\|_{\mathcal H}
 \le M_\Phi+\frac a2.
\]
The distance used by the predictor is then
\[
 \widehat\rho(G,G')
 =
 \|\vw(G)-\vw(G')\|_{\mathcal H}.
\]
If $V=\{0\}$, take a zero head and interpret these coordinate maps as empty.

Let $\vu(G)=\vw(G)-P_V\Phi(G)\in V$.
For any $G,G'\in\mathcal K$, the difference between the predicted and ideal difference vectors is
\[
\left(\vw(G)-\vw(G')\right)
 -\left(\Phi(G)-\Phi(G')\right)=
 \left(\vu(G)-\vu(G')\right)
 -
 \left(\vr_V(G)-\vr_V(G')\right).
\]
The first term lies in $V$ and the second lies in $V^\perp$.
Thus the reverse triangle inequality and orthogonality give
\[
 \begin{aligned}
 |\widehat\rho(G,G')-\rho(G,G')|
 &\le
 \left(
   \|\vu(G)-\vu(G')\|_{\mathcal H}^2
   +
   \|\vr_V(G)-\vr_V(G')\|_{\mathcal H}^2
 \right)^{1/2}\\
 &\le
 \sqrt{4(\alpha+a)^2+\beta_V^2}.
 \end{aligned}
\]
In addition, we also have $
 \rho(G,G')\le2M_\Phi$ and  $ \widehat\rho(G,G')\le2M_\Phi+a$,
thus  both distances lie in the stated Lipschitz interval.
Therefore, we have
\[
 \begin{aligned}
 |\widehat y(G,G')-y(G,G')|
 &\le
 |q(\widehat\rho(G,G'))-q(\rho(G,G'))|+
 |q(\rho(G,G'))-y(G,G')|\\
 &\le
 L_q\sqrt{4(\alpha+a)^2+\beta_V^2}+\kappa.
 \end{aligned}
\]
\end{proof}

\subsection{Score error, layer retrieval and stitching}
\label{app:Score error, layer retrieval and stitching}

\begin{proposition}[Stability of retrieval and Hausdorff aggregation]
Let $\mathcal X$ be a set of sites, let
$y,\widehat y:\mathcal X\times\mathcal X\to\R$,
and let $E\ge 0$.

\smallskip
\noindent\emph{(i) Retrieval.}
Fix a query $u\in\mathcal X$ and a nonempty finite candidate set
$\mathcal V\subseteq\mathcal X$, and suppose that
$
 |\widehat y_{uv}-y_{uv}|\le E$ for every $v\in\mathcal V$.
Choose, with arbitrary tie-breaking,
$
 v^\star\in\arg\min_{v\in\mathcal V}y_{uv}$ and 
$
 \widehat v\in\arg\min_{v\in\mathcal V}\widehat y_{uv}$.
Then $
 0\le y_{u\widehat v}-y_{uv^\star}\le 2E.$
Denote $n=|\mathcal V|$, and let $
 y_{u,(1)}\le \cdots\le y_{u,(n)}$
be the target scores in nondecreasing order, counted with multiplicity.
For $1\le k<n$, if $
 y_{u,(k+1)}-y_{u,(k)}>2E$,
then $
 T_k=\{v\in\mathcal V:y_{uv}\le y_{u,(k)}\}$
has cardinality $k$, and every selection of the $k$ smallest predicted scores recovers $T_k$, regardless of tie-breaking.
Specifically, if $n\ge 2$ and $
 y_{u,(2)}-y_{u,(1)}>2E$,
then $v^\star$ is the unique minimizer of both the target and predicted scores, so $\widehat v=v^\star$. For $k=n$, top-$k$ recovery is obtained  and the nearest-site recovery is
also obtained when $n=1$.

\smallskip
\noindent\emph{(ii) Hausdorff aggregation.}
For nonempty finite sets $A,B\subseteq\mathcal X$ and
$z\in\{y,\widehat y\}$, define
\begin{equation}
 \mathfrak H_z(A,B)
 =
 \max\left\{
   \max_{a\in A}\min_{b\in B}z_{ab},
   \max_{b\in B}\min_{a\in A}z_{ab}
 \right\}.
\end{equation}
If $
 |\widehat y_{ab}-y_{ab}|\le E$ for every $(a,b)\in A\times B$,
then $
 \left|\mathfrak H_{\widehat y}(A,B)
       -\mathfrak H_y(A,B)\right|\le E$.
If $y$ is a pseudometric on $\mathcal X$, then $\mathfrak H_y$ is a pseudometric on the collection of nonempty finite subsets
of $\mathcal X$. If $y$ is a metric, then $\mathfrak H_y$ is a metric on that collection.
\end{proposition}

\begin{proof}
For the retrieval bound, the optimality of $\widehat v$ for the predicted scores gives
$$
 y_{u\widehat v}
 \le \widehat y_{u\widehat v}+E
 \le \widehat y_{uv^\star}+E
 \le y_{uv^\star}+2E.
$$
The lower bound $y_{u\widehat v}-y_{uv^\star}\ge 0$ follows from optimality of $v^\star$ for the target scores.
Now suppose that $1\le k<n$ and
$y_{u,(k+1)}-y_{u,(k)}>2E$.
Specifically, $y_{u,(k)}<y_{u,(k+1)}$, so $T_k$ has $k$ elements.
For every $v\in T_k$ and $w\in\mathcal V\setminus T_k$, we have
$$
 \widehat y_{uv}
 \le y_{u,(k)}+E
 < y_{u,(k+1)}-E
 \le \widehat y_{uw}.
$$
Thus every candidate in $T_k$ has a strictly smaller predicted score
than every candidate outside $T_k$. Therefore, every predicted
top-$k$ selection equals $T_k$.
The nearest-site statement follows by taking $k=1$. The cases $k=n$ and $n=1$ are then obtained.

For the aggregation bound, the entrywise error assumption implies, for each $a\in A$, we have
$$
 \min_{b\in B}y_{ab}-E
 \le \min_{b\in B}\widehat y_{ab}
 \le \min_{b\in B}y_{ab}+E.
$$
Taking the maximum over $a\in A$ preserves these inequalities.
The same argument applies to the column minima and their maximum
over $b\in B$.
Taking the maximum of the two resulting terms then gives
$$
 \mathfrak H_y(A,B)-E
 \le \mathfrak H_{\widehat y}(A,B)
 \le \mathfrak H_y(A,B)+E.
$$
Finally, suppose that $y$ is a pseudometric, and define
$$
 h_y(A,B)=\max_{a\in A}\min_{b\in B}y_{ab}.
$$
By symmetry of $y$, we have
$$
 \mathfrak H_y(A,B)
 =\max\{h_y(A,B),h_y(B,A)\}.
$$

Let $A,B,C$ be nonempty finite subsets of $\mathcal X$.
For each $a\in A$, choose $
 b_a\in\arg\min_{b\in B}y_{ab}$ and $
 c_a\in\arg\min_{c\in C}y_{b_a c}$.
These minimizers exist since the sets are nonempty and finite. The triangle inequality for $y$ then gives
$$
 \min_{c\in C}y_{ac}
 \le y_{a c_a}
 \le y_{a b_a}+y_{b_a c_a}
 \le h_y(A,B)+h_y(B,C).
$$
Taking the maximum over $a\in A$ shows that
$$
 h_y(A,C)\le h_y(A,B)+h_y(B,C).
$$
Applying the same argument in the reverse direction gives
$$
 h_y(C,A)\le h_y(C,B)+h_y(B,A).
$$
Thus, we have
$$
 \begin{aligned}
 \mathfrak H_y(A,C)
 &=\max\{h_y(A,C),h_y(C,A)\}\\
 &\le
 \max\left\{
   h_y(A,B)+h_y(B,C),
   h_y(C,B)+h_y(B,A)
 \right\}\\
 &\le \mathfrak H_y(A,B)+\mathfrak H_y(B,C),
 \end{aligned}
$$
and we also have nonnegativity and symmetry.
In addition, note that  $\mathfrak H_y(A,A)=0$ because $y_{aa}=0$ for every $a\in A$.
Thus $\mathfrak H_y$ is a pseudometric.
If $y$ is a metric and $\mathfrak H_y(A,B)=0$, then every $a\in A$ has some $b\in B$ with $y_{ab}=0$, and thus $a=b$. Thus $A\subseteq B$.  The reverse inclusion follows similarly.
Thus, $A=B$ and $\mathfrak H_y$ is a metric.
\end{proof}

\begin{proposition}
Let $a,b$ index two supervised pairs, with real-valued targets $y_a,y_b$ and predictions $\widehat y_a,\widehat y_b$.
Fix $\tau>0$, use the convention $\operatorname{sign}(0)=0$, and define $
 s_{ab}=\operatorname{sign}(y_a-y_b)$, $
 m_{ab}=s_{ab}(\widehat y_a-\widehat y_b)$,
and $
 \ell_\tau(a,b)
 =\log\left(1+\exp(-\tau m_{ab})\right)$.
Then, we have
\begin{equation}
 \mathbf 1\{y_a\ne y_b,\ m_{ab}\le 0\}
 \le
 \frac{\mathbf 1\{y_a\ne y_b\}\,\ell_\tau(a,b)}{\log 2}
 \le
 \frac{\ell_\tau(a,b)}{\log 2}.
 \label{eq:theory-ranking}
\end{equation}
Thus, under any joint distribution of the compared targets and predictions, the ranking error probability $
 R_{\mathrm{rank}}
 :=\mathbb P\left(y_a\ne y_b,\ m_{ab}\le 0\right)$
satisfies
\begin{equation}
 R_{\mathrm{rank}} \le
 \frac{\mathbb E[
   \mathbf 1\{y_a\ne y_b\}\,\ell_\tau(a,b)
 ]}{\log 2}=
 \frac{\mathbb E[\ell_\tau(a,b)]}{\log 2}
 -\mathbb P(y_a=y_b)\le
 \frac{\mathbb E[\ell_\tau(a,b)]}{\log 2}.
\end{equation}
The reversed predictions and predicted ties are counted as errors only when the targets are unequal and the multiplicative constant $1/\log 2$ is optimal.
\end{proposition}

\begin{proof}
Note that if $y_a=y_b$, then the error indicator and the masked loss
$\mathbf 1\{y_a\ne y_b\}\ell_\tau(a,b)$ both vanish.
Suppose now that $y_a\ne y_b$.
If $m_{ab}\le 0$, then $\tau>0$ implies
$$
 \ell_\tau(a,b)
 =\log\left(1+\exp(-\tau m_{ab})\right)
 \ge \log(1+\exp(0))
 =\log 2.
$$
If $m_{ab}>0$, the error indicator is zero, while the loss is nonnegative.

When $y_a=y_b$, we have $s_{ab}=0$ and thus
$\ell_\tau(a,b)=\log 2$.
Thus, we have
$$
 \ell_\tau(a,b)
 =
 \mathbf 1\{y_a\ne y_b\}\,\ell_\tau(a,b)
 +
 \mathbf 1\{y_a=y_b\}\,\log 2.
$$
The final inequality follows from $\mathbb P(y_a=y_b)\ge 0$.
If the expected loss is infinite, the bounds hold in the extended-real sense.
Finally, take $y_a=1$, $y_b=0$, and
$\widehat y_a=\widehat y_b=0$.
Then the error indicator equals $1$ and
$\ell_\tau(a,b)=\log 2$ for every $\tau>0$.
Thus any universal constant $C$ satisfying
$$
 \mathbf 1\{y_a\ne y_b,\ m_{ab}\le 0\}
 \le C\,\ell_\tau(a,b)
$$
need to satisfy $C\ge 1/\log 2$.
\end{proof}

We fix one relation $\mathsf h$ and a finite collection of supervised site pairs $\mathcal P=\{p_1,\ldots,p_n\}$, with $n\ge2$.
Denote  $
 y_i=y_{p_i}^{(\mathsf h)}$, $
 \widehat y_i=\widehat y_{p_i}^{(\mathsf h)}$, and 
 $\varepsilon_i=\widehat y_i-y_i$, 
and suppress the relation superscript below.

If an index $I$ is chosen uniformly from $\{1,\ldots,n\}$, then $Y=y_I$ and $\widehat Y=\widehat y_I$ are the target and prediction on the same random site pair.
For $\beta>0$, we define
\[
 h_\beta(t)=
 \begin{cases}
   t^2/(2\beta),& |t|\le\beta,\\
   |t|-\beta/2,& |t|>\beta,
 \end{cases}
\]
and the empirical regression loss
\[
 \mathcal L_{\rm reg}
 =\mathbb E[h_\beta(\widehat Y-Y)]
 =\frac1n\sum_{i=1}^n h_\beta(\varepsilon_i).
\]
For $\tau>0$, let $\operatorname{sign}(0)=0$ and define $
 s_{ij}=\operatorname{sign}(y_i-y_j)$ and 
 $\ell_{ij}
 =\log\left(
   1+\exp[-\tau s_{ij}(\widehat y_i-\widehat y_j)]
 \right)$.
For the analysis, we use the complete-comparison ordinal average $
 M=\binom n2$ and  
 $\mathcal L_{\rm rank}
 =\frac1M\sum_{i<j}\ell_{ij}$.
Therefore, the comparisons are averaged uniformly over distinct examples, and the target ties are retained. 

For $t\ge0$, we define the fraction of comparisons with target gap at most $t$ and their contribution to the ordinal loss by
\[
 F_n(t)
 =\frac1M\sum_{i<j}\mathbf1\{|y_i-y_j|\le t\} 
 \quad\text{ and } \quad
 \mathcal L_{\rm rank}^{\le t}
 =\frac1M\sum_{i<j}
   \mathbf1\{|y_i-y_j|\le t\}\ell_{ij}.
\]
Both quantities are normalized by all $M$ comparisons.

\begin{proposition}[Joint loss control of empirical Spearman correlation]
\label{prop:loss-spearman}
Suppose that $|\widehat Y-Y|\le B$ almost surely for some $0\le B<\infty$, equivalently $|\varepsilon_i|\le B$ for every $i$. Assume that $\vy=(y_1,\ldots,y_n)^\top$ 
$\widehat{\vy}=(\widehat y_1,\ldots,\widehat y_n)^\top$ are not constant.
Let $\vr$ and $\widehat{\vr}$ be their increasing
rank vectors, assigning average ranks to tied values.
Then, we have 
\begin{equation}
 1-\rho_{\rm S}
 \le \Gamma_n\inf_{t>0}\left[\min\left\{
     F_n(t),
     \frac{\mathcal L_{\rm rank}^{\le t}}{\log2}
   \right\}+\frac{2n\,C_\beta(B)}{(n-1)t^2}
      \mathcal L_{\rm reg}
 \right],
\end{equation}
where   $
 \bar r=\frac{n+1}{2}$, $
 A=\|\vr-\bar r\mathbf1_n\|_2$, $
 \widehat A=\|\widehat{\vr}-\bar r\mathbf1_n\|_2$,
 $\rho_{\rm S}
 =\frac{
   (\vr-\bar r\mathbf1_n)^\top
   (\widehat{\vr}-\bar r\mathbf1_n)
 }{A\widehat A}$,
 $\Gamma_n=\frac{nM}{2A\widehat A}$, and $
 C_\beta(B)=2\max\{B,\beta\}$.
If neither vector contains ties, then $
 \Gamma_n=\frac{3n}{n+1}$.
If in addition $Y,\widehat Y\in[0,1]$ almost surely and $\beta=1$, we can take $B=1$ and $C_1(1)=2$ and in this case, we have  $
 \mathbb E[(\widehat Y-Y)^2]=2\mathcal L_{\rm reg}$.
\end{proposition}

\begin{proof}
Note that for $|x|\le\beta$, we have $x^2=2\beta h_\beta(x)$.
For $\beta<|x|\le B$, we have  $
 h_\beta(x)=|x|-\frac\beta2\ge\frac{|x|}{2}$ and  
 $x^2\le2B h_\beta(x)$.
Thus, we have $
 r_n\le C_\beta(B)\mathcal L_{\rm reg}$ and 
 $\frac1n\sum_{i=1}^n|\varepsilon_i|\le\sqrt{r_n}$, where $r_n=\frac1n\sum_i\varepsilon_i^2$ and the second inequality is Cauchy-Schwarz.
If $B\le\beta$, every error lies in the quadratic branch and thus $r_n=2\beta\mathcal L_{\rm reg}$.

Let  $
 \widehat s_{ij}
 =\operatorname{sign}(\widehat y_i-\widehat y_j)$, 
 $\xi_{ij}=\frac{s_{ij}-\widehat s_{ij}}2$, and  
 $Q_n=\sum_{i<j}\xi_{ij}^2$.
Average ranks satisfy $
 r_i=\bar r+\frac12\sum_{j=1}^n s_{ij}$ and 
 $\widehat r_i=\bar r+\frac12\sum_{j=1}^n\widehat s_{ij}$.
Thus, for $\vd=\vr-\widehat{\vr}$, we have  
 $d_i=\sum_j\xi_{ij}$, 
 $\xi_{ji}=-\xi_{ij}$, 
 and $\sum_i d_i=0$.
By antisymmetry and Cauchy-Schwarz, we obtain
\[
 \|\vd\|_2^2
 =\sum_{i<j}(d_i-d_j)\xi_{ij}\le
 \left(\sum_{i<j}(d_i-d_j)^2\right)^{1/2}Q_n^{1/2}=\sqrt n\,\|\vd\|_2Q_n^{1/2},
\]
where the last equality uses
$\sum_{i<j}(d_i-d_j)^2=n\|\vd\|_2^2$ when $\sum_i d_i=0$.
Dividing when $\|\vd\|_2>0$, and treating the zero case
directly, gives $\|\vd\|_2^2\le nQ_n$.
Thus, we have 
\[
 2A\widehat A(1-\rho_{\rm S})
 =\|\vd\|_2^2-(A-\widehat A)^2\le nQ_n,
\]
and therefor 
 $1-\rho_{\rm S}\le\Gamma_n\frac{Q_n}{M}$.

Consider  $t>0$. If $\xi_{ij}\ne0$, the target and predicted signs differ, and thus  $
 s_{ij}(\widehat y_i-\widehat y_j)\le0$.
This includes reversals and comparisons tied in exactly one of the two vectors. It implies $\ell_{ij}\ge\log2$.
Since $|\xi_{ij}|\le1$, we have
$\xi_{ij}^2\le\ell_{ij}/\log2$ for every comparison.
Therefore, we have 
\begin{equation*}
 \frac1M\sum_{\substack{i<j\\|y_i-y_j|\le t}}\xi_{ij}^2
 \le
 \min\left\{
   F_n(t),\frac{\mathcal L_{\rm rank}^{\le t}}{\log2}
 \right\}.
\end{equation*}

For a comparison with $|y_i-y_j|>t$ and $\xi_{ij}\ne0$, the predicted order is reversed or tied, and thus $
 |\varepsilon_i-\varepsilon_j|
 \ge |y_i-y_j|>t$.
It then follows that
\[
 \frac1M\sum_{\substack{i<j\\|y_i-y_j|>t}}\xi_{ij}^2
\le\frac1{Mt^2}\sum_{i<j}(\varepsilon_i-\varepsilon_j)^2=\frac{
    n\sum_i\varepsilon_i^2-(\sum_i\varepsilon_i)^2
   }{Mt^2}\le\frac{2n}{(n-1)t^2}r_n\le\frac{2n\,C_\beta(B)}{(n-1)t^2}\mathcal L_{\rm reg}.
\]

Finally, without ties both rank vectors are permutations of $(1,\ldots,n)$, and thus  $
 A^2=\widehat A^2=\frac{n(n^2-1)}{12}$.
Substitution into the definition of $\Gamma_n$ then gives
$\Gamma_n=3n/(n+1)$.
\end{proof}

\begin{theorem}
\label{thm:retrieval}
Fix one relation and a trained metanetwork.
Draw a query and a candidate list of fixed size $K\ge2$ from a specified joint distribution.
All expectations and probabilities below are conditional on the trained metanetwork and are taken over this query-list distribution.

For $j=1,\ldots,K$, let $Y_j$ and $\widehat Y_j$ be the real-valued target and predicted relation values for candidate $j$.
Denote $
 \varepsilon_j=\widehat Y_j-Y_j$ and 
 $r_K=\mathbb E\left[
   \frac1K\sum_{j=1}^K\varepsilon_j^2
 \right]<\infty$.
Let $
 \widehat j\in\arg\min_{1\le j\le K}\widehat Y_j$ and  $
 j_\star\in\arg\min_{1\le j\le K}Y_j$,
using fixed measurable tie-breaking rules, and define the retrieval regret as $
 \mathcal R=Y_{\widehat j}-Y_{j_\star}$.
Then, we have 
$
 \mathbb E[\mathcal R^2]\le2Kr_K$, and 
 $0\le\mathbb E[\mathcal R]\le\sqrt{2Kr_K}$.

If the best target candidate is unique almost surely, let
\begin{equation}
 \Delta=Y_{(2)}-Y_{(1)}>0
 \qquad\text{almost surely},
\end{equation}
where $Y_{(1)}\le Y_{(2)}$ are the smallest two target order statistics. Then, for every $\gamma>0$, we have 
\begin{equation}
 \Pr\{\widehat j\ne j_\star\}
 \le
 \min\left\{
   1,\;
   \Pr\{\Delta\le\gamma\}+\frac{2Kr_K}{\gamma^2}
 \right\}.
\end{equation}
If in addition $Y_j\in[0,B_{\rm tar}]$ almost surely for every $j$, then we have  $
 \mathbb E[\mathcal R]
 \le\min\{B_{\rm tar},\sqrt{2Kr_K}\}$.
\end{theorem}

\begin{proof}
Consider a fixed realized query and candidate list.
By the optimality of $j_\star$ and $\widehat j$, we have 
\[
 0\le\mathcal R
 =Y_{\widehat j}-Y_{j_\star}=\widehat Y_{\widehat j}-\widehat Y_{j_\star}
   +\varepsilon_{j_\star}-\varepsilon_{\widehat j}\le\varepsilon_{j_\star}-\varepsilon_{\widehat j}.
\]
If $\widehat j=j_\star$, then $\mathcal R=0$.
Otherwise the two indices are different, and thus 
\[
 \mathcal R^2
 \le
 (\varepsilon_{j_\star}-\varepsilon_{\widehat j})^2\le
 2(\varepsilon_{j_\star}^2+\varepsilon_{\widehat j}^2)\le
 2\sum_{j=1}^K\varepsilon_j^2, 
\]
where the last bound also holds when the indices are the same. 
Taking expectations and applying Cauchy-Schwarz, then we have  $
 \mathbb E[\mathcal R^2]\le2Kr_K$ and $
 \mathbb E[\mathcal R]
 \le\sqrt{\mathbb E[\mathcal R^2]}
 \le\sqrt{2Kr_K}$.

Now,  suppose that the best target candidate is unique almost surely. On the event $\{\widehat j\ne j_\star\}$, the selected target is at
least the second-smallest target, and thus we have  $
 \mathcal R\ge\Delta$.
Therefore, for every $\gamma>0$, we have 
 $\{\widehat j\ne j_\star\}
 \subseteq
 \{\Delta\le\gamma\}\cup\{\mathcal R>\gamma\}$.
By the union bound and Markov's inequality, we have
\[
 \begin{aligned}
 \Pr\{\widehat j\ne j_\star\}
 &\le
 \Pr\{\Delta\le\gamma\}
 +\Pr\{\mathcal R>\gamma\}\\
 &\le
 \Pr\{\Delta\le\gamma\}
 +\frac{\mathbb E[\mathcal R^2]}{\gamma^2}\\
 &\le
 \Pr\{\Delta\le\gamma\}
 +\frac{2Kr_K}{\gamma^2}.
 \end{aligned}
\]
\end{proof}

We analyze a two-stage use of the metanetwork for stitching-candidate selection. First, predicted relation values determine a list of
candidate pairs. Second, adapters are fitted for the
retained candidates, and an independent validation sample is selected among the resulting stitched models. Model stitching assesses functional
compatibility through the performance of a fitted composition \citep{bansal2021revisiting,csiszarik2021similarity}.
An activation-based relation is a screening surrogate and thus transferring a distance guarantee to stitching requires a separate surrogate-fidelity assumption.

Let $\mathcal C$ be a nonempty finite candidate bank. Fix a candidate-wise adapter-fitting procedure, its training data, and its random seeds. For each $c\in\mathcal C$, let $f_c$ denote the stitched model produced by that procedure and define its population risk under the task law $\mathsf D$ by
\begin{equation}
 R_c=\int\ell(f_c(x),y)\,d\mathsf D(x,y).
\end{equation}
For candidates not fitted in deployment, $f_c$ is the potential output of the same specified procedure. Thus, the reference oracle is over these procedure-specific fitted models, not over all possible adapters. Let $\mathcal F_0$ collect all pre-validation information, including these potential fitted models, the metanetwork outputs, and the list.

\begin{theorem}
Assume $0\le\ell\le B_\ell<\infty$.
Let $\mathcal S\subseteq\mathcal C$ be a nonempty,
$\mathcal F_0$-measurable list, and write $s=|\mathcal S|$.
Let $
 \mathcal V_n=((X_i,Y_i))_{i=1}^n\sim\mathsf D^n$ for $n\ge1$ 
be independent of $\mathcal F_0$.
Specifically, this validation sample is not used to construct the list or to fit or otherwise adapt the candidate models.
Define
\begin{equation}
 \widehat R_c=\frac1n\sum_{i=1}^n\ell(f_c(X_i),Y_i),
 \qquad
 \widehat c\in\arg\min_{c\in\mathcal S}\widehat R_c,
\end{equation}
using measurable tie-breaking, and let $
 R_\star=\min_{c\in\mathcal C}R_c$ and  
 $B_{\rm screen}=\min_{c\in\mathcal S}R_c-R_\star$.
For $0<\alpha<1$, let $
 a_n(s,\alpha)
 =B_\ell\sqrt{\frac{\log(2s/\alpha)}{2n}}$.
Then, conditionally on $\mathcal F_0$, with probability at least
$1-\alpha$ over $\mathcal V_n$, we have 
\begin{equation}
 R_{\widehat c}-R_\star
 \le B_{\rm screen}+2a_n(s,\alpha) \quad \text{ and } \quad
 \mathbb E[R_{\widehat c}-R_\star\mid\mathcal F_0]
 \le B_{\rm screen}
    +B_\ell\sqrt{\frac{\log s}{2n}}.
\end{equation}

Suppose additionally that $\delta_c$ and $\widehat\delta_c$ are,
respectively, the true and predicted relation values used for
screening, all measurable with respect to $\mathcal F_0$.
Let $F:I\to\mathbb R$ be nondecreasing and $L_F$-Lipschitz on an
interval $I$ containing every $\delta_c$, and assume that $\max_{c\in\mathcal C}|R_c-F(\delta_c)|\le\zeta$
for some $\zeta\ge0$.
Define the relation-screening regret
$A_\delta(\mathcal S)
 =\min_{c\in\mathcal S}\delta_c-\min_{c\in\mathcal C}\delta_c$.
Then, deterministically conditional on $\mathcal F_0$, we have $
 B_{\rm screen}\le L_F A_\delta(\mathcal S)+2\zeta$.
Specifically, suppose that
$\max_{c\in\mathcal C}|\widehat\delta_c-\delta_c|\le E$
for some $E\ge0$. If $\mathcal S$ contains at least one minimizer
of $\widehat\delta_c$ over $\mathcal C$, then
$B_{\rm screen}\le2L_FE+2\zeta$.
If $\mathcal S$ contains a minimizer of $\delta_c$, then instead
$B_{\rm screen}\le2\zeta$. The latter condition is guaranteed by
\begin{equation}
 \mathcal S\supseteq\mathcal S_{\rm safe}(E)
 :=\left\{
   c\in\mathcal C:
   \widehat\delta_c\le
   \min_{d\in\mathcal C}\widehat\delta_d+2E
 \right\}.
\end{equation}
Therefore, a list containing a predicted
relation minimizer satisfies
$
 R_{\widehat c}-R_\star
 \le2L_FE+2\zeta+2a_n(s,\alpha)$.
Retaining the complete safe set instead gives
 $R_{\widehat c}-R_\star
 \le2\zeta+2a_n(s,\alpha)$.
\end{theorem}

\begin{proof}
Condition on $\mathcal F_0$. The list and all fitted models are now fixed, and the validation observations remain independent
with law $\mathsf D$. If $B_\ell=0$, the validation bounds are immediate and for their derivation below, assume $B_\ell>0$.

For a random variable $Z\in[0,B_\ell]$, the centered log moment generating function has second derivative
$\operatorname{Var}_t(Z)\le B_\ell^2/4$ under exponential tilting.
Since its value and first derivative vanish at zero, integration has $
 \mathbb E\exp(t(Z-\mathbb EZ))
 \le\exp(t^2B_\ell^2/8)$ for $t\in\mathbb R$.
Independence across validation examples and exponential Markov inequality give
\[
 \Pr\{|\widehat R_c-R_c|>a\mid\mathcal F_0\}
 \le2\exp(-2na^2/B_\ell^2),\qquad a>0.
\]
A union bound over the $s$ retained candidates has $
 \max_{c\in\mathcal S}|\widehat R_c-R_c|
 \le a_n(s,\alpha)$
with conditional probability at least $1-\alpha$.
For a fixed population minimizer
$c_{\mathcal S}\in\arg\min_{c\in\mathcal S}R_c$, we have 
\[
 R_{\widehat c}
 \le\widehat R_{\widehat c}+a_n(s,\alpha)\le\widehat R_{c_{\mathcal S}}+a_n(s,\alpha)
 \le R_{c_{\mathcal S}}+2a_n(s,\alpha).
\]

For the expectation bound, let $
 Z_{\max}=\max_{c\in\mathcal S}(R_c-\widehat R_c)$. 
By the empirical optimality, we have $
 R_{\widehat c}-R_{c_{\mathcal S}}
 \le Z_{\max}+\widehat R_{c_{\mathcal S}}-R_{c_{\mathcal S}}.$
The difference $\widehat R_{c_{\mathcal S}}-R_{c_{\mathcal S}}$ has zero conditional mean.
The same exponential-moment bound gives, for $t>0$, we have 
\[
 \mathbb E[Z_{\max}\mid\mathcal F_0]
 \le\frac1t\log\left(
   \sum_{c\in\mathcal S}
   \mathbb E[e^{t(R_c-\widehat R_c)}\mid\mathcal F_0]
 \right)\le\frac{\log s}{t}+\frac{tB_\ell^2}{8n}.
\]
For $s>1$, optimizing over $t$ gives
$B_\ell\sqrt{\log s/(2n)}$.
For $s=1$, selection adds no error and the same bound is zero.

Now, we choose $
 c_{\mathcal S,\delta}\in\arg\min_{c\in\mathcal S}\delta_c$,  
 $c_\delta\in\arg\min_{c\in\mathcal C}\delta_c$,
and 
 $c_R\in\arg\min_{c\in\mathcal C}R_c$, then we have 
\[
 \begin{aligned}
 \min_{c\in\mathcal S}R_c
 &\le R_{c_{\mathcal S,\delta}}\\
 &\le F(\delta_{c_{\mathcal S,\delta}})+\zeta\\
 &\le F(\delta_{c_\delta})
      +L_FA_\delta(\mathcal S)+\zeta\\
 &\le F(\delta_{c_R})+L_FA_\delta(\mathcal S)+\zeta\\
 &\le R_\star+L_FA_\delta(\mathcal S)+2\zeta.
 \end{aligned}
\]

If $\widetilde c\in\mathcal S$ minimizes the predicted relation
value over $\mathcal C$, then
\[
 \delta_{\widetilde c}
 \le\widehat\delta_{\widetilde c}+E
 \le\widehat\delta_{c_\delta}+E
 \le\delta_{c_\delta}+2E.
\]
Thus $A_\delta(\mathcal S)\le2E$.
If $c_\delta\in\mathcal S$, then $A_\delta(\mathcal S)=0$,
giving $B_{\rm screen}\le2\zeta$.

For any true relation minimizer $c_\delta$ and any predicted minimizer $\widetilde c$, we have
\[
 \widehat\delta_{c_\delta}
 \le\delta_{c_\delta}+E
 \le\delta_{\widetilde c}+E
 \le\widehat\delta_{\widetilde c}+2E.
\]
Therefore every true relation minimizer lies in
$\mathcal S_{\rm safe}(E)$.
\end{proof}

\section{Experimental setup and details
}\label{app:relation_retrieval_setup}

\subsection{Base-model populations and training}
\label{app:relation_retrieval_shared}

\noindent\textbf{Datasets and model populations.}
We consider four dataset-architecture combinations: ResNets \citep{he2016deep, zagoruyko2016wide} and Vision Transformers (ViTs) \citep{dosovitskiy2021an} trained on CIFAR-10, and the same two architecture families trained on CIFAR-100.
For the CIFAR datasets, we use the standard data augmentation, consisting of random flips, random translations, and Cutout.
For the ResNet family, we vary both network depth and width. Specifically, we consider ResNet20, ResNet26, ResNet44, and ResNet56, each with width multipliers of $\times 1$, $\times 2$, $\times 4$, and $\times 8$. 
We train each model for 300 epochs using stochastic gradient descent (SGD) with a momentum of 0.9, a batch size of 128, an initial learning rate of 0.1, and a weight decay of $5\times10^{-4}$.
The learning rate is decayed according to a cosine schedule throughout training.
For the ViT family, we vary the number of transformer blocks, attention heads, and per-head embedding dimensions. Specifically, we consider models with 2, 3, and 6 transformer blocks, 2, 3, and 6 attention heads, and per-head dimensions of 16, 32, and 64.
We train each ViT for 300 epochs using AdamW with $(\beta_1,\beta_2)=(0.9,0.999)$, a batch size of 128, and a weight decay of 0.05. The learning rate is linearly increased from $10^{-6}$ to $3\times10^{-4}$ during the first 10 epochs and subsequently decayed to $10^{-6}$ using a cosine annealing schedule over the remaining epochs.
We construct a model population containing 2,000 ResNet models and 1,350 ViT models, trained across the considered architecture and dataset configurations. Training the complete model population requires approximately 658 GPU hours.

\noindent\textbf{Model population split.}
We partition models into disjoint meta-training, validation, and test populations into 50\%, 25\%, and 25\%, respectively, before constructing representation-site pairs. 
Site pairs are constructed only within their respective partitions.

\subsection{Representation sites}
\label{app:representation_sites}

A representation site consists of a block-output location and a
fixed rule for extracting one feature vector per reference image.
Sites are identified by their position in the forward computation.

\noindent\textbf{ResNet sites.}
We select the output of every complete residual block in the three-stage CIFAR ResNets. For a model of depth $d_m$, each stage contains $(d_m-2)/6$ two-convolution residual blocks. Thus, ResNet20, ResNet26, ResNet44, and ResNet56 contribute
9, 12, 21, and 27 sites, respectively. The width multiplier changes the channel dimensions but not the number of sites.
Each site is identified by its stage index and block index within that stage. For the post-activation blocks specified here, we
record the block output after the shortcut addition and final ReLU, rather than an intermediate convolution or normalization
output. We average the resulting feature map over its spatial dimensions, separately for each channel, to obtain one feature vector per image. Its dimension equals the output channel count of the selected block. Spatial positions are neither concatenated nor treated as separate reference examples. The initial stem, final pooling operation, and classifier are not additional sites.

\noindent\textbf{ViT sites.}
We select the output of every complete Transformer block.
The configurations with 2, 3, and 6 blocks have 2, 3, and 6 sites, respectively, independently of the number
of attention heads and the per-head attention dimension.
For the pre-layer-normalized blocks specified here, the selected tensor is the output after the MLP residual addition. The final block is observed before the separate final encoder normalization and classification head. At each block, we extract the class-token vector at token index zero. Its feature dimension is the residual-stream token embedding dimension, not the per-head attention dimension. Patch tokens are not pooled, concatenated, or treated as separate examples. The patch-embedding output, individual attention heads, and classifier logits are not additional sites.

\noindent\textbf{Reference matrices and preprocessing.} We compute representation relations using a shared reference sample $\mX_{\rm ref}=(\vx_i)_{i=1}^{n}$ of $n=1{,}000$ images. For each representation site $u$, we stack the extracted feature vectors into the activation matrix with one row per reference image and one column per feature coordinate. The reference images and their ordering are identical across all models and sites evaluated on the same reference sample. All activations are collected with the base models in evaluation mode. 
Site extraction applies only the spatial averaging specified for ResNets or the class-token selection specified for ViTs.
Any relation-specific centering, normalization, neighborhood construction, or fitted transformation is applied subsequently as part of the corresponding target-construction procedure (App.~\ref{app:relations}).
The validation is determined separately for each relation because the conditions under which a relation is well defined differ across relation types. For a given $\mathsf h$, all predictors are evaluated on the same set of valid site pairs. We report the number of excluded sites and site pairs for each relation. We do not remove reference examples independently for individual site pairs, since doing so would change the underlying reference sample across comparisons.
In addition, we use batch aggregation as in \cite{nguyen2021do, huh2024position}.

\subsection{Method construction 
}
\label{app:exp_details}
Our method's central point is that all learned estimators are supervised using labeled pairs. The terms independent embedding and joint comparison refer to how a labeled pair is processed by the metanetwork.
There are two levels of learning: \textit{(i)} each base model is trained in the ordinary way on its task, and \textit{(ii)} the base models are frozen and treated as \textit{data} for a metanetwork. One meta-training instance is constructed by the following sequence.
\begin{enumerate}[leftmargin=1.75em,itemsep=2pt,topsep=0pt, itemsep=0pt, parsep=0pt, partopsep=0pt]
\item Select two frozen base models and a representation site in each.
\item Extract the architecture and trained parameters of the computation upstream of each site.
\item Run the frozen base models on the data examples and compute the activation-defined target distances.
\item Present the two parameterized computations and the target distances to the metanetwork loss.
\end{enumerate}

\subsubsection{A two-layer example}

Consider the two-layer linear network \begin{equation} \vh = \mW_1\vx, \qquad \vy = \mW_2\vh, \end{equation} where $\vx\in\mathbb{R}^{d}$, $\vh\in\mathbb{R}^{p}$, $\vy\in\mathbb{R}^{q}$, $\mW_1\in\mathbb{R}^{p\times d}$, and $\mW_2\in\mathbb{R}^{q\times p}$. The hidden activation $\vh$ defines one representation site, say $u$. Its upstream computation $\mathcal C_u$ contains the input interface and the parameterized map $\mW_1$. The output $\vy$ defines a second site, say $v$, whose upstream computation $\mathcal C_v$ contains both linear maps and implements the composed transformation $\mW_2\mW_1$. Now suppose that site $u$ is compared with a representation site $u'$ from another independently trained and frozen network. Evaluating the two sites on the same reference sample $\mX$ produces activation matrices $\mH_u(\mX)$ and $\mH_{u'}(\mX)$. An activation-derived relation $\delta_{\mathsf h}$, such as CKA, then defines the target \begin{equation} y_{uu'}^{(\mathsf h)} = \delta_{\mathsf h} \left( \mH_u(\mX), \mH_{u'}(\mX) \right). \end{equation} The resulting meta-learning example is $(\mathcal C_u,\mathcal C_{u'},y_{uu'}^{(\mathsf h)})$. In this simplified setting, $\mathsf{PARIS}$-Ind processes the two upstream computations independently to obtain reusable site embeddings, \begin{equation} \vz_u = \operatorname{Enc}_{\phi}(\mathcal C_u), \qquad \vz_{u'} = \operatorname{Enc}_{\phi}(\mathcal C_{u'}), \end{equation} and predicts their relation from a relation-specific distance between these embeddings. By contrast, $\mathsf{PARIS}$-Joint processes $\mathcal C_u$ and $\mathcal C_{u'}$ jointly, allowing information from the two parameterized computations to interact before the relation is predicted. Thus, this interaction can capture pair-dependent relationships between the two parameter matrices that cannot be represented by processing each site independently before comparison.

\subsubsection{Computation graph}

For each representation site $u=(m,s)$, we have
$\mathcal C_u
=
\mathcal C_m[
\operatorname{Anc}_{\mathcal C_m}(u)\cup{u}
]$.
This graph includes trained parameters and state used during inference. We encode this computation into a graph
\begin{equation}
G_u=\Psi(\mathcal C_u)
=(V_u,E_u,\mX_u,\mA_u,t_u),
\end{equation}
where $V_u$ and $E_u$ are nodes and edges, $\mX_u$ and $\mA_u$ store their attributes, and $t_u$ marks the target coordinates.
Following \citet{lim2024graph}, we attach each numerical parameter to the coordinates it connects instead of treating a whole parameter tensor as one feature vector. We also encode operator type, inference state, and the rule used to extract the target representation. The graph describes the trained model, and its activations are not included for graph features.

\noindent\textbf{Graph structure.}
Coordinate nodes represent feature or channel dimensions at tensor boundaries. Operator nodes represent computations such as linear maps, convolutions, normalization, pooling, and nonlinearities. We add position, constant, attention-head, or normalization-group nodes only when the computation requires them.
Edges describe both structure and numerical parameters. For example, a linear edge carries one weight, a convolution edge carries one kernel value and its offset, and bias edges connect a constant node to output coordinates. We also use structural edges to preserve input--output relations, residual connections, coordinate mappings, and normalization groups.
We keep the inference state that affects the output. For example, BatchNorm includes its learned scale and bias, running mean and variance, and $\epsilon$. We do not include training-only state that has no effect during evaluation.

\noindent\textbf{Attention.}
For attention, we represent each head through its query-key and value-output circuits:
\begin{equation}
\mC_a^{\mathrm{QK}}
=
\mW_{Q,a}\mW_{K,a}^{\top},
\qquad
\mC_a^{\mathrm{VO}}
=
\mW_{V,a}\mW_{O,a}.
\end{equation}

We store these circuit values directly on graph edges and group the QK and VO circuits by attention head. This keeps their functional relation.

\noindent\textbf{Target readout and message passing.}
We encode the rule used to extract the target representation as a final readout operation. This rule can select, for example, the full tensor, a class token, or an average over tokens or spatial positions. We mark only the output coordinates of this readout as targets.
After $R$ message-passing rounds, we combine the mean target state with the mean state of the full graph.
We add a reverse message-passing edge for every forward edge. Reverse edges share the same numerical attributes but have a separate direction flag. They support information flow in both directions and do not represent inverse operations.
Finally, we normalize numerical attributes using statistics from the meta-training graphs only. We keep all represented parameters, including zeros, and do not sample or prune nodes or edges.

\subsubsection{Architecture}

We use the same fixed-width graph backbone for $\mathsf{PARIS}$-Ind and
$\mathsf{PARIS}$-Joint.
They share the site-graph schema and architecture, not parameter values.
The
backbone uses node width $d=128$, message width $128$, and six
message-passing rounds.
Each categorical field has an embedding width of $8$.
The node input MLP has dimensions $d_x\to128\to128$, each
round-specific edge-gate MLP has dimensions
$d_e\to128\to128$, and each node-update MLP has dimensions
$384\to256\to128$.
We use GELU nonlinearities, concatenate the sum and mean of incoming
messages, and apply LayerNorm after each residual node update.
Different message-passing rounds use different parameters.
We use no dropout and do not sample nodes, edges, or tokens.

For $\mathsf{PARIS}$-Ind, the readout has dimensions
$256\to256\to128$ and produces a persistent site embedding
$\vz_u\in\mathbb R^{128}$.
Each comparison rule $h\in\mathcal H$ uses a bias-free metric head
$\mW_h\in\mathbb R^{64\times128}$ followed by the monotone calibration
$1-\exp(-\varepsilon_h d)$.

For $\mathsf{PARIS}$-Joint, the graph backbone uses width $128$.
The comparator keeps operator tokens, attention-head tokens, and two
summary tokens per site.
It uses two bidirectional cross-attention blocks with four heads each,
model width $128$, and per-head width $32$.
Each cross-attention feed-forward network has dimensions
$128\to512\to128$.
The contextual readout has dimensions
$256\to256\to128$, and the final symmetric prediction head has
dimensions $384\to256\to H$.
The joint estimator uses a sigmoid output rather than the metric heads
and exponential calibrations of the independent estimator.
These widths do not depend on the depth, width, or number of attention
heads of the represented base model, so the same metanetwork can process
the variable-size ResNet and ViT site graphs considered in our
experiments.

\noindent\textbf{Input attributes.}
Each categorical node or edge field has its own learned lookup table of
width $8$.
If a field does not apply, it contributes a fixed zero vector.
Let $F_x$ and $F_e$ denote the numbers of categorical node and edge
fields, and let $n_x$ and $n_e$ count the normalized numerical values,
flags, and presence masks.
The resulting node and edge widths are
\begin{equation}
    d_x=n_x+8F_x,
    \qquad
    d_e=n_e+8F_e.
    \label{eq:paris_attribute_widths}
\end{equation}
We fix the field vocabularies, their ordering, and all normalization
constants from the feature manifest.
We do not fit vocabularies or normalization statistics on validation or
test graphs.
The manifest determines the actual values of $d_x$, $d_e$,
and the vocabulary sizes.

For a node feature vector $\bm x_i\in\mathbb R^{d_x}$, we initialize
\begin{equation}
    \bm h_i^{(0)}
    =
    \operatorname{LN}_{\mathrm{in}}\left(
        \mA_{\mathrm{in},2}\,
        \operatorname{GELU}\left(
            \mA_{\mathrm{in},1}\bm x_i+\vb_{\mathrm{in},1}
        \right)
        +\vb_{\mathrm{in},2}
    \right),
    \label{eq:paris_initial_node_state}
\end{equation}
where
$\mA_{\mathrm{in},1}\in\mathbb R^{d\times d_x}$ and
$\mA_{\mathrm{in},2}\in\mathbb R^{d\times d}$.
All node types use the same MLP, and their types enter through the input
attributes.

\noindent\textbf{Message-passing rounds.}
We use $R=6$.
For a directed edge $e$ with source $s(e)$, target $t(e)$, and
attributes $\bm a_e\in\mathbb R^{d_e}$, round $r$ computes
\begin{align}
    \bm g_e^{(r)}
    &=
    \mB_{e,2}^{(r)}
    \operatorname{GELU}\left(
        \mB_{e,1}^{(r)}\bm a_e+\vb_{e,1}^{(r)}
    \right)
    +\vb_{e,2}^{(r)},
    \label{eq:paris_edge_gate}\\
    \bm m_e^{(r)}
    &=
    \left(W^{(r)}\bm h_{s(e)}^{(r)}\right)
    \odot
    \bm g_e^{(r)}.
    \label{eq:paris_round_message}
\end{align}
Here,
$\mB_{e,1}^{(r)}\in\mathbb R^{d\times d_e}$,
$\mB_{e,2}^{(r)}\in\mathbb R^{d\times d}$, and
$\mW^{(r)}\in\mathbb R^{d\times d}$.
The source projection $\mW^{(r)}$ has no bias.
The edge MLP has biases in both affine layers, no normalization, and no
final sigmoid.
Its output is a signed gate.
Edge attributes remain fixed across rounds, and we do not maintain a
separate recurrent edge state.
For node $j$, let $\mathcal E^{-}(j)$ denote the multiset of incoming
edges, including reverse and parallel edges.
We compute both their sum and mean:
\begin{equation}
    \bm s_j^{(r)}
    =
    \sum_{e\in\mathcal E^{-}(j)}
    \bm m_e^{(r)},
    \qquad
    \bar{\bm m}_j^{(r)}
    =
    \begin{cases}
        \bm s_j^{(r)}/|\mathcal E^{-}(j)|,
        & |\mathcal E^{-}(j)|>0,\\
        \bm 0,
        & |\mathcal E^{-}(j)|=0.
    \end{cases}
    \label{eq:paris_sum_mean}
\end{equation}
We concatenate the current node state with both aggregates and update
\begin{align}
    \Delta\bm h_j^{(r)}
    &=
    B_{n,2}^{(r)}
    \operatorname{GELU}\left(
        \mB_{n,1}^{(r)}
        \left[
            \bm h_j^{(r)}
            \Vert
            \bm s_j^{(r)}
            \Vert
            \bar{\bm m}_j^{(r)}
        \right]
        +\vb_{n,1}^{(r)}
    \right)
    +\vb_{n,2}^{(r)},
    \label{eq:paris_node_increment}\\
    \bm h_j^{(r+1)}
    &=
    \operatorname{LN}_{r}\left(
        \bm h_j^{(r)}+\Delta\bm h_j^{(r)}
    \right),
    \label{eq:paris_node_update}
\end{align}
where
$\mB_{n,1}^{(r)}\in\mathbb R^{2d\times3d}$ and
$\mB_{n,2}^{(r)}\in\mathbb R^{d\times2d}$.
Each round has its own source projection, edge MLP, node MLP, and
LayerNorm.
Within a round, these parameters are shared across all graph locations
and all site graphs.
The categorical lookup tables and the input encoder are shared across
rounds.

LayerNorm acts on the $d$ features of each node independently:
\begin{equation}
    \operatorname{LN}(\bm a)
    =
    \bm\gamma\odot
    \frac{
        \bm a-\mu(\bm a)\bm 1
    }{
        \sqrt{
            d^{-1}\sum_{j=1}^{d}
            (a_j-\mu(\bm a))^2
            +10^{-5}
        }
    }
    +\bm\beta,
    \qquad
    \mu(\bm a)
    =
    d^{-1}\sum_{j=1}^{d}a_j.
    \label{eq:paris_layernorm}
\end{equation}

\noindent\textbf{Pooling and site embedding.}
Let $S_u\subseteq V_u$ be the nonempty set of coordinate nodes marked by
the target readout.
After the final message-passing round, we compute
\begin{equation}
    \bm p_u
    =
    \frac{1}{|S_u|}
    \sum_{i\in S_u}\bm h_i^{(R)},
    \qquad
    \bm g_u
    =
    \frac{1}{|V_u|}
    \sum_{i\in V_u}\bm h_i^{(R)}.
    \label{eq:paris_ind_pools}
\end{equation}
The whole-graph mean includes every node type and excludes padding.
The target mean uses only the marked target-coordinate nodes.

The independent encoder maps the two pooled vectors to
\begin{equation}
    \vz_u
    =
    \mA_{\rho,2}
    \operatorname{GELU}\left(
        \mA_{\rho,1}
        [\bm p_u\Vert\bm g_u]
        +b_{\rho,1}
    \right)
    +b_{\rho,2}
    \in\mathbb R^k,
    \label{eq:paris_ind_embedding}
\end{equation}
with dimensions $2d\to2d\to k$, i.e.,
$256\to256\to128$.

For each comparison rule $h\in\mathcal H$, a bias-free matrix
$\mW_h\in\mathbb R^{r_h\times k}$ with $r_h=64$ defines
\begin{equation}
    d_{\phi,h}(u,v)
    =
    \left\|
        \mW_h(\vz_u-\vz_v)
    \right\|_2.
    \label{eq:paris_ind_metric}
\end{equation}
We optimize $\mW_h$ directly and do not constrain its positivity,
orthogonality, spectral norm, or Frobenius norm.
The induced matrix
$\mM_h=\mW_h^\top\mW_h$
is positive semidefinite by construction and has rank at most $64$.
Thus, $d_{\phi,h}$ is a pseudometric: it is symmetric, satisfies the
triangle inequality, and gives
$d_{\phi,h}(u,u)=0$, but two distinct sites may still have zero
distance.
We use the ordinary Euclidean norm and a zero subgradient at the zero
vector.

We map the latent distance to the target dissimilarity scale using
\begin{align}
    \widehat y_{uv}^{(h)}
    &=
    1-\exp\left[
        -\varepsilon_h d_{\phi,h}(u,v)
    \right],
    \label{eq:paris_ind_calibration}\\
    \varepsilon_h
    &=
    \operatorname{softplus}(\theta_h)
    +\varepsilon_{\min},
    \qquad
    \varepsilon_{\min}=10^{-6},
    \label{eq:paris_positive_scale}
\end{align}
where
$\operatorname{softplus}(t)=\log(1+e^t)$.
Each comparison rule has one learned scalar $\theta_h$.
We initialize $\varepsilon_h=1$.
We evaluate it numerically as
$-\operatorname{expm1}(-\varepsilon_h d_{\phi,h})$
with a stable softplus.

Let $\mathcal B$ be one batch of supervised site pairs and
define the set of unique sites
\begin{equation}
    \mathcal U_{\mathcal B}
    =
    \operatorname{unique}\left(
        \{u:(u,v)\in\mathcal B\}
        \cup
        \{v:(u,v)\in\mathcal B\}
    \right),
    \qquad
    M=|\mathcal U_{\mathcal B}|.
    \label{eq:paris_unique_batch_sites}
\end{equation}

For each dataset-architecture setting, we train
$\mathsf{PARIS}$-Ind and $\mathsf{PARIS}$-Joint on the same sequence of
sampled batches.
Each batch $\mathcal B$ contains $B=64$ unordered pairs of distinct
sites, and all sites come from the meta-training partition.
For $M\geq2$, let
\begin{equation}
    \mu_j
    =
    \frac1M
    \sum_{u\in\mathcal U_{\mathcal B}}z_{u,j},
    \qquad
    \sigma_j
    =
    \sqrt{
        \frac1M
        \sum_{u\in\mathcal U_{\mathcal B}}
        (z_{u,j}-\mu_j)^2
    }.
    \label{eq:paris_unique_site_std}
\end{equation}
We use
\begin{equation}
    \mathcal L_{\mathrm{var}}
    =
    \max\left(
        0,
        \eta-\frac1k\sum_{j=1}^{k}\sigma_j
    \right).
    \label{eq:paris_deduplicated_variance_loss}
\end{equation}
The standard deviation uses denominator $M$, not $M-1$.
When $M<2$, we set $\mathcal L_{\mathrm{var}}=0$.
We compute the statistic from the unnormalized embeddings $\vz_u$.
It discourages global collapse.

The joint estimator $\mathsf{PARIS}$-Joint first applies the same $d=128$, $R=6$
message-passing architecture to $G_u$ and $G_v$ separately.
To keep the pairwise comparator tractable, we do not expose every
coordinate node as a cross-attention token.
Let $\mathcal O_u$ and $\mathcal A_u$ denote the operator and
attention-head nodes of $G_u$.
Before adding role embeddings, we use
\begin{equation}
    \mathcal T_u^{\mathrm{raw}}
    =
    \{\bm p_u\}_{\mathrm{target}}
    \uplus
    \{\bm g_u\}_{\mathrm{global}}
    \uplus
    \{\bm h_i^{(R)}:i\in\mathcal O_u\}
    \uplus
    \{\bm h_i^{(R)}:i\in\mathcal A_u\}.
    \label{eq:paris_joint_token_set}
\end{equation}
Each token receives a learned width-$128$ role embedding that marks one
of four roles: target summary, global summary, operator, or attention
head.
The resulting number of tokens is
\begin{equation}
    n_u
    =
    |\mathcal O_u|
    +
    |\mathcal A_u|
    +
    2,
    \qquad
    T_u^{(0)}
    \in
    \mathbb R^{n_u\times128}.
    \label{eq:paris_joint_token_count}
\end{equation}
For ResNet sites, $|\mathcal A_u|=0$.
For a ViT site after block $\ell_u$,
\begin{equation}
    |\mathcal A_u|
    =
    \sum_{\ell=1}^{\ell_u}
    A_{\ell,\mathrm{base}}.
    \label{eq:paris_population_head_token_count}
\end{equation}
For a uniform-head ViT this becomes
$\ell_uA_{\mathrm{base}}$.
In the stated ViT population, the largest complete prefix
contains at most $6\times6=36$ attention-head tokens.

\noindent\textbf{Bidirectional cross-attention.}
We use $L_{\times}=2$ cross-attention blocks.
Each block has $a_{\times}=4$ comparator heads, model width $128$, and
per-head query/key/value width $32$.

For block $\ell$, we first normalize the two token streams,
\begin{equation}
    \mN_u^{(\ell)}
    =
    \operatorname{LN}_{A,\ell}(\mT_u^{(\ell)}),
    \qquad
    \mN_v^{(\ell)}
    =
    \operatorname{LN}_{A,\ell}(\mT_v^{(\ell)}).
\end{equation}
For comparator head $b$, the update from $v$ to $u$ is
\begin{equation}
    \operatorname{Att}_{b,\ell}(\mN_u,\mN_v)
    =
    \operatorname{softmax}\left(
        \frac{
            Q_{b,\ell}(\mN_u)
            K_{b,\ell}(\mN_v)^\top
        }{
            \sqrt{32}
        }
        +
        \mM_{u\leftarrow v}
    \right)
    V_{b,\ell}(\mN_v),
    \label{eq:paris_joint_head_attention}
\end{equation}
where the query, key, and value projections are affine and
$\mM_{u\leftarrow v}$ masks padded partner tokens.
The two directions are updated by
\begin{align}
    \bar \mT_u^{(\ell)}
    &=
    \mT_u^{(\ell)}
    +
    \operatorname{MHA}_{\ell}
    \left(
        \mN_u^{(\ell)},
        \mN_v^{(\ell)}
    \right),\\
    \bar \mT_v^{(\ell)}
    &=
    \mT_v^{(\ell)}
    +
    \operatorname{MHA}_{\ell}
    \left(
        \mN_v^{(\ell)},
        \mN_u^{(\ell)}
    \right),
    \label{eq:paris_joint_synchronous_crossattention}\\
    \mT_u^{(\ell+1)}
    &=
    \bar \mT_u^{(\ell)}
    +
    \operatorname{FFN}_{\ell}\left(
        \operatorname{LN}_{F,\ell}
        (\bar \mT_u^{(\ell)})
    \right),\\
    \mT_v^{(\ell+1)}
    &=
    \bar \mT_v^{(\ell)}
    +
    \operatorname{FFN}_{\ell}\left(
        \operatorname{LN}_{F,\ell}
        (\bar \mT_v^{(\ell)})
    \right).
    \label{eq:paris_joint_feedforward}
\end{align}
The feed-forward network has dimensions
$128\to512\to128$, uses GELU after the first affine layer, and
has biases in both layers.
We use no token self-attention inside the comparator.
Each sublayer uses pre-LayerNorm with learned gain and bias and
$\epsilon=10^{-5}$.
After the second cross-attention block, we apply one final shared
LayerNorm.

Let
$
    F_{u\mid v}
    =
    \operatorname{LN}_{\mathrm{out}}
    \left(
        \mT_u^{(L_{\times})}
    \right).$
We concatenate the final target-summary token with the mean of all valid
tokens and compute
\begin{equation}
    \bm c_{u\mid v}
    =
    \rho_J\left(
        F_{u\mid v,\mathrm{target}}
        \Vert
        \frac1{n_u}
        \sum_{t=1}^{n_u}
        F_{u\mid v,t}
    \right)
    \in
    \mathbb R^{128}.
    \label{eq:paris_joint_contextual_readout}
\end{equation}
The readout $\rho_J$ has dimensions
$256\to256\to128$ with one intermediate GELU.
It is shared between the two sides.
Let
$\bm c=\bm c_{u\mid v}$
and
$\bm c'=\bm c_{v\mid u}$.
We form the symmetric pair feature
\begin{equation}
    \bm s_{uv}
    =
    \left[
        \bm c+\bm c'
        \;\Vert\;
        |\bm c-\bm c'|
        \;\Vert\;
        \bm c\odot\bm c'
    \right]
    \in
    \mathbb R^{384}.
\end{equation}
The final predictor is
\begin{equation}
    \widehat{\bm y}_{uv}
    =
    \operatorname{sigmoid}\left(
        \mA_{J,2}
        \operatorname{GELU}\left(
            \mA_{J,1}\bm s_{uv}+\vb_{J,1}
        \right)
        +\vb_{J,2}
    \right)
    \in
    (0,1)^H,
    \label{eq:paris_joint_prediction_head}
\end{equation}
where
$\mA_{J,1}\in\mathbb R^{256\times384}$ and
$\mA_{J,2}\in\mathbb R^{H\times256}$.
The $h$-th coordinate is the predicted score
$\widehat y_{uv}^{(h)}$ for comparison rule $h$.

\paragraph{Initialization.}
Except for the metric heads and calibration scalars described above, we
initialize affine weight matrices with Xavier-uniform initialization
with gain one \citep{glorot2010understanding} and set affine biases to
zero.
We initialize categorical embeddings and joint token-role embeddings
independently from $\mathcal N(0,0.02^2)$.
LayerNorm gains start at one and biases at zero.

\noindent\textbf{Objectives and optimizer.} The objectives are
\begin{align}
 \mathcal L_{\mathrm{Ind}}
 &=\frac1H\sum_{h=1}^{H}
   \left(\mathcal L_{\mathrm{reg}}^{(h)}
          +\lambda_{\mathrm{rank}}\mathcal L_{\mathrm{rank}}^{(h)}\right)
    +\lambda_{\mathrm{var}}\mathcal L_{\mathrm{var}},\\
 \mathcal L_{\mathrm{Joint}}
 &=\frac1H\sum_{h=1}^{H}
   \left(\mathcal L_{\mathrm{reg}}^{(h)}
          +\lambda_{\mathrm{rank}}\mathcal L_{\mathrm{rank}}^{(h)}\right).
 \label{eq:paris_opt_total_loss}
\end{align}
We use $\tau=8$ and $\eta=0.04$ and $\lambda_{\mathrm{rank}} =0.14$ and $\lambda_{\mathrm{var}}=0.5$.
Both estimators use AdamW with weight decay coefficient $2\times10^{-4}$.
The initial learning rate is $2\times10^{-3}$ for $\mathsf{PARIS}$-Ind and
$1.5\times10^{-3}$ for $\mathsf{PARIS}$-Joint. After accumulating all gradients for one batch, we clip their global Euclidean norm to $5$ and perform one
optimizer step. 

\noindent\textbf{Trainable parameter counts.}
Let
\begin{equation}
    P_{\mathrm{cat}}
    =
    8\left(
        \sum_{f=1}^{F_x}K_{x,f}
        +
        \sum_{f=1}^{F_e}K_{e,f}
    \right)
\end{equation}
count the learned categorical lookup entries, excluding fixed zero rows
for inapplicable fields.
With the biases and affine LayerNorm parameters above, the shared graph
backbone contains
$
    P_B
    =
    P_{\mathrm{cat}}
    +
    d\,d_x
    +
    d^2
    +
    4d
    +
    R\left(
        10d^2+d\,d_e+7d
    \right).
$
The independent readout contains
$
    P_{\rho}
    =
    4d^2+2dk+2d+k,
$
so
$
    P_{\mathrm{Ind}}
    =
    P_B
    +
    P_{\rho}
    +
    \sum_{h=1}^{H}
    (r_hk+1).
$
The additional scalar for each comparison rule is its learned
calibration parameter.

For $\mathsf{PARIS}$-Joint, the four token-role embeddings add $4d$
parameters.
Each cross-attention block adds $12d^2+13d$ parameters from the four
attention projections, the $d\to4d\to d$ feed-forward network,
and two affine LayerNorms.
The final token LayerNorm adds $2d$, the contextual readout adds
$P_{\rho}$, and the symmetric prediction head adds
$6dk+2d+(2d+1)H$.
Thus,
\begin{equation}
    P_{\mathrm{Joint}}
    P_B
    +
    4d
    +
    L_{\times}(12d^2+13d)
    +
    2d
    +
    P_{\rho}
    +
    6dk
    +
    2d
    +
    (2d+1)H.
\label{eq:paris_joint_parameter_count}
\end{equation}

\noindent\textbf{Edge-block processing.}
We use float32 for metanetwork parameters,
numerical graph attributes, node states, and reductions.
Attention-circuit values and static graph topology may
remain in host memory and be transferred in directed-edge blocks of at
most
$C_E=65{,}536$.
For each block, we compute the categorical embeddings, edge gates, and
messages and accumulate them into the destination-node sums.
We compute the incoming degree over the complete graph, so the mean is
formed only after all edge blocks have contributed.
Edge blocking changes only the execution order.

We do not need to retain a complete
$|E_u|\times d$
array of gates or messages.
During training, we checkpoint message-passing rounds and recompute
edge-block intermediates during the backward pass.
Gradients from each block accumulate into the shared parameters and
source-node states.
Static graph attributes and categorical indices may be cached, but
learned edge gates are recomputed after every parameter update.

\subsubsection{Inference-time computational cost}
\label{app:inference-cost}
 
We compare inference costs after the metanetwork has been trained and frozen.
We compare two ways to obtain representation relations: $\mathsf{PARIS}$ predicts a
relation from trained computations, whereas direct evaluation measures the
relation from newly computed representations. 
 
We also include every computation needed to score a candidate
model that has not been processed before. For $\mathsf{PARIS}$, this includes graph
construction and encoding. For direct evaluation, it includes feature
extraction and processing required by the chosen relation. If a direct
relation fits a readout to a candidate model's features, we also include that
cost.

Fix one reference protocol and one relation $h$. Let $\mathcal{M}$ be the set
of distinct models, $\mathcal{U}$ the set of $N$ requested sites, and
$\mathcal{P}$ the set of $K$ distinct site pairs to score. Let $n$ be the
total number of distinct shared inputs required by the protocol. This count
includes inputs used to fit or tune a readout when needed. Let
$\mH_u\in\mathbb{R}^{n\times p_u}$ denote the features extracted at site $u$.
We count computation at fixed precision and assume conventional dense linear
algebra. 

\noindent\textbf{Direct evaluation from representations.}
Let $F_m$ denote the work per input required to run model $m$ far enough to
collect \emph{all requested sites}, including the work required by their
extraction rules. In a sequential model, one traversal to the deepest
requested site is sufficient. The model does not need a separate traversal
for each site. For fixed input shapes, collecting the representations costs
\begin{equation}
    \mathcal{F}(n)
    = n\sum_{m\in\mathcal{M}}F_m.
    \label{eq:cost-feature-extraction}
\end{equation}
We pay this cost once for each model and reference sample, not once for each
site pair. For an implementation that preprocesses each site, let $a_h(u)$
denote the work needed to prepare the comparison descriptor for site $u$
after feature extraction. Let $b_h(u,v)$ denote the work needed to score one
pair from the prepared quantities. Then
\begin{equation}
    C_{\mathrm{dir}}^{(h)}
    = \mathcal{F}(n)
    + \sum_{u\in\mathcal{U}}a_h(u)
    + \sum_{(u,v)\in\mathcal{P}}b_h(u,v).
    \label{eq:cost-direct-general}
\end{equation}
The preparation term includes any required centering, kernel construction,
neighbor search, or readout fitting and evaluation. For a readout-based
relation, feature extraction must also include the examples used to fit and
tune the readout, not only the examples used to measure disagreement.

For the independent estimator, write
\begin{equation}
    \vz_u=\operatorname{Enc}_{\phi}(G_u)\in\mathbb{R}^{k},
    \qquad
    \widetilde \vz_{u,h}=\mW_h\vz_u\in\mathbb{R}^{k_h}.
    \label{eq:cost-projected-embedding}
\end{equation}
Because the projection is linear,
$\|\mW_h(\vz_u-\vz_v)\|_2
=\|\widetilde \vz_{u,h}-\widetilde \vz_{v,h}\|_2$.
We can project each site once and reuse the result across pairs.
Let $\Gamma_{\mathcal{U}}$ denote the total work needed to build the required
graphs and compute all $\vz_u$, after reusing any common preprocessing that the
implementation supports. A dense projection costs $O(kk_h)$ per site. The
distance calculation and scalar calibration cost $O(k_h)$ per pair. Hence,
\begin{equation}
    C_{\mathrm{Ind}}^{(h)}
    = \Gamma_{\mathcal{U}}
    + O\left(Nkk_h+Kk_h\right).
    \label{eq:cost-independent}
\end{equation}
For one fixed relation, storing only the projected embeddings requires
$O(Nk_h)$ values. If we evaluate several relations, all of them share the
graph encoding, while their projection and scoring costs add across heads.
Direct methods can share feature extraction across relations.

Let $v_u=|V_u|$ and $e_u=|E_u|$, including reverse edges. Consider the stated
gated message-passing update with $R$ rounds, hidden dimension $d$, and
fixed-depth node-update MLPs with hidden widths of $O(d)$. At each round,
computing $Wh_i$ once for every node costs $O(v_ud^2)$. Computing edge gates
and aggregating messages costs $O(e_ud)$. Let $c_{\mathrm{edge}}$ denote the
total per-edge work of the distinct edge-MLP evaluations used across all
rounds. Encoding one site then costs
\begin{equation}
    \gamma_u
    = C_{\mathrm{aux},u}
    + O\left(
        e_uc_{\mathrm{edge}}
        + R(v_ud^2+e_ud)
      \right),
    \label{eq:cost-site-encoder}
\end{equation}
where $C_{\mathrm{aux},u}$ covers graph construction, attribute preprocessing,
node initialization, pooling, and the embedding readout. If the model uses
the same static edge gates across rounds, we can cache them. If it uses
different edge maps in different rounds, we must count each map separately.
Note that a dense width-$d$ hidden layer in the edge MLP costs $O(d^2)$
per edge for each distinct map.

\noindent\textbf{Comparison example.}
Let $\bar \mH_u$ denote the features after centering them over the $n$ inputs,
and assume that the centered Gram matrices are nonzero. In feature space,
\begin{equation}
    s_{\mathrm{CKA}}(u,v)
    = \frac{\|\bar \mH_u^{\top}\bar \mH_v\|_F^2}
    {\|\bar \mH_u^{\top}\bar \mH_u\|_F
     \|\bar \mH_v^{\top}\bar \mH_v\|_F}.
    \label{eq:cost-cka-feature-form}
\end{equation}
Computing each denominator costs $O(np_u^2)$. For each pair, computing the
cross-product costs $O(np_up_v)$. If the features are cached, the total cost
is
\begin{equation}
    C_{\mathrm{CKA,feat}}
    = \mathcal{F}(n)
    + O\left(
        n\sum_{u\in\mathcal{U}}p_u^2
        + n\sum_{(u,v)\in\mathcal{P}}p_up_v
      \right).
    \label{eq:cost-cka-features}
\end{equation}
We can instead prepare the normalized Gram descriptor
\begin{equation}
    \mB_u
    = \frac{\bar \mH_u\bar \mH_u^{\top}}
    {\|\bar \mH_u\bar \mH_u^{\top}\|_F},
    \qquad
    s_{\mathrm{CKA}}(u,v)=\langle \mB_u,\mB_v\rangle_F.
    \label{eq:cost-cka-gram-form}
\end{equation}
Constructing this descriptor costs $O(n^2p_u)$ per site, and comparing two
descriptors costs $O(n^2)$ per pair. Thus,
\begin{equation}
    C_{\mathrm{CKA,Gram}}
    = \mathcal{F}(n)
    + O\left(n^2\sum_{u\in\mathcal{U}}p_u+Kn^2\right).
    \label{eq:cost-cka-grams}
\end{equation}
We center the features with their means.
Converting the similarity to chordal dissimilarity
adds only constant work per pair. Both forms compute the same full-sample
relation. Which form is cheaper depends on $n$, the feature dimensions, the
number of requested pairs, and the available memory.

For one relation, we assume $p_u=p$ for simplicity. We also assume that no candidate descriptors are cached before preparation, so each method pays its preparation cost once. For linear CKA with feature caching, preparation costs $\mathcal{F}(n)+O(Nnp^2)$, each pairwise comparison costs $O(np^2)$, and the cache requires $O(Nnp)$ space. For linear CKA with Gram caching, preparation costs $\mathcal{F}(n)+O(Nn^2p)$, each pairwise comparison costs $O(n^2)$, and the cache requires $O(Nn^2)$ space. For $\mathsf{PARIS}$-Ind, preparation costs $\Gamma_{\mathcal{U}}+O(Nkk_h)$, each pairwise comparison costs $O(k_h)$, and the cache requires $O(Nk_h)$ space. 
They exclude model weights, encoder workspace, and the optional $O(K)$ space used to store output scores.

These costs apply to the stated full-sample implementations.  Consider a CKA estimator that uses one shared partition of the data
into batches of size at most $b$. If it builds and aggregates kernels only
within each batch, kernel construction costs $O(nbp_u)$ per site and
comparison costs $O(nb)$ per pair. These bounds exclude any extra work for
bandwidth selection or repeated partitions. The cost of direct
comparison depends on the relation. Cached sample-space GULP operators or
normalized RBF kernels allow $O(n^2)$ comparison. MutualNN neighbor sets
stored in example-index order allow $O(nk_{\mathrm{NN}})$ comparison. Cached
hard readout predictions allow $O(n_{\mathrm{eval}})$ disagreement
evaluation. We still include the cost of constructing these descriptors in
$a_h(u)$.

The joint estimator cannot reuse a final site embedding independently of its
comparison partner. Let $\Gamma_{\mathrm{Joint}}$ denote the reusable work for
graph and token preparation, and let $j_h(u,v)$ denote the remaining work for
pair $(u,v)$. Then
\begin{equation}
    C_{\mathrm{Joint}}^{(h)}
    = \Gamma_{\mathrm{Joint}}
    + \sum_{(u,v)\in\mathcal{P}}j_h(u,v).
    \label{eq:cost-joint}
\end{equation}
Suppose sites $u$ and $v$ contain $\tau_u$ and $\tau_v$ tokens. For $B$ dense
bidirectional cross-attention blocks with total hidden width $d_{\mathrm{J}}$
and feed-forward widths of $O(d_{\mathrm{J}})$, the cross-attention blocks
cost
\begin{equation}
    O\left(
       B\big[(\tau_u+\tau_v)d_{\mathrm{J}}^2
            +\tau_u\tau_vd_{\mathrm{J}}\big]
    \right)
    \label{eq:cost-cross-attention}
\end{equation}
per pair, in addition to the output head and any other work that depends on
the pair. Therefore, the $O(k_h)$ cached-scoring bound for $\mathsf{PARIS}$-Ind does not
apply to $\mathsf{PARIS}$-Joint. Avoiding evaluation on reference inputs is not, by
itself, enough to show that $\mathsf{PARIS}$-Joint scores pairs more cheaply than a
cached direct method.

Let $D_0$ and $I_0$ denote the
realized preparation costs of direct evaluation and $\mathsf{PARIS}$-Ind. Let
$\bar b_{\mathrm{dir}}$ and $\bar b_{\mathrm{Ind}}$ denote their average
per-pair scoring costs. $\mathsf{PARIS}$-Ind has lower total inference cost when $
    K\big(\bar b_{\mathrm{dir}}-\bar b_{\mathrm{Ind}}\big)
    > I_0-D_0$.
If $\bar b_{\mathrm{dir}}>\bar b_{\mathrm{Ind}}$, this condition is equivalent
to
$K>(I_0-D_0)/(\bar b_{\mathrm{dir}}-\bar b_{\mathrm{Ind}})$.

If candidate descriptors are already cached, we exclude their preparation
cost from the incremental comparison for \emph{both} methods. A new query
model still requires its own preparation and query--bank scoring, but we do
not process the cached candidate models again. Exhaustive all-pairs scoring
still requests $K=\Theta(N^2)$ scores under both methods. $\mathsf{PARIS}$-Ind changes
the cost of preparing each site and scoring each pair.

After training, $\mathsf{PARIS}$ does not need to evaluate a new candidate model on the
reference inputs. $\mathsf{PARIS}$-Ind then compares sites by computing distances in
$\mathbb{R}^{k_h}$. Its deployment cost has no explicit dependence
on the number of reference examples used to define the learned target.
$\mathsf{PARIS}$ has an inference-time advantage only when graph preparation,
encoding, and scoring cost less in total than the corresponding direct
pipeline. 

\subsubsection{Computation cost and practical efficiency}

Training the $\mathsf{PARIS}$-Ind and $\mathsf{PARIS}$-Joint metanetworks requires approximately 5.56 and 7.64 GPU hours, respectively.
These costs represent an upfront investment in learning to predict representation similarities from trained computations. Once trained, $\mathsf{PARIS}$ can estimate similarities between internal sites of previously unseen models without collecting new activations or retraining the metanetwork. 
Specifically, $\mathsf{PARIS}$-Ind produces reusable site embeddings, making repeated comparisons substantially cheaper after initial graph construction and encoding. Thus, the practical benefit depends on the intended workload: the training and preparation costs are substantial, but can be amortized when the trained metanetwork is reused across many models and comparisons. 

We evaluate the practical computational cost of $\mathsf{PARIS}$ using a representative workload of 200 representation sites drawn from 200 distinct base models. We compare $\mathsf{PARIS}$-Ind and $\mathsf{PARIS}$-Joint against direct activation-based evaluation with cached descriptors, distinguishing three stages: \textit{(i)} graph construction or feature acquisition, \textit{(ii)} site encoding or descriptor preparation, and \textit{(iii)} pairwise comparison.

For $\mathsf{PARIS}$-Ind, graph construction and site encoding require around 47 and 130 seconds, respectively, while $\mathsf{PARIS}$-Joint requires around 47 and 136 seconds. Direct activation-based evaluation requires around 30 seconds for feature acquisition and around 44 seconds for descriptor preparation. Thus, direct evaluation has lower initial preparation costs. However, once site representations are prepared, $\mathsf{PARIS}$-Ind evaluates representation relations in approximately 1.1 ms per pair, compared with 40.5 ms for direct evaluation, corresponding to an approximately 37-fold speedup. $\mathsf{PARIS}$-Joint requires approximately 265 ms per pair due to its pair-dependent cross-attention.

We additionally measure peak memory usage and quantify persistent descriptor storage. $\mathsf{PARIS}$-Ind produces a shared 128-dimensional site embedding and five 64-dimensional relation-specific projections. Storing these five float32 projections requires 0.256 MB. For $\mathsf{PARIS}$-Joint, caching 128 tokens per site requires approximately 13.1 MB. In contrast, the illustrative direct baseline stores three dense $1000\times1000$ float32 descriptors per site, with neighbor lists and held-out predictions, requiring approximately 2.408 GB.
The measured peak memory usage is 0.34 GB for $\mathsf{PARIS}$-Ind, 0.67 GB for $\mathsf{PARIS}$-Joint, and 3.82 GB for direct evaluation. Thus, $\mathsf{PARIS}$-Ind requires approximately 11 times less peak memory than the direct baseline in this workload, while also providing substantially more compact persistent site representations.

These measurements show that the computational advantage of $\mathsf{PARIS}$ depends on the workload. Direct activation-based evaluation can be preferable for one-off comparisons, when suitable probe data and cached descriptors are already available. In contrast, $\mathsf{PARIS}$-Ind incurs a one-time graph-construction and encoding cost but supports substantially faster repeated comparisons and compact persistent storage. $\mathsf{PARIS}$-Joint provides pair-dependent interactions at a higher computational cost.
Beyond these computational trade-offs, $\mathsf{PARIS}$ predicts representation dissimilarities for unseen models without requiring a probe dataset or additional base-model inference at query time. This also avoids variability associated with newly sampling probe examples. That is, the advantage is the ability to reuse information learned from reference similarity measurements across unseen models without collecting new activations.

\subsubsection{Relation to contrastive learning and supervised metric learning}

Classical contrastive metric learning uses a binary label that marks two
examples as similar or dissimilar and then attracts or repels their
embeddings, often with a margin for dissimilar pairs
\citep{hadsell2006dimensionality}.
Modern contrastive objectives such as InfoNCE and SimCLR define positive
views for an anchor and distinguish them from sampled negatives through a
normalized log-softmax loss
\citep{oord2018representation,chen2020simple}.
Supervised contrastive learning replaces augmentation-based positives by
examples with the same class label, but it still separates positives
from negatives \citep{khosla2020supervised}.

Our supervision is different.
For each comparison rule $h\in\mathcal H$ and site pair $(u,v)$, we
compute a continuous target dissimilarity
\begin{equation}
    y_{uv}^{(h)}
    =
    \Delta_h(\mH_u,\mH_v)
    \in[0,1].
    \label{eq:graded-reference-distance}
\end{equation}
For example, a pair with score $0.08$
should be closer than one with score $0.35$, which should in turn be
closer than one with score $0.82$.
Turning these values into binary positive and negative labels would
discard both their scale and much of their ordering.

We train directly on the continuous scores.
The regression term encourages the
predicted value $\widehat y_{uv}^{(h)}$ to match
$y_{uv}^{(h)}$, while the ranking term encourages the model to preserve
the ordering between two supervised site pairs.
The ranking loss does not define
one pair as positive and the remaining pairs as negatives.
Thus, $\mathsf{PARIS}$-Ind is more naturally viewed as supervised
metric learning with continuous target distances, or as distance
distillation.
$\mathsf{PARIS}$-Joint is even less similar to standard contrastive
learning: it predicts the score for a pair directly and does not need
to embed every site into one shared metric space.

These differences also clarify the role of the output geometry.
Binary contrastive learning, InfoNCE, and supervised contrastive
learning primarily specify which examples should be close or far apart. 
In contrast, our objective uses the full target dissimilarity
$y_{uv}^{(h)}$.
For $\mathsf{PARIS}$-Ind, this supervision is realized through the
pseudometric
$d_{\phi,h}(u,v)$
and its monotone calibration.
For $\mathsf{PARIS}$-Joint, the model predicts
$\widehat y_{uv}^{(h)}$
with a pair-dependent comparator, without imposing a common metric
geometry.

A contrastive variant of our method could potentially be defined.
For a query site $u$ and a candidate set $\mathcal V_u$, the target
scores can define a soft neighborhood distribution
\begin{equation}
    p_h(v\mid u)
    =
    \frac{
        \exp\!\left[-y_{uv}^{(h)}/T_y\right]
    }{
        \sum_{w\in\mathcal V_u}
        \exp\!\left[-y_{uw}^{(h)}/T_y\right]
    }.
    \label{eq:reference-neighborhood}
\end{equation}
For $\mathsf{PARIS}$-Ind, the learned distances can define
\begin{equation}
    q_{\phi,h}(v\mid u)
    =
    \frac{
        \exp\!\left[-d_{\phi,h}(u,v)/T_z\right]
    }{
        \sum_{w\in\mathcal V_u}
        \exp\!\left[-d_{\phi,h}(u,w)/T_z\right]
    }.
    \label{eq:predicted-neighborhood}
\end{equation}
One could then minimize
\begin{equation}
    \mathcal L_{\mathrm{softcon}}^{(h)}
    =
    \mathbb E_u\,
    \operatorname{KL}\!\left(
        p_h(\cdot\mid u)
        \,\|\, 
        q_{\phi,h}(\cdot\mid u)
    \right).
    \label{eq:soft-contrastive}
\end{equation}
This objective would use normalized neighborhoods in a contrastive
manner while preserving graded similarity through soft weights.
It could serve as an auxiliary retrieval objective, and we leave it for future exploration.
Unlike direct regression to $y_{uv}^{(h)}$, this objective mainly fixes
relative neighborhood structure and does not directly recover the
absolute target scale.

Verified reparameterizations provide another possible source of
positive pairs.
If a transformation $\gamma$ leaves the computation unchanged, then
$G_u$ and $\gamma G_u$ describe equivalent computations.
For the hidden-coordinate, feed-forward-coordinate, and complete-head
permutations represented by our graph construction, however,
$\mathsf{PARIS}$ is already invariant by construction.
Likewise, the supported query--key and value--output changes of basis
are removed by the paired attention encoding before the metanetwork
processes the graph.
Using these transformations as contrastive augmentations would
be unnecessary for enforcing the invariances that our architecture
already guarantees.
Contrastive augmentation could still be useful for other
symmetries that are not built into the current graph representation, but
such invariance would be learned empirically.

\subsubsection{Whole-network encoder with site-specific outputs}
\label{app:whole_network_encoder}

\begin{figure*}[t]
    \centering
     \includegraphics[width=0.75\textwidth]{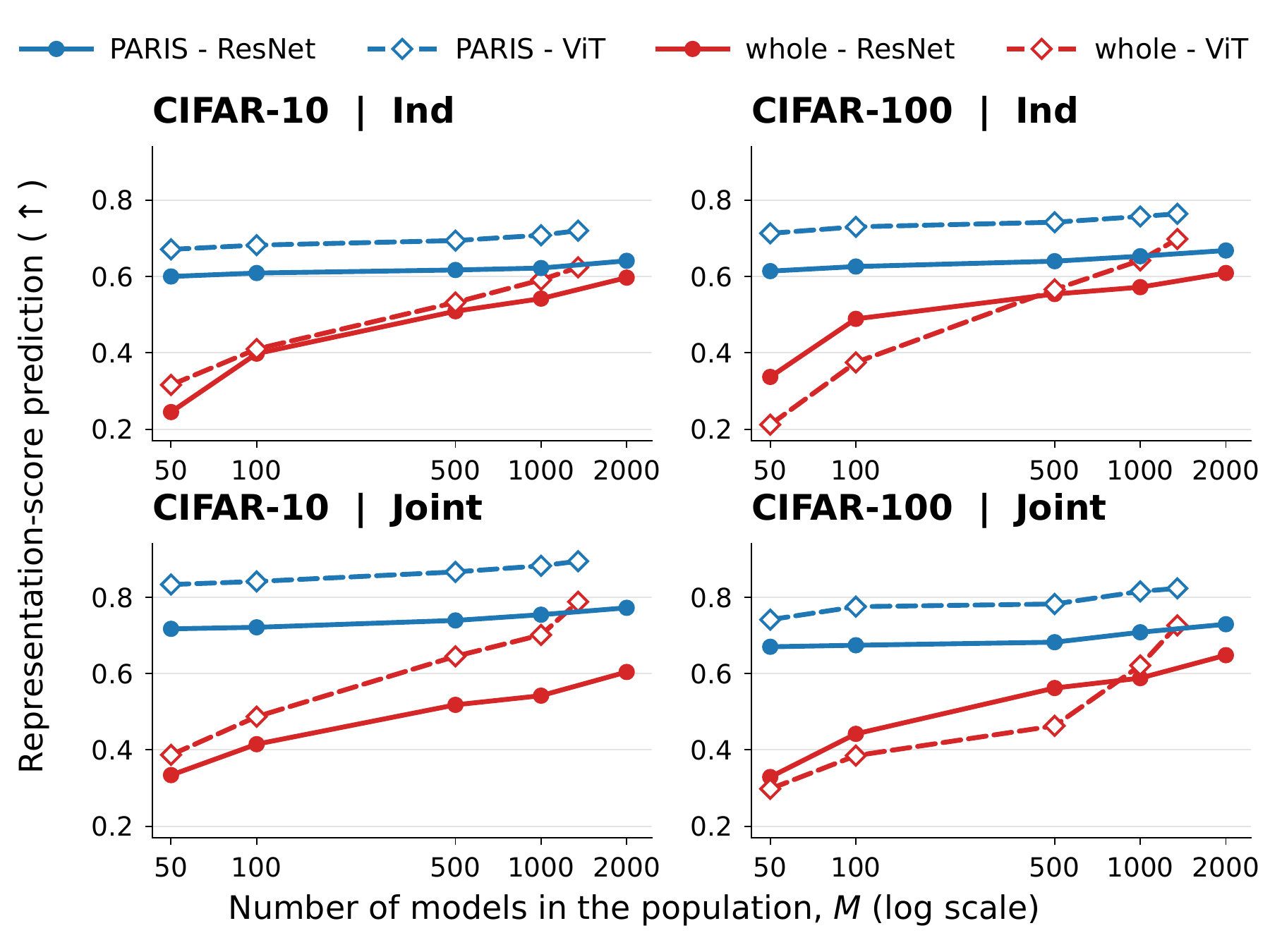}
\caption{ 
Representation-score prediction performance of various model population sizes.
}
\label{fig:paris_vs_whole}

\end{figure*}

To investigate the benefit of site-specific upstream computation graphs, we compare $\mathsf{PARIS}$ against an alternative that encodes the entire network instead of restricting the input to the computations leading to the queried site. We evaluate this alternative for both independent and joint comparison designs, varying the number of population.

Fig.~\ref{fig:paris_vs_whole} reports the representation-score prediction performance on ResNet and ViT populations trained on CIFAR-10 and CIFAR-100. Across these settings, upstream encoding outperforms whole-network encoding. The difference is pronounced for smaller populations, where the whole-network alternative exhibits substantially lower prediction correlations. As the number of training models increases, whole-network encoding improves and the gap narrows, but upstream encoding generally retains an advantage.

These results support the use of site-specific upstream graphs as an effective inductive bias for learning representation similarity, especially in limited-data settings. By excluding parameters that cannot influence the selected representation, our method also avoids requiring the metanetwork to learn to disregard them from training examples. We note that this benefit comes with a computational trade-off: separately encoding multiple upstream graphs may duplicate shared computations, whereas whole-network encoding can potentially amortize this cost across the internal sites.

\subsection{Baseline adaptations}
\label{app:baseline adaptations}

\noindent\textbf{Common comparison protocol.}
All methods operate on the same representation sites, site pairs, and
candidate banks.
For a comparison rule $h\in\mathcal H$, the target score for sites
$u$ and $v$ is
\begin{equation}
    y_{uv}^{(h)}
    =
    \Delta_h(\mH_u,\mH_v),
\end{equation}
where smaller values indicate more similar representations.
For retrieval, a query site $u$ ranks candidate sites $v$ in ascending
order of either a predicted dissimilarity or a directly measured
comparison score.
Within each experiment, all methods use the same reference distribution
and the same valid site pairs.
CIFAR-10 and CIFAR-100 define separate evaluation settings.
Candidates from the query models are excluded.

\noindent\textbf{Information available at test time.}
Response-based methods evaluate the compared models on a shared set of
inputs and use the resulting representation matrices.
weightspace methods instead receive only the parameterized upstream
computations $\mathcal C_u$ and $\mathcal C_v$.
These computations include the architecture, trained parameters, and
evaluation-time state needed to produce the two representations.
They exclude downstream weights, model identifiers, source-dataset
labels, and any statistics computed from reference responses.
During meta-training, learned weightspace methods use target scores
computed from reference responses.
At test time, they do not need reference inputs or base-model forward
passes.

\noindent\textbf{Shared training protocol.}
All learned comparators use the same model-disjoint splits, sampled
site pairs, and target scores.
We train them with the regression and ranking losses from the main
text.
Independent encoders also use the same embedding-variance regularizer.
We select models by the same mean-correlation criterion on
meta-validation models.
We compute normalization statistics and any self-supervised pretraining
only from meta-training models.
We tune baseline-specific hyperparameters on meta-validation models
without using test targets.

\noindent\textbf{Common head for independent encoders.}
For each independent baseline $b$, we map its observation of
$\mathcal C_u$ to a site embedding
\begin{equation}
    \vz_u^{(b)}
    =
    E_b\left(\operatorname{Obs}_b(\mathcal C_u)\right)
    \in\R^k.
\end{equation}
For comparison rule $h$, we then compute
\begin{equation}
    d_{b,h}(u,v)
    =
    \left\|
    \mW_{b,h}
    \left(
    \vz_u^{(b)}-\vz_v^{(b)}
    \right)
    \right\|_2
\end{equation}
and calibrate this distance to the target scale, $
    \widehat y_{uv,b}^{(h)}
    =
    q_{b,h}\left(d_{b,h}(u,v)\right)$.
We use the same embedding dimension and metric-head family for all
independent learned methods, but each method has its own parameters.
The calibration is monotone, so it does not change candidate rankings.
For retrieval, we use the uncalibrated distance
$d_{b,h}$, which also avoids numerical ties when the calibration
saturates.

These common heads make the comparison focus on the encoder:
each method receives the same supervision and predicts the same target
scores, but it represents the upstream computation differently.
The heads and supervised objectives are benchmark adaptations and are
not part of the original applications of every cited method.

\noindent\textbf{Response-based comparison methods.}
Response-based baselines evaluate the compared sites on a shared
comparison set $\mX_{\mathrm{cmp}}$.
We keep $\mX_{\mathrm{cmp}}$ disjoint from the reference examples used
to define the final evaluation targets.
This avoids giving these baselines the responses from which the
target score is computed.

\begin{enumerate}[leftmargin=1.75em,itemsep=2pt,topsep=0pt, itemsep=0pt, parsep=0pt, partopsep=0pt]
\item \noindent\textbf{SVCCA.}
Singular Vector Canonical Correlation Analysis (SVCCA)
\citep{raghu2017svcca} first reduces each representation by singular
value decomposition and then applies canonical correlation analysis
(CCA).
We evaluate $\mH_u(\mX_{\mathrm{cmp}})$ and
$\mH_v(\mX_{\mathrm{cmp}})$ using the same extraction rules as in the
main benchmark.
After centering each feature over examples, we retain principal
directions that explain $99\%$ of the variance and compute their
canonical correlations $\rho_1,\ldots,\rho_r$.
We rank candidates by
\begin{equation}
    d_{\mathrm{SVCCA}}(u,v)
    =
    1-\frac{1}{r}\sum_{i=1}^{r}\rho_i.
\end{equation}
We fix the numerical rank tolerance and CCA regularization on
meta-validation data.
SVCCA does not use a learned weightspace encoder or a
comparison-rule-specific head.
\item \noindent\textbf{PWCCA.}
Projection-Weighted CCA (PWCCA) \citep{morcos2018insights} weights the
canonical correlations by how strongly their canonical directions
contribute to the original representation.
We compute CCA on centered responses and obtain normalized projection
weights from the orthonormalized canonical variates.
Because the original weighting is directional, we compute the weights
in both directions.
Let $\alpha_i^{u\mid v}$ and $\alpha_i^{v\mid u}$ denote the two sets
of weights, each normalized to sum to one.
We use
\begin{equation}
    d_{\mathrm{PWCCA}}(u,v)
    =
    1-\frac{1}{2}\left(
      \sum_i\alpha_i^{u\mid v}\rho_i
      +
      \sum_i\alpha_i^{v\mid u}\rho_i
    \right).
\end{equation}

\end{enumerate}

Our benchmark compares selected internal sites.
A whole-network embedding reused at every site cannot distinguish
different locations in the same model, while an isolated layer tensor
can omit upstream information that changes the representation.
For example,
$\vr_u(\vx)=\mW_2\sigma(\mW_1\vx)$ depends on $\mW_1$ even when $\mW_2$ is fixed.
We then give learned weightspace baselines the upstream
computation $\mathcal C_u$ and mark the selected site when the method
requires it.

\noindent\textbf{Independent weightspace encoders.}

\begin{enumerate}[leftmargin=1.75em,itemsep=2pt,topsep=0pt, itemsep=0pt, parsep=0pt, partopsep=0pt]
\item \noindent\textbf{Metadata.}
This architecture-only control maps a fixed-length descriptor of
$\mathcal C_u$ to a site embedding with an MLP.
The descriptor contains upstream computational depth, counts of
operator types, the target tensor shape, and the number of upstream
parameters.
We standardize numerical features using meta-training statistics.
We do not normalize site depth by the full model depth because this
would reveal information downstream of the site.
This baseline uses no trained weight values and tests how much of the
target score can be predicted from architecture and site location
alone.
\item \noindent\textbf{Tensor marginal.}
This control summarizes the marginal distribution of values in each
upstream parameter tensor, following the general idea of
weightstatistics predictors \citep{unterthiner2020predicting}.
For each tensor, we compute its mean, standard deviation,
root-mean-square magnitude, and fixed quantiles, and append descriptors
for tensor role and shape.
We pool these summaries within tensor-role groups using fixed sums,
means, and maxima.
An MLP maps the pooled summary, together with the Metadata descriptor,
to the common embedding space.
This baseline keeps coarse information about parameter distributions
but discards coordinate assignments within tensors and correspondences
across tensors.

\item \noindent\textbf{DeepSets.}
We adapt the permutation-invariant architecture of
\citet{zaheer2017deep} to a set of upstream weight tokens.
We split tensors into fixed-width blocks, mask padding, and append the
tensor type, shape, and position in the upstream computation to each
token.
If $\xi_{u,j}$ denotes an attributed token, we compute
\begin{equation}
    \vz_u^{\mathrm{DS}}
    =
    \rho_{\mathrm{DS}}\left(
      \left[
      \sum_j\varphi_{\mathrm{DS}}(\xi_{u,j}),
      \;
      \frac{1}{N_u}\sum_j\varphi_{\mathrm{DS}}(\xi_{u,j}),
      \;
      \bm m_u
      \right]
    \right),
\end{equation}
where $\bm m_u$ is the Metadata descriptor.
Unlike Tensor marginal, DeepSets learns features directly from raw
weight tokens.
It does not use graph message passing or interactions between tokens.
Its permutation invariance concerns the order in which attributed
tokens are listed. Note that it does not make the encoder invariant to arbitrary
reparameterizations of the network.
\item \noindent\textbf{SANE-FT.}
SANE \citep{schurholt2024towards} represents model weights as
fixed-width tokens processed by a Transformer-based sequential
autoencoder.
For our site-level task, we tokenize only $\mathcal C_u$, retain layer
and within-layer position information, and restrict all context windows
to this upstream computation.
We pool the token embeddings with masks and project the result to the
common site-embedding space.
We pretrain on meta-training sites using SANE's reconstruction and
contrastive objectives, discard the decoder, and fine-tune the encoder
with the shared comparison-score supervision.
We call this adaptation \emph{SANE-FT}.
Our heterogeneous-prefix setting does not use whole-network canonical
alignment, so it should not be read as a reproduction of the aligned
SANE configuration.
When we report an aligned variant separately, its reference and all
alignment statistics come only from meta-training lineages.
\item \noindent\textbf{ProbeX-Prefix.}
ProbeX \citep{horwitz2025learning} applies learned probes to weight
matrices and combines their responses with factorized encoders.
Its original formulation can combine information from multiple layers.
For our task, we reshape compatible upstream tensors into matrices,
apply ProbeX modules, and pool their fixed-dimensional outputs together
with tensor-role and prefix-position descriptors.
This produces one embedding for the selected site, which is then
compared with the common metric heads.
We call this adaptation \emph{ProbeX-Prefix}.
ProbeX modules are shared across tensors with compatible shapes.
When a matrix shape is unsupported, we report the missing coverage and
use a labeled matched-subset comparison. 
The probes act on weights, not on CIFAR inputs, and the method receives
no lineage or model-identity labels.
\item \noindent\textbf{Graph Metanetwork.}
We adapt the parameter-graph encoder of \citet{lim2024graph} to
$\mathcal C_u$.
We retain its parameter-attributed graph construction and
message-passing backbone, including learned edge features.
A target marker identifies the selected site.
We replace the original task-specific output with a readout that pools
features from the upstream graph and the target nodes and maps them to
the common embedding space.
We retain the published parameter-graph representation. 
This baseline isolates the effect of our graph construction
under the same site-level supervision and metric-head family.
\item \noindent\textbf{Neural Graph}.
We adapt neural graph (NG) \citet{kofinas2024graph} to each selected site’s upstream computation $\mathcal C_u$. We retain its neural-graph parameter encoding: neurons or channels form nodes, biases become node features, and weights become edge features. Convolution kernels are padded and flattened into vector-valued edge features. We process this graph with the PNA-based NG-GNN encoder, including its learned edge updates and edge-conditioned messages. For ViT, we follow \cite{kofinas2024graph}, which retains the query, key, and value projections as separate components of the edge features. A target marker identifies the selected site. We replace the original task-specific output with a readout that pools features from the upstream graph and the target nodes and projects them to the common site-embedding space. At an internal target site, coordinates share the same layer embedding and target marker, and the readout pools them without assigning distinct output-coordinate identifiers. Each site is encoded independently and compared using the common relation-specific metric heads. We use the shared training protocol described above. We omit the optional activation-probe features because they require evaluating the compared networks on learned inputs. The resulting baseline predicts the same representation relations from the upstream neural artifacts without collecting new base-model activations at inference.

\end{enumerate}

\noindent\textbf{Graph Matching Network.}
We adapt the Graph Matching Network (GMN) of \citet{li2019graph}, which
combines within-graph message passing with cross-graph matching.
GMN receives the same parameter-attributed
site graphs and target markers as the joint $\mathsf{PARIS}$ model.
Cross-graph matching compares each node state with an
attention-weighted summary of the other graph, and gated pooling
produces pair-dependent graph summaries.
We replace the original task output and loss with $H$ bounded
comparison-score predictions trained with the shared regression and
ranking losses.
When needed, we symmetrize the GMN score by averaging the two input
orders.
This adaptation evaluates its joint matching architecture under our
site-level supervision.

\noindent\textbf{Stitching-selection baselines.}
The stitching experiment asks whether a comparison score can screen
useful connections before adapter fitting.
We use frozen-model stitching \citep{bansal2021revisiting}.
We do not compare complete model-assembly systems.
Instead, every selector receives the same candidate bank, chooses a
fixed-size shortlist, and then uses the same adapter-fitting and
validation procedure.
The experiment isolates candidate-selection quality.
This experiment tests whether these scores provide a useful
screening signal under a limited adapter-fitting budget.

A stitching candidate
$c=(A,s,B,t)$
connects $(A,s)$ to  $(B,t)$: $
    f_c
    =
    B_{>t}\circ S_c\circ A_{\leq s}$,
where $S_c$ is the trainable adapter and $B_{>t}$ starts after receiver
site $t$.
For each selection problem, all methods receive the same eligible
candidate set $\mathcal K$ and choose $k$ candidates, with
$k\leq|\mathcal K|$.
Eligibility fixes the architecture family, stitching direction,
adapter compatibility, and any compute constraint before a selector
scores the candidates.
Both constituent networks remain frozen, including their
evaluation-time normalization state.
We fit a separate adapter for every selected candidate using the same
initialization and optimization protocol.

\noindent\textbf{Using learned comparison scores for stitching.}
For candidate $c=(A,s,B,t)$, let
$u=(A,s)$ and $v=(B,t)$.
We choose the comparison rule $h^\star$ on meta-validation and
keep it fixed for test evaluation.
An independent learned method ranks candidates by $
    s_b(c)
    =
    d_{b,h^\star}(u,v)$,
while a joint method uses $
    s_b(c)
    =
    \widehat y_{uv,b}^{(h^\star)}$.

\noindent\textbf{SN-Net-H.}
SN-Net uses nearest-anchor selection and paired/unpaired block matching
to define a stitching space \citep{pan2023stitchable}.
Its full method also trains an elastic collection of stitched models,
which differs from our frozen-model protocol.
We isolate its architectural selection heuristic and call the
resulting baseline \emph{SN-Net-H}.

We order anchor configurations by standalone FLOPs at the benchmark
input resolution.
Independent runs with the same configuration share the same position.
Within compatible stages, we use the published paired/unpaired
construction, including nearest-neighbor interpolation for unequal
depths, with window size two and stride one.
A one-block stage uses a singleton window.
After intersecting these configurations with $\mathcal K$, we obtain a
preferred set $\mathcal K_H$.

The SN-Net heuristic does not define a unique top-$k$ ranking.
We shuffle $\mathcal K_H$ with a recorded random seed and
take its first $k$ candidates.
If $|\mathcal K_H|<k$, we keep all preferred candidates and sample the
remaining candidates uniformly without replacement from
$\mathcal K\setminus\mathcal K_H$.
We report the coverage of $\mathcal K_H$ and how often this fallback is
used.
We reuse the same shuffled order across shortlist sizes to obtain
nested lists.
This sampling rule and the frozen-model fitting protocol are our
benchmark adaptations. They are not part of the original SN-Net
training procedure.

\noindent\textbf{KLAS-score.}
KLAS uses task probes to construct a normalized directional KL score
for stitching selection \citep{sanyal2026klas}.
We evaluate this score within our common candidate bank and call the
resulting baseline \emph{KLAS-score}.
For every required site, we fit an affine classifier on a fitting set
$\mathcal B_{\mathrm{fit}}$ while keeping the base model frozen.
For site $(A,s)$,
\begin{equation}
    \vp_{A,s}(x)
    =
    \operatorname{softmax}\left(
        \mU_{A,s}\vr_{(A,s)}(\vx)+\vb_{A,s}
    \right)
    \in\R^C,
\end{equation}
where $C$ is the number of downstream classes.
All probes use the same downstream label space.
We choose probe hyperparameters on the development model and compute
scores on a disjoint ranking set $\mathcal B_{\mathrm{rank}}$.
Write
\begin{equation}
    \Theta\left((A,s),(B,t)\right)
    =
    \frac{1}{|\mathcal B_{\mathrm{rank}}|}
    \sum_{x\in\mathcal B_{\mathrm{rank}}}
    D_{\mathrm{KL}}\left(
        p_{A,s}(x)\,\|\,p_{B,t}(x)
    \right).
\end{equation}
For candidate $c=(A,s,B,t)$, KLAS-score ranks by
\begin{equation}
    \Gamma(c)
    =
    \frac{
        \Theta\left((A,s),(B,t)\right)
    }{
        \max\left\{
        \Theta\left((B,t),(B,t+1)\right),
        \epsilon
        \right\}
    },
    \qquad
    \epsilon=10^{-8}.
    \label{eq:klas_score_adaptation}
\end{equation}
The denominator compares the receiver site with the output after its
next block.
We keep the directed KL divergence and take a ratio of the
sample-averaged divergences.

The full KLAS procedure also performs anchor filtering and
threshold-based selection across FLOPs buckets.
To compare its score on the same candidate bank as the other methods,
we omit the preliminary anchor filter and select the $k$ candidates
with the smallest $\Gamma(c)$ inside the prescribed compute interval.
Thus, this baseline evaluates the KLAS selection score. 
KLAS-score retains its response and task-label access, and we do not remove this information to imitate our weights-only setting.

\section{Additional experiments}
\label{app:more_exp}

\subsection{Assessing cross-graph interaction}

\begin{figure*}[h]
    \centering
     \includegraphics[width=0.65\textwidth]{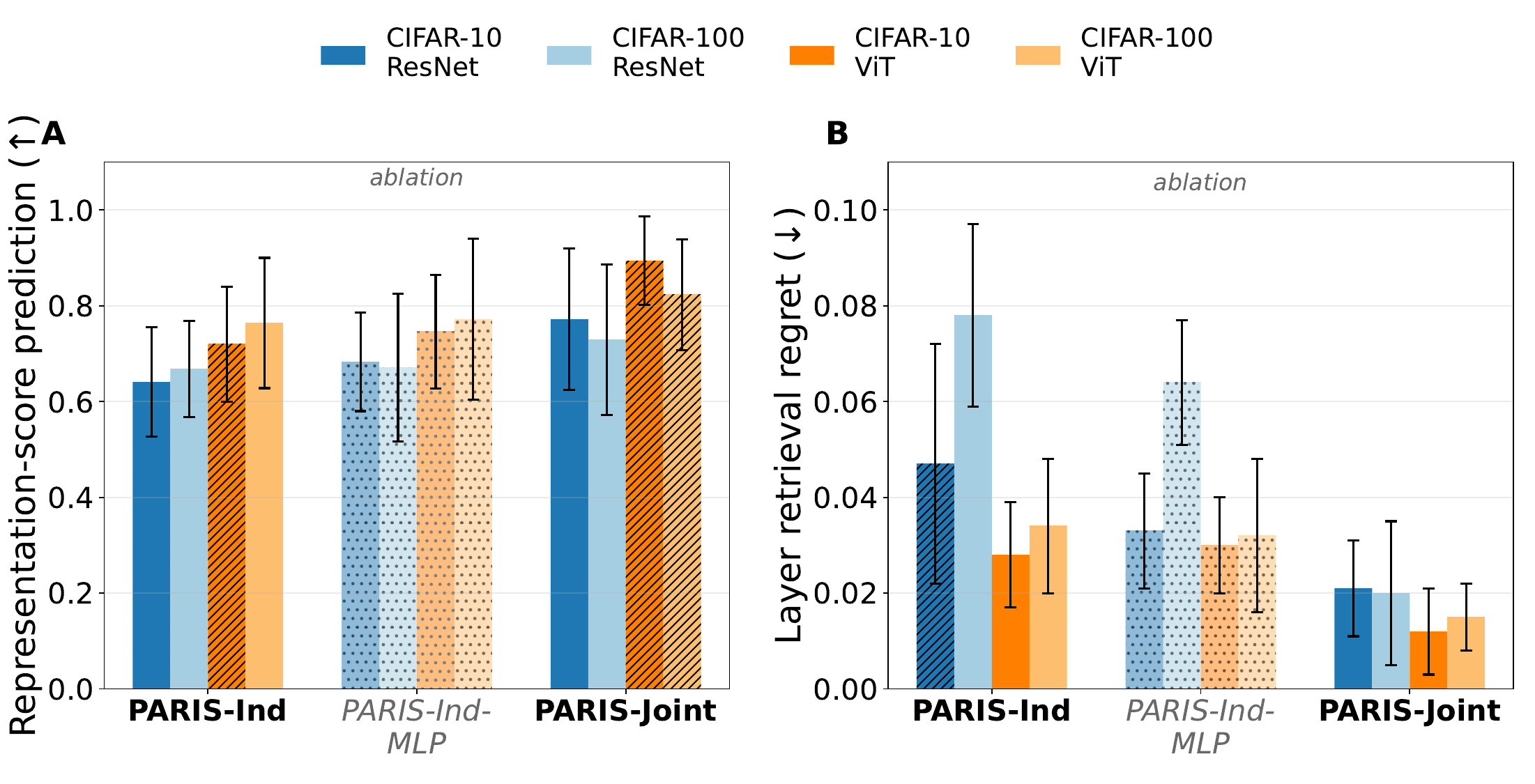}
\caption{ 
Cross graph interaction analysis. 
$\mathsf{PARIS}$-Ind-MLP replaces the independent estimator's
calibrated metric head with a symmetric MLP while retaining partner-independent site embeddings.
The MLP improves prediction scores but does not close the gap to $\mathsf{PARIS}$-Joint.
}
\label{fig:cross_graph_ablation}
\end{figure*}

The comparison between $\mathsf{PARIS}$-Ind and $\mathsf{PARIS}$-Joint changes both how the site graphs are processed and how their relation is predicted. To examine whether the joint estimator's advantage can
be explained by its more flexible comparison head, we evaluate $\mathsf{PARIS}$-Ind-MLP. This variant retains partner-independent site embeddings but replaces the projected Euclidean distance and exponential calibration with a symmetric nonlinear one:
\begin{equation}
    \widehat{\boldsymbol y}_{uv}
    =
    \sigma\left(
        \operatorname{MLP}\left(
            [\vz_u+\vz_v
            \;\Vert\;
            |\vz_u-\vz_v|
            \;\Vert\;
            \vz_u\odot \vz_v]
        \right)
    \right),
    \label{eq:ind_mlp_ablation}
\end{equation}
where $\vz_u$ and $\vz_v$ are computed independently and $\sigma$ is an elementwise sigmoid. Thus, the final comparison is no longer restricted to a calibrated projected distance, but there is no cross-graph interaction before site-level pooling.
Fig.~\ref{fig:cross_graph_ablation} shows that replacing the metric head improves the reported prediction score. 
Yet, $\mathsf{PARIS}$-Joint remains ahead of the MLP
control. Thus, decoder flexibility contributes to
the observed improvement, but simply replacing it can not compete with the joint estimator. 
The effect of the MLP control is less uniform for retrieval. It improves the reported retrieval criterion in three of the four settings, but the ViT/CIFAR-10 value increases.  The advantage of the joint design  extends to the downstream correspondence-selection task and not appearing only in aggregate relation-prediction scores.

\subsection{Ablating regression and ranking supervision}

\begin{figure*}[h]
    \centering
     \includegraphics[width=0.65\textwidth]{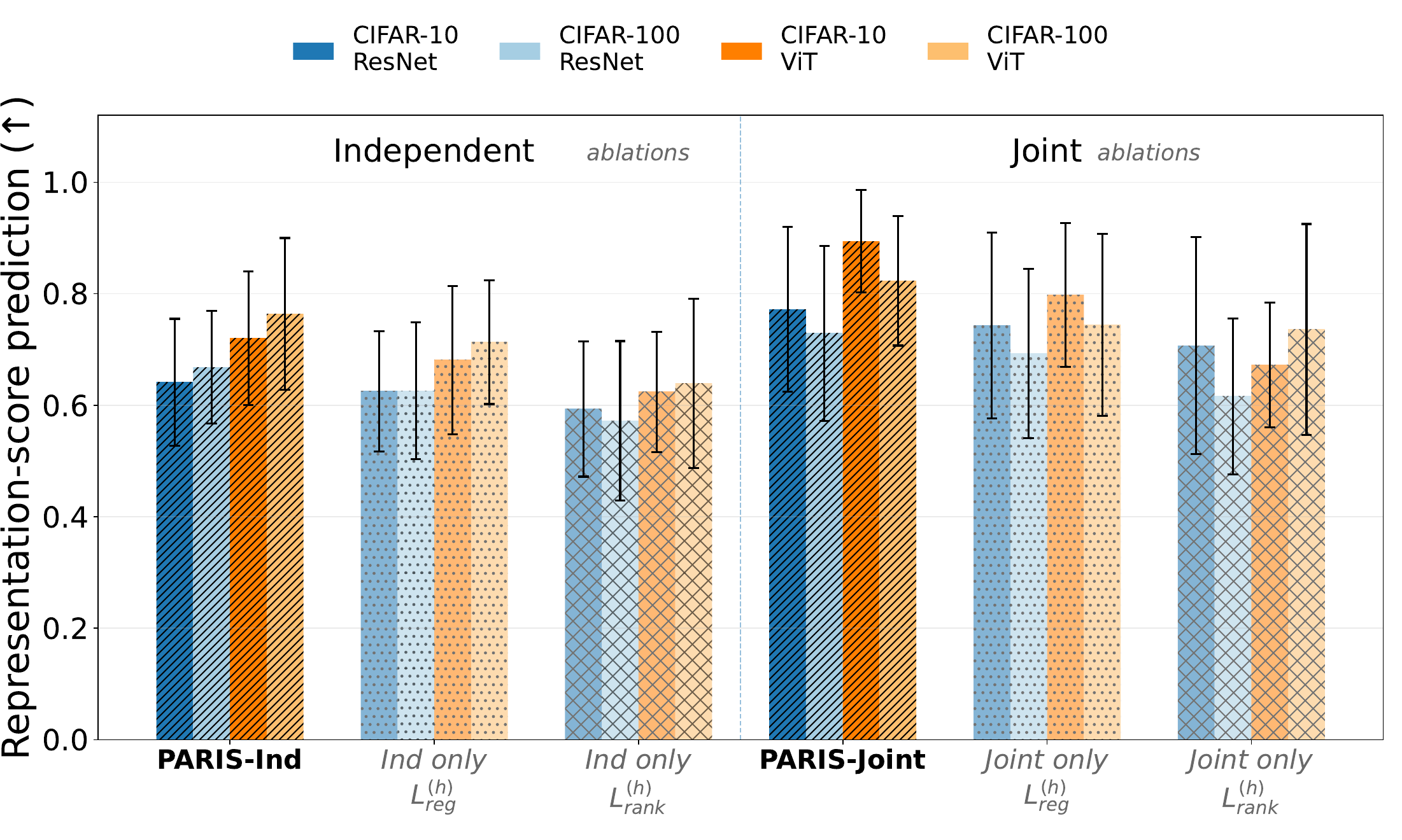}
\caption{ 
Complementary regression and ranking supervision.
Representation-score prediction ($\uparrow$) for
$\mathsf{PARIS}$-Ind and $\mathsf{PARIS}$-Joint using combined regression and ranking pair supervision, regression alone, or ranking alone. 
}
\label{fig:loss_ablation}
\end{figure*}

The supervised objective combines a smooth regression loss $\mathcal L_{\mathrm{reg}}^{(h)}$, which fits the numerical target dissimilarities, with an ordinal loss $\mathcal L_{\mathrm{rank}}^{(h)}$, which encourages their relative ordering. To assess the contribution of each component, we compare the combined objective with regression only and ranking-only pair supervision for both $\mathsf{PARIS}$-Ind and
$\mathsf{PARIS}$-Joint. Fig.~\ref{fig:loss_ablation} reports representation-score prediction across the four dataset-architecture settings.
The combined objective achieves the highest reported mean for both estimators in all settings.  Regression-only supervision outperforms ranking-only supervision. However, both losses combined still perform the best. Regression directly constrains the numerical dissimilarities, while ranking explicitly penalizes incorrect relative ordering across supervised site pairs.
Although the ranking loss also encourages separation between predictions, it does not directly fit the magnitudes of the target dissimilarities. Conversely, regression alone does not include an explicit pair-ordering objective.

\subsection{Sensitivity to the sample size}

\begin{figure*}[h]
    \centering
     \includegraphics[width=0.65\textwidth]{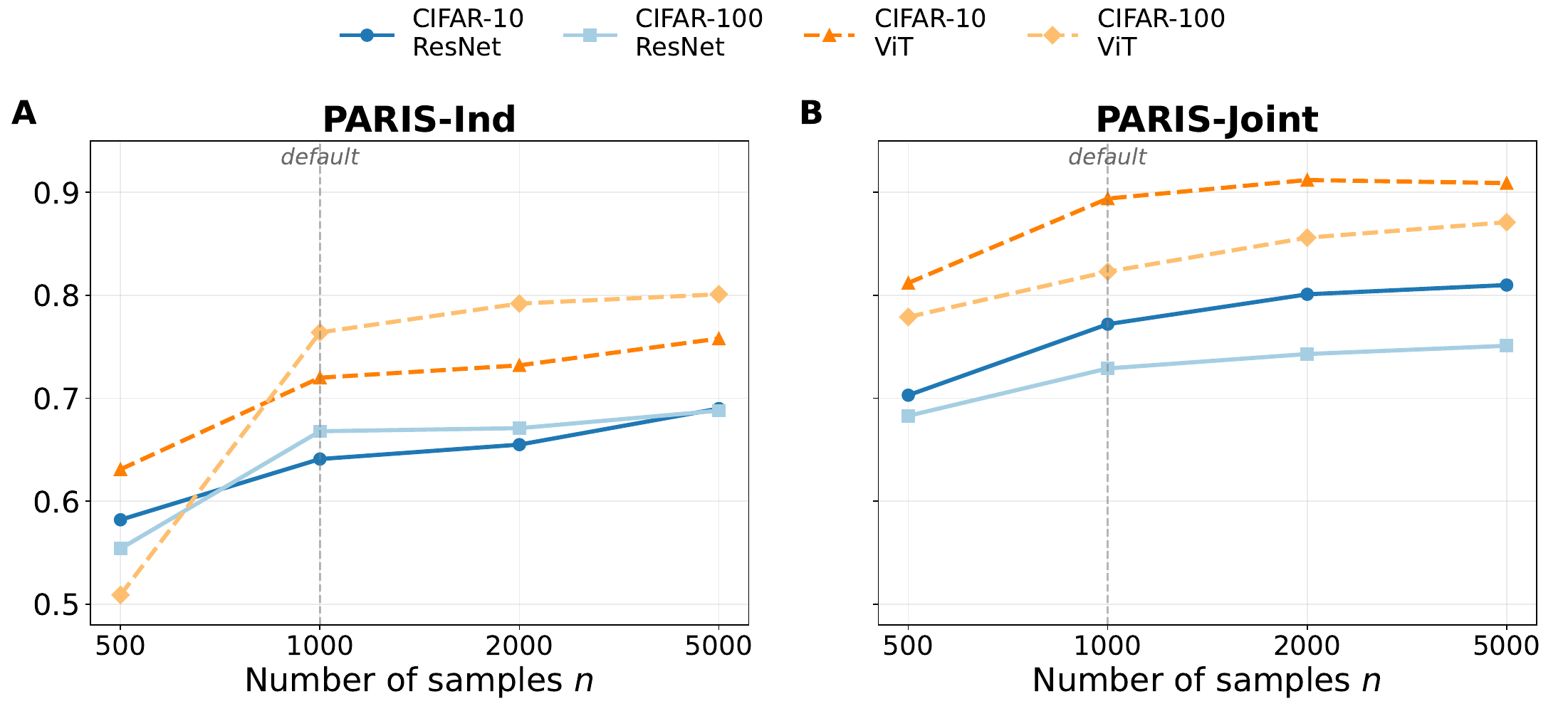}
\caption{ 
Effect of reference-sample size on predicted-reference correlation. Correlation between predicted and activation-derived reference dissimilarities for $\mathsf{PARIS}$-Ind and $\mathsf{PARIS}$-Joint. 
}
\label{fig:sample_size}

\end{figure*}

We examine the effect of the number of inputs in the ordered reference sample used to compute activation-derived
dissimilarities.
Fig.~\ref{fig:sample_size} reports predicted-reference dissimilarity correlations for $n\in\{500,1000,2000,5000\}$ across the four dataset-architecture settings. 
Increasing the reference size from 500 to 1000 improves the correlation for both estimators. 
The larger-reference regime has smaller gains than the initial increase to inputs.

\subsection{Per-representation score correlation}

\begin{figure*}[h!]
    \centering
     \includegraphics[width=0.65\textwidth]{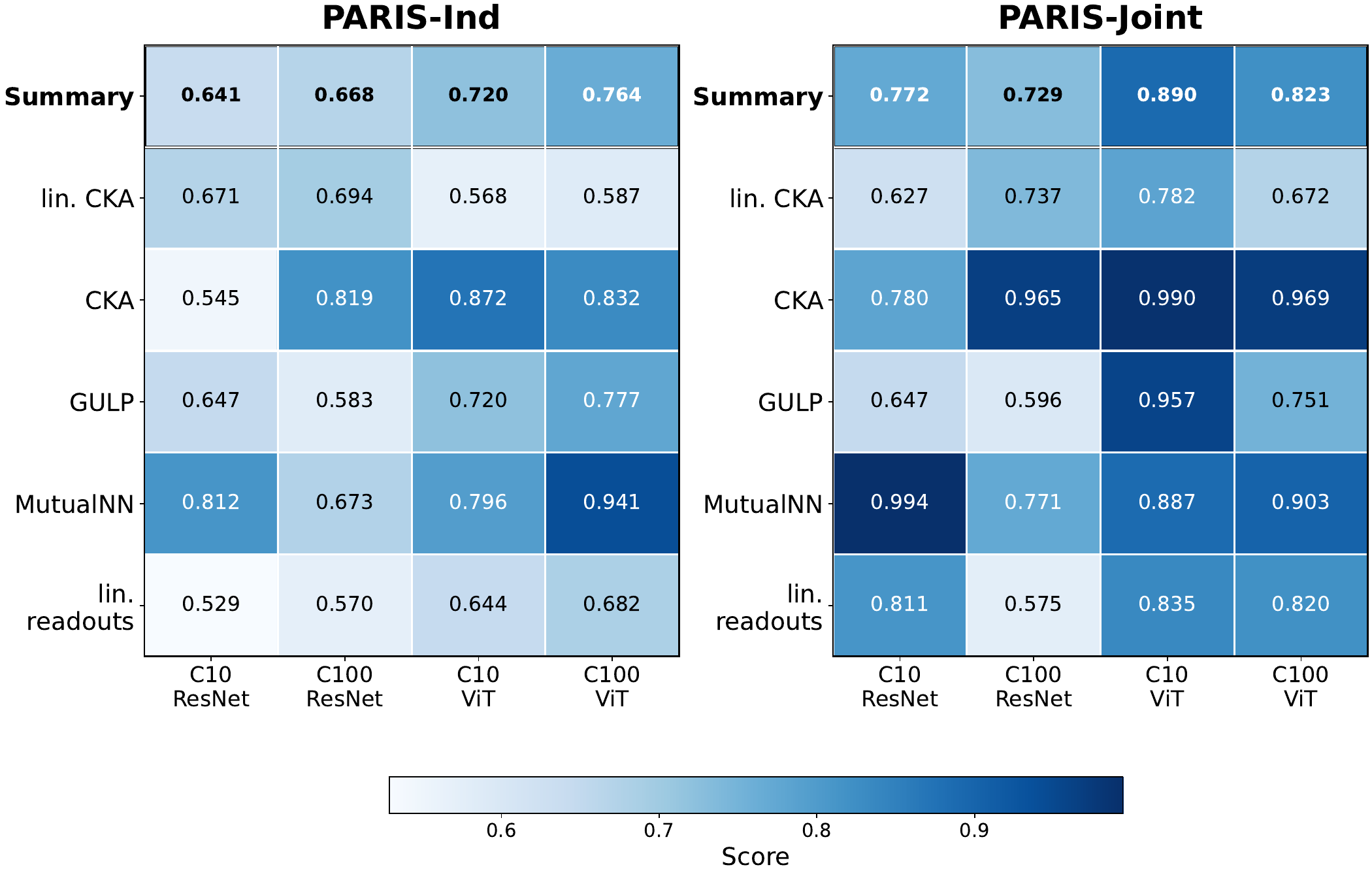}
\caption{ 
Per-representation score correlation between predicted and reference dissimilarities.
}
\label{fig:per_relation_results}
\end{figure*}

Fig.~\ref{fig:per_relation_results} reports the correlations between predicted and corresponding reference dissimilarities.
We see that  $\mathsf{PARIS}$-Joint yields higher predicted benchmark comparisons. These results indicate that the joint
estimator's advantage in correlation extends across several
reference internal neural representations.
The independent estimator is also competitive and has higher correlation for linear CKA on ResNet/CIFAR-10, and for GULP and MutualNN on ViT/CIFAR-100. The benefit varies with both the representation similarity and model population.

\begin{figure*}[h!]
    \centering
     \includegraphics[width=0.65\textwidth]{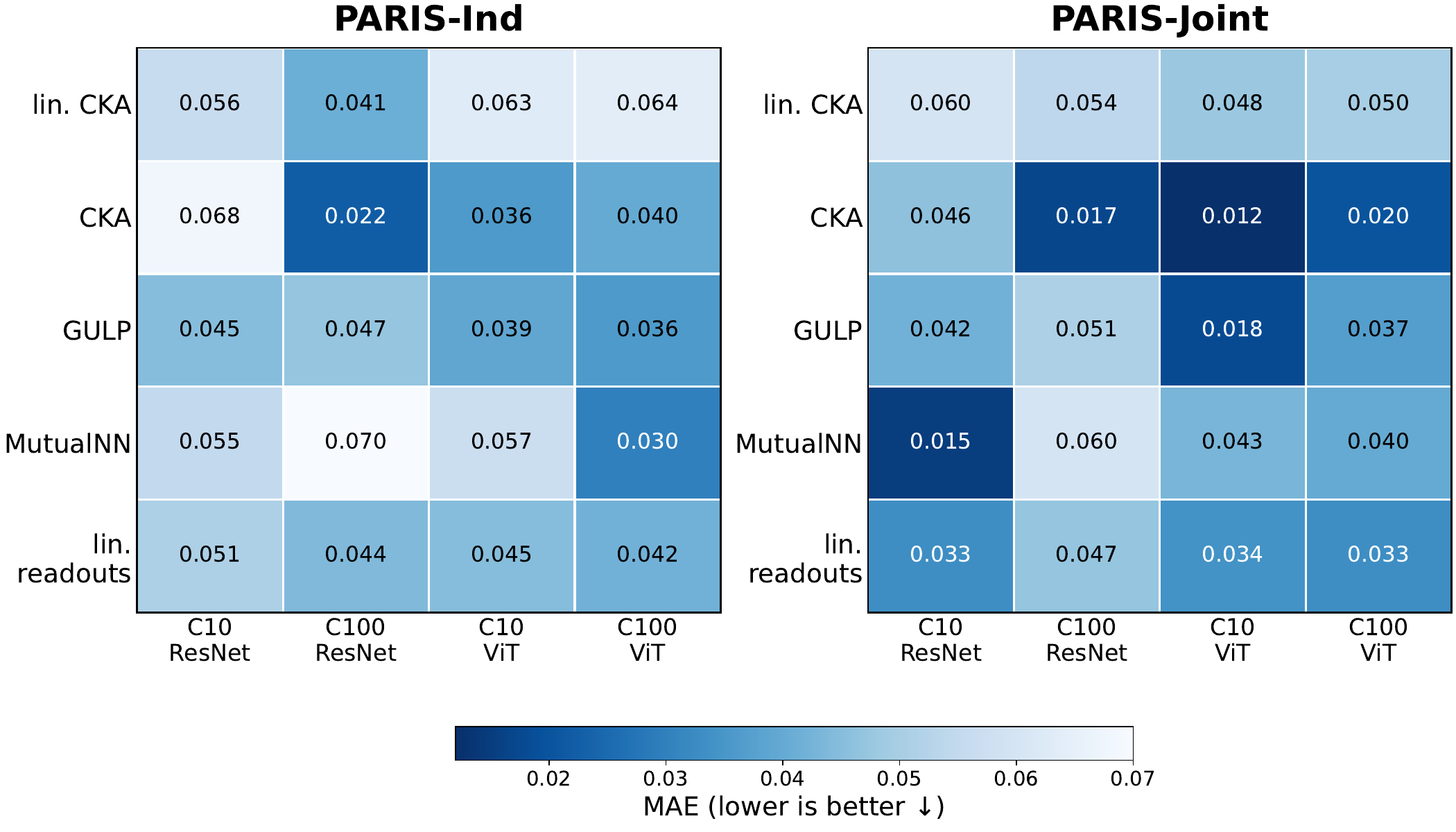}
\caption{ 
Per-representation score MAE between predicted and reference dissimilarities.
}
\label{fig:mae}
\end{figure*}

\begin{figure*}[h!]
    \centering
     \includegraphics[width=0.65\textwidth]{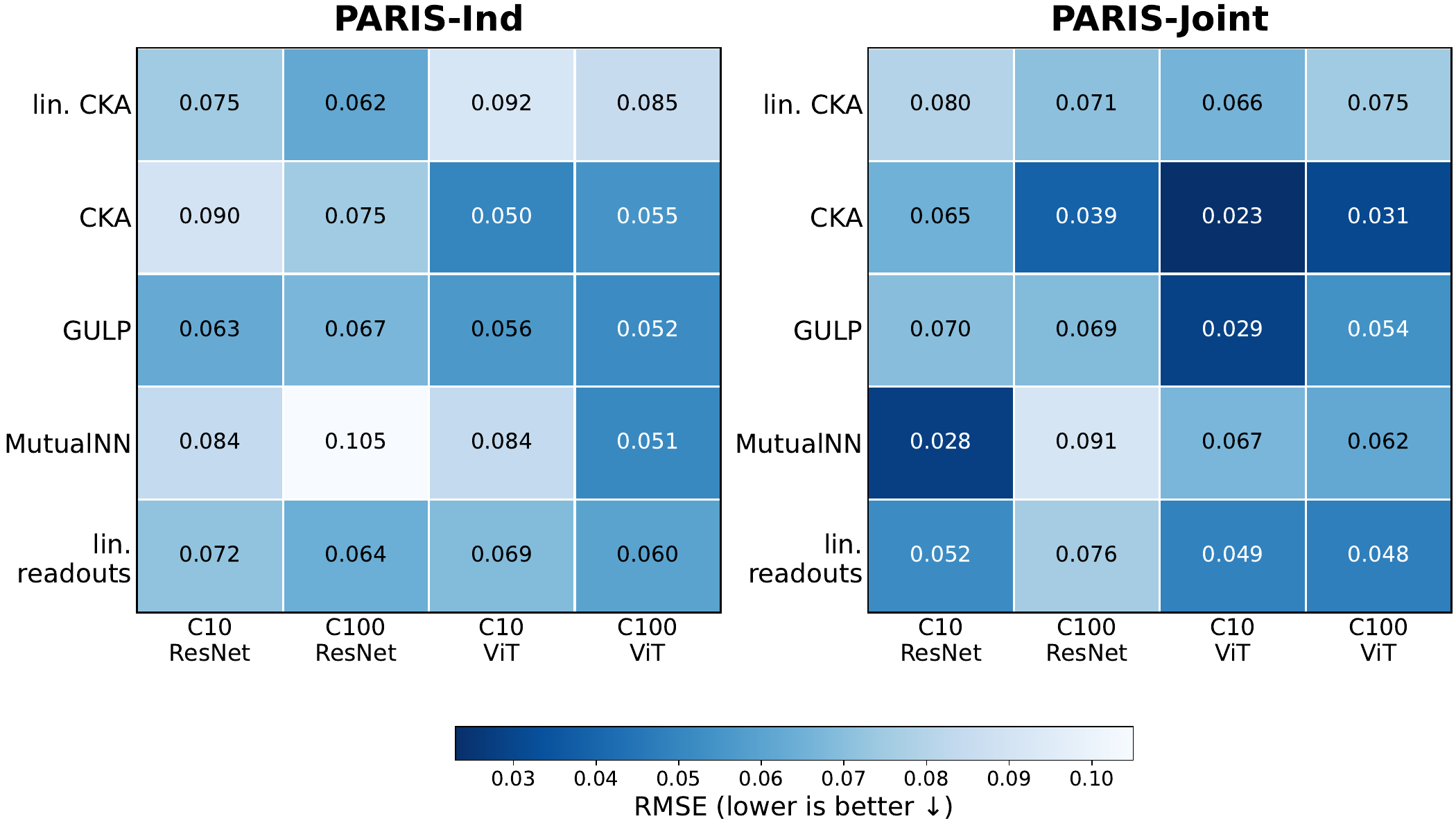}
\caption{ 
Per-representation score RMSE between predicted and reference dissimilarities.
}
\label{fig:rmse}
\end{figure*}

To complement the rank-based evaluation, we additionally report mean absolute error (MAE) and root mean squared error (RMSE) between predicted and reference dissimilarities. 
Fig.~\ref{fig:mae} and Fig.~\ref{fig:rmse} respectively show the MAE and RMSE between predicted and corresponding reference dissimilarities.
We observe that $\mathsf{PARIS}$-Joint generally achieves lower numerical errors and $\mathsf{PARIS}$-Ind performs better for certain measures and model populations. These results complement the correlations by quantifying prediction errors on the reference-score scale, providing a more complete evaluation of $\mathsf{PARIS}$'s ability to estimate representation dissimilarities beyond ranking agreement.

\subsection{Meta-training across diverse architectures}
\label{app:Meta-training across diverse architectures}
We study how the architectures used for meta-training affect the prediction of representation similarity, that is, we investigate how architecture diversity during meta-training affects representation similarity prediction and whether $\mathsf{PARIS}$ can generalize across different architectural configurations.

We compare two settings.
In \emph{Same family}, we train a separate metanetwork for each architecture family. In \emph{Diverse}, we train one metanetwork on a shared population that includes ResNets and ViTs. We then evaluate each predictor separately on ResNets and ViTs trained on CIFAR-10 and CIFAR-100. The experiment tests whether a predictor trained on different architecture families can still generalize well to each family.
We compare $\mathsf{PARIS}$ with  Deep Meta Classifier (DMC) \citep{eilertsen2020classifying} and Graph Metanetwork \citep{lim2024graph}.

\begin{figure*}[h]
    \centering
     \includegraphics[width=0.9\textwidth]{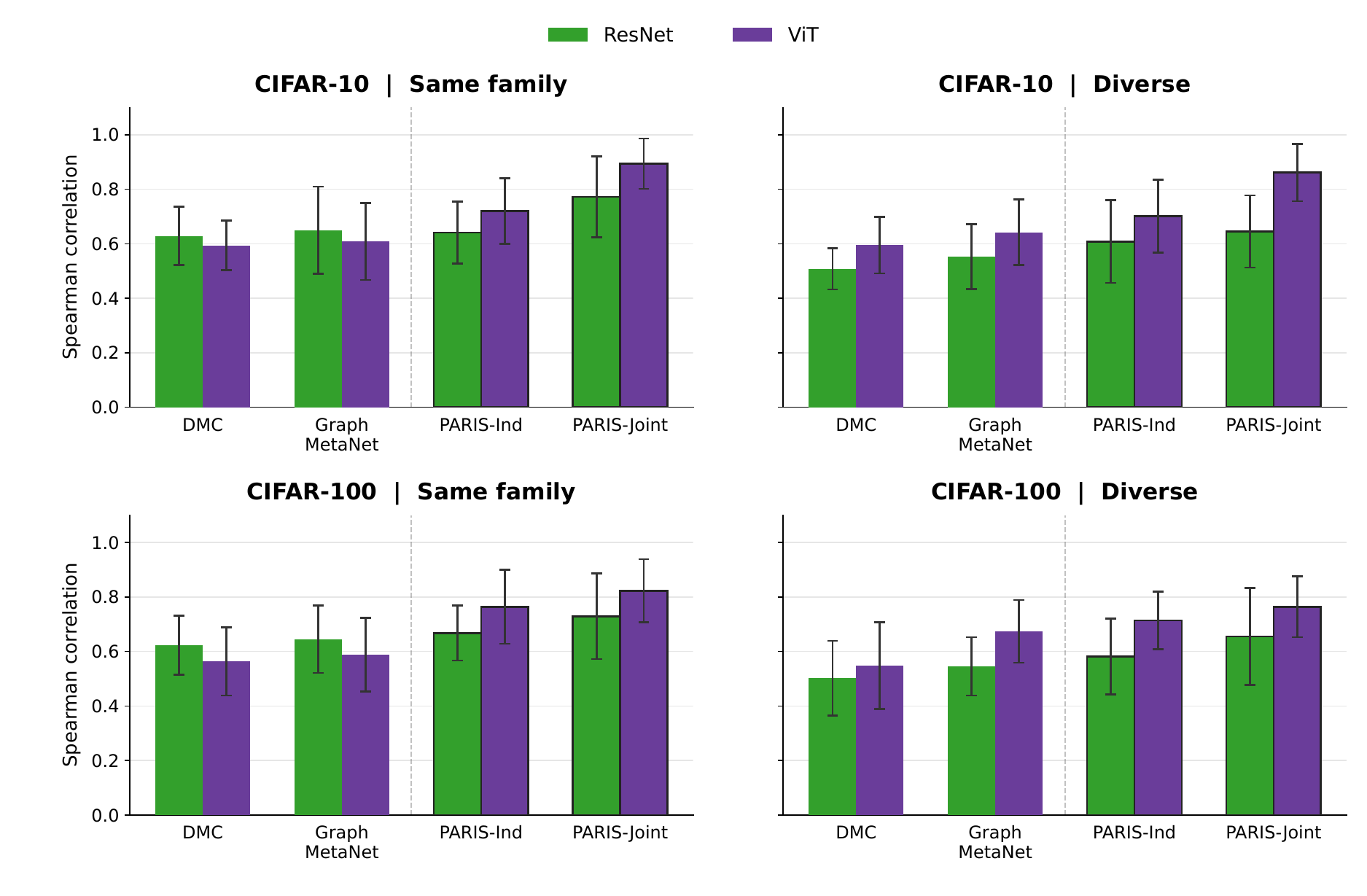}
\caption{ 
Representation-score prediction under same-family and diverse meta-training. Mean Spearman correlations for DMC, Graph Metanetwork, $\mathsf{PARIS}$-Ind, and $\mathsf{PARIS}$-Joint on held-out ResNet and ViT populations trained on CIFAR-10 and CIFAR-100.
}
\label{fig:diverse_architectures}
\end{figure*}

Fig.~\ref{fig:diverse_architectures}  shows that both $\mathsf{PARIS}$ variants outperform DMC and Graph Metanetwork across all four architecture/dataset settings.
These results demonstrate that a shared metanetwork can be trained on heterogeneous architecture populations and remain effective on held-out models from both families, while also showing the accuracy trade-off associated with diverse meta-training.

\begin{figure*}[h]
    \centering
     \includegraphics[width=0.45\textwidth]{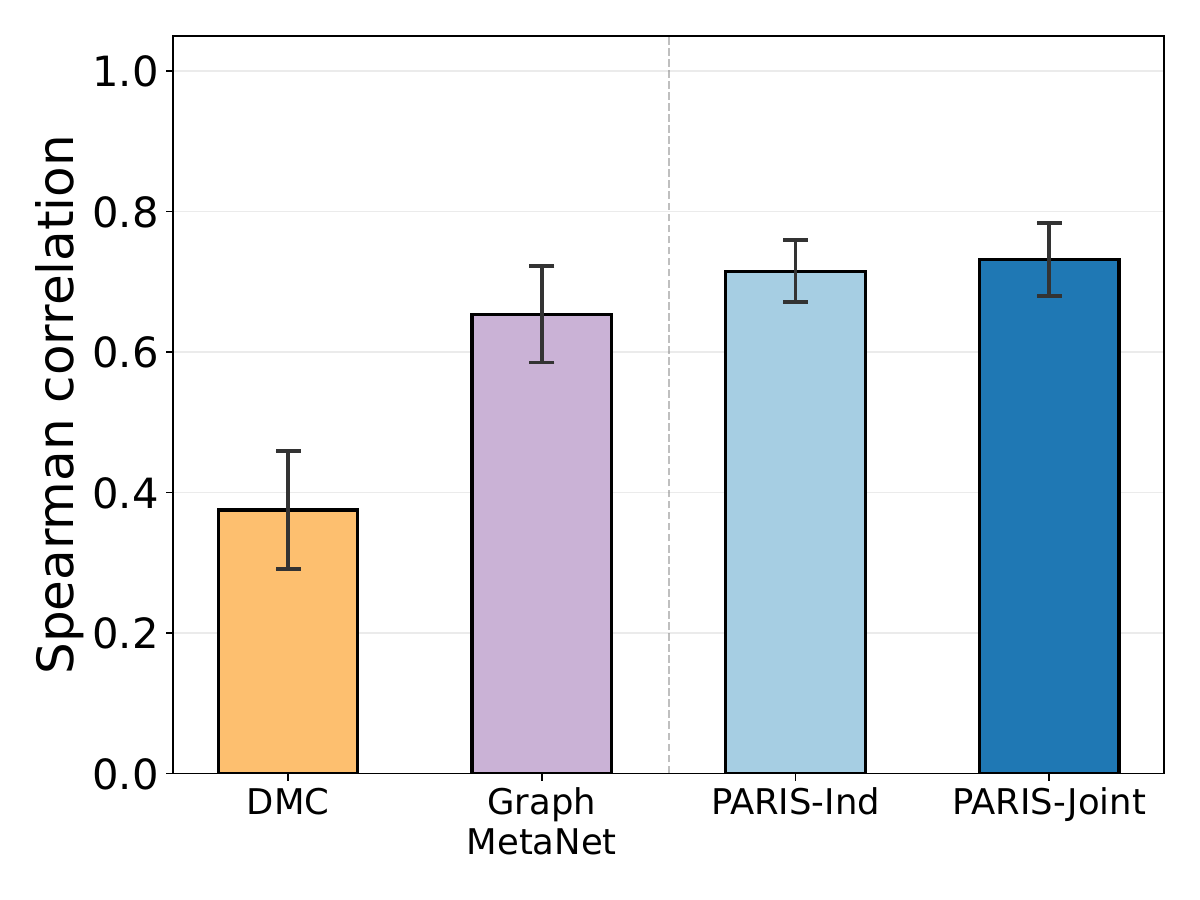}
\caption{ 
Out-of-distribution generalization across architecture configurations.
}
\label{fig:ood}
\end{figure*}

We further investigate out-of-distribution (OOD) architectural generalization by training metanetworks on a reduced set of architectures with smaller hidden dimensions and evaluating them on held-out architectures with larger hidden dimensions. Fig.~\ref{fig:ood} demonstrates that both $\mathsf{PARIS}$ variants outperform the competing metanetwork baselines, suggesting that their learned computation representations transfer to the evaluated architectural configurations.
These experiments demonstrate that $\mathsf{PARIS}$ is not limited to training within a single architecture family or to the exact hidden-dimension configurations observed during meta-training.

\subsection{Cross-task prediction with partially overlapping label spaces}

\begin{figure*}[h]
    \centering
     \includegraphics[width=0.4\textwidth]{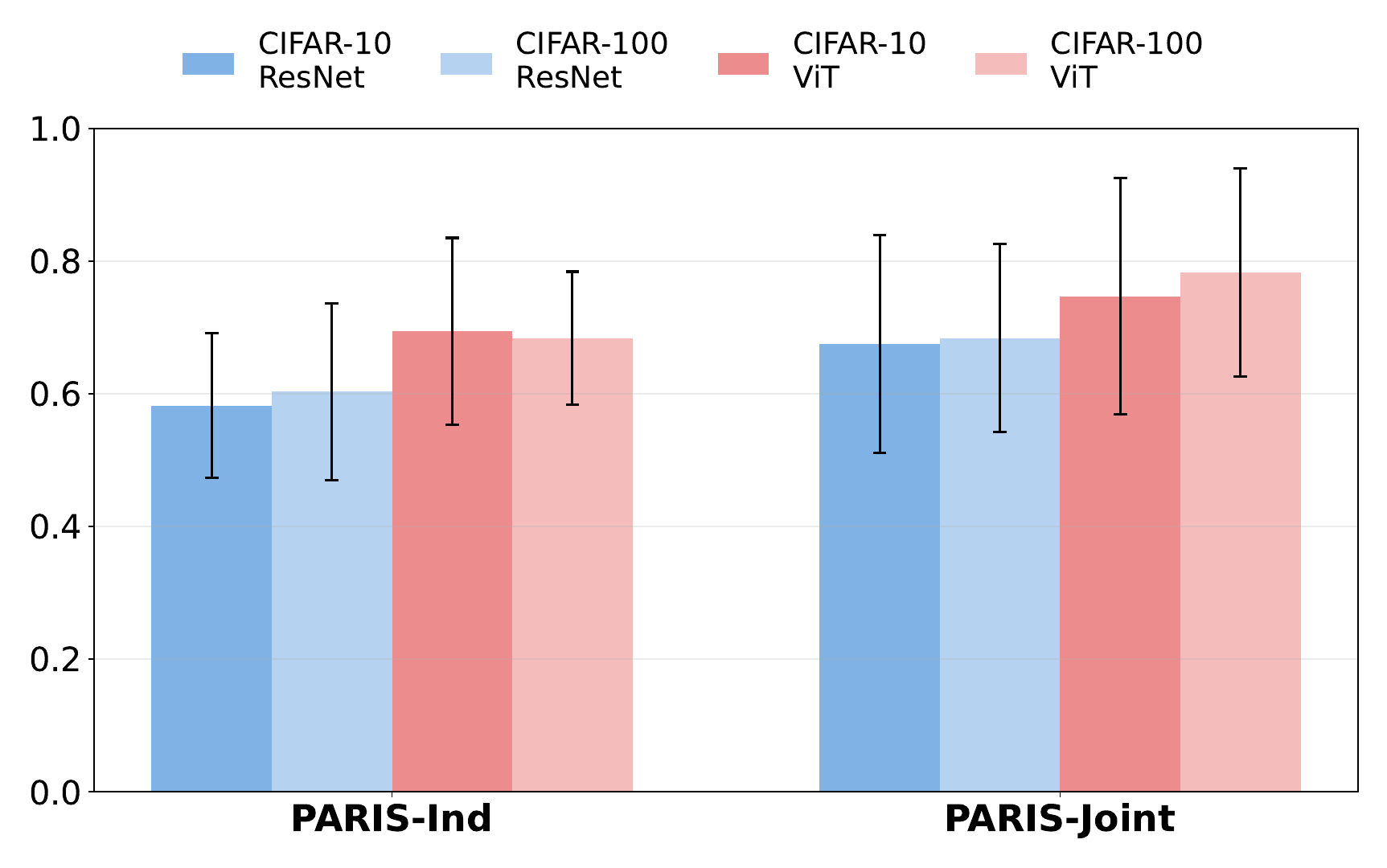}
\caption{ 
Cross-task correlation under partially overlapping label spaces.
We partition each parent dataset's classes into $30\%$ task-A-only, $40\%$ shared, and $30\%$ task-B-only subsets. Each task contains $70\%$ of the original classes.
For pairs containing one site from each task-specific base-model population, entries report aggregate correlation between predicted and activation-derived reference dissimilarities.
}
\label{fig:partial_overlap_tasks}
\end{figure*}

We examine representation-relation prediction between base models trained on related but nonidentical classification tasks. For each parent dataset, let $\mathcal R$ be its set of $K$ class labels. We partition $\mathcal R$ into three disjoint subsets: task-A-only classes $\mathcal R_A$, shared classes $\mathcal R_{\cap}$, and task-B-only classes $\mathcal R_B$, with $
    \bigl(
        |\mathcal R_A|,
        |\mathcal R_{\cap}|,
        |\mathcal R_B|
    \bigr)
    = (0.3K,0.4K,0.3K).$
The two task label sets are
$
    \mathcal Y_A=\mathcal R_A\cup\mathcal R_{\cap}$ and 
    $\mathcal Y_B=\mathcal R_B\cup\mathcal R_{\cap}$.
Each task contains $70\%$ of the parent dataset's classes, and the intersection contains $40\%$ of the parent class universe. For CIFAR-10, this gives two seven-class tasks with four shared classes.  For CIFAR-100, it gives two 70-class tasks with 40 shared classes. The class partition defines the source tasks.
Base-model training data are restricted to the respective task label sets. We evaluate site pairs with one endpoint from an A-trained model and the other from a B-trained model, separately for ResNets and ViTs within each parent dataset.
The metanetwork receives the corresponding site graphs and trained state. For each relation, both sites are evaluated on the same ordered reference inputs to construct the activation-derived target. Fig.~\ref{fig:partial_overlap_tasks} reports the correlation
between predicted and reference dissimilarities across the evaluated site pairs.
We see that the advantage of the joint pipeline in predicted-reference correlation persists when the two base models are trained on different, partially overlapping class sets.

\subsection{Comparison with Filter-Subspace Similarity and Tensor Similarity}

In Sec.~\ref{sec:exp}, we compare our method with metanetwork baselines, including graph metanetworks,  and in App.~\ref{app:extend_related_work},
we discuss two activation-free similarity approaches that are not metanetwork-based: \textit{(i)} filter-subspace similarity \citep{chen2023inner} and \textit{(ii)} tensor similarity \citep{gonzalez2026two}.
Both are relevant to our goal of comparing representations without collecting new activations.
Yet, we remark that the setup in these activation-free similarity approaches is different than ours. 
Specifically: 
\begin{itemize} [leftmargin=1.75em,itemsep=2pt]
    \item
    \cite{chen2023inner} define the filter subspace distance via a layerwise cosine similarity between model-specific filter atoms under a decomposition whose coefficients are shared across the compared models. 
    In contrast, we follow prior metanetwork work and use standard ResNet and ViT checkpoints trained without this constraint. 
    Thus, applying the published score to these checkpoints requires an explicitly specified common decomposition, or a separate population constructed 
    under the required parameterization.
    \item \cite{gonzalez2026two} introduced tensor similarity that supports polynomial or multilinear computations.
    Their construction does not directly cover the complete upstream computations in our benchmark, which contain non-polynomial operations such as ReLU and softmax attention. 
    Their Gaussian functional similarity also differs from the representation similarities that our method learns to predict. 
    We could approximate these operations with polynomials or compare only linear parts of the network, but doing so would change the computations being compared.
\end{itemize}

To compare our method with filter-subspace similarity and tensor similarity, we could consider the following complementary settings.
\begin{itemize} [leftmargin=1.75em,itemsep=2pt]
    \item For filter subspace distance, we consider two settings:
    \begin{enumerate}[leftmargin=1.75em,itemsep=2pt]
        \renewcommand{\labelenumi}{\textit{(\roman{enumi})}}
    \item We can relax the subspace constraints in the filter subspace similarity for compatible tensor shapes, which gives us a common-coefficient representation. In this case, the atom-cosine score would simply be kernel cosine with no reconstruction error. Under this setup, we use 50 ResNet models for the CIFAR-10 dataset. 
    The full-rank identity-coefficient version achieves correlations of $0.27/0.28/0.24/0.17/0.19$ with linear CKA, CKA, GULP, MutualNN, and readout disagreement, respectively. In comparison, PARIS-Ind achieves $0.64/0.54/0.63/0.79/0.52$, while PARIS-Joint achieves $0.61/0.75/0.67/0.93/0.80$. 
    \item We can also construct a small model population whose atoms are trained under fixed shared coefficients, so that our method also receives the exact effective kernels and inference state of those same models. For the preliminary test, we consider an initial model collection with 60 ResNet models trained on CIFAR-10.
    On the shared-coefficient population, the filter-subspace atom-cosine score achieves correlations of $0.93/0.95/0.89/0.74/0.81$ with linear CKA, CKA, GULP, MutualNN, and readout disagreement, respectively. In comparison, PARIS-Ind achieves $0.96/0.99/0.90/0.78/0.85$, while PARIS-Joint achieves $0.95/0.99/0.92/0.76/0.88$.
    \end{enumerate}
    \item For tensor similarity, we consider a model population of 300 independently trained three-layer linear models on varied regression problems so the population contains meaningful functional diversity. Tensor similarity achieves correlations of $0.94/0.64/0.67/0.72/0.52$ with linear CKA, CKA, MutualNN, GULP, and readout disagreement, respectively. In comparison, PARIS-Ind achieves $0.95/0.77/0.82/0.84/0.63$, and PARIS-Joint achieves $0.97/0.80/0.84/0.89/0.67$.
\end{itemize}
We can see that under these complementary settings, both PARIS variants achieve higher correlations across all five targets.

\subsection{Sensitivity to positive-rescaling}
\label{app:Sensitivity to positive-rescaling}

\begin{figure*}[h]
    \centering
     \includegraphics[width=0.6\textwidth]{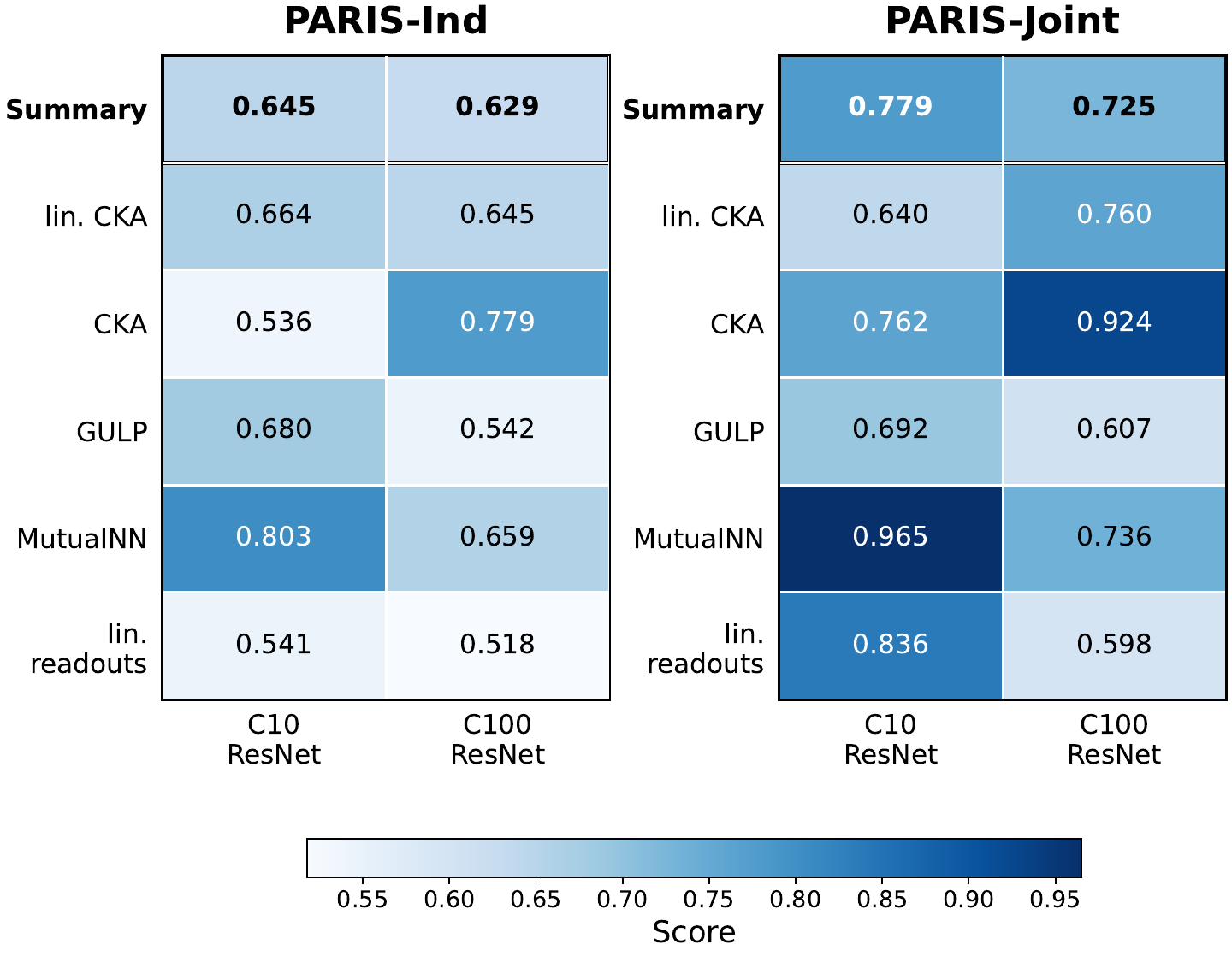}
\caption{ 
Score-prediction performance under compensated positive ReLU rescaling.
}
\label{fig:positive_scaling}
\end{figure*}

To investigate the limitations of our symmetry guarantees, we evaluate the sensitivity of $\mathsf{PARIS}$ to positive rescaling using ResNet models on CIFAR-10 and CIFAR-100. Specifically, we consider a positive rescaling within residual branches, where we multiply the affine scale and bias of a BatchNorm layer preceding ReLU by positive channel-wise factors and divide the following convolution's corresponding input-channel weights by the same factors. By the positive homogeneity of ReLU, the rescaling and its compensation cancel before the selected complete-block output. We repeat the score-prediction evaluation under positive scaling and compare the resulting correlations with the original results. Fig.~\ref{fig:positive_scaling} reports these comparisons for both $\mathsf{PARIS}$ variants and all five representation relations. We see that the prediction correlations remain relatively stable. These experiments illustrate that PARIS can retain performance under some parameter rescalings, but its predictions are not guaranteed to be invariant to transformations outside the supported symmetry families.

\subsection{Evaluating the influence of relative depth}

To investigate the role of relative depth in representation similarity prediction, we conduct two complementary experiments. Although $\mathsf{PARIS}$ does not receive relative depth explicitly, the structure of its upstream computation graphs allows it to infer information about the number and organization of preceding operations. So, in this subsection, we examine whether normalized site positions can explain its prediction performance, and whether $\mathsf{PARIS}$ captures additional information when relative depth is controlled.

\begin{figure*}[h]
    \centering
    \captionsetup{skip=3pt}

    \begin{minipage}[t]{0.49\textwidth}
        \centering
        \includegraphics[width=\linewidth]{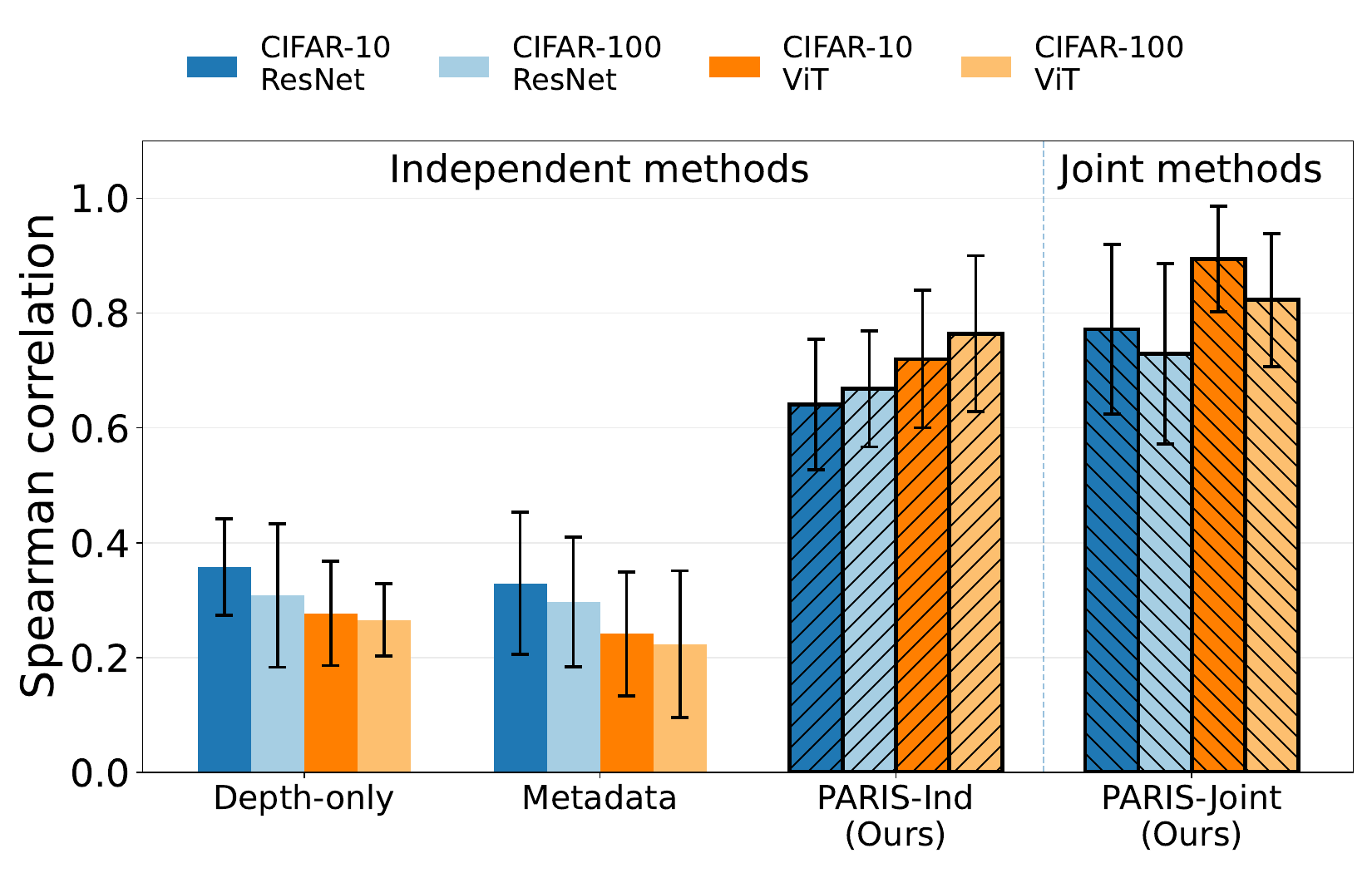}
        \caption{Spearman correlations  between predicted and ground-truth representation relations for depth-only and metadata-based predictors, compared with $\mathsf{PARIS}$-Ind and $\mathsf{PARIS}$-Joint}.
        \label{fig:depth_only_original_dataset}
    \end{minipage}
    \hfill
    \begin{minipage}[t]{0.49\textwidth}
        \centering
        \includegraphics[width=\linewidth]{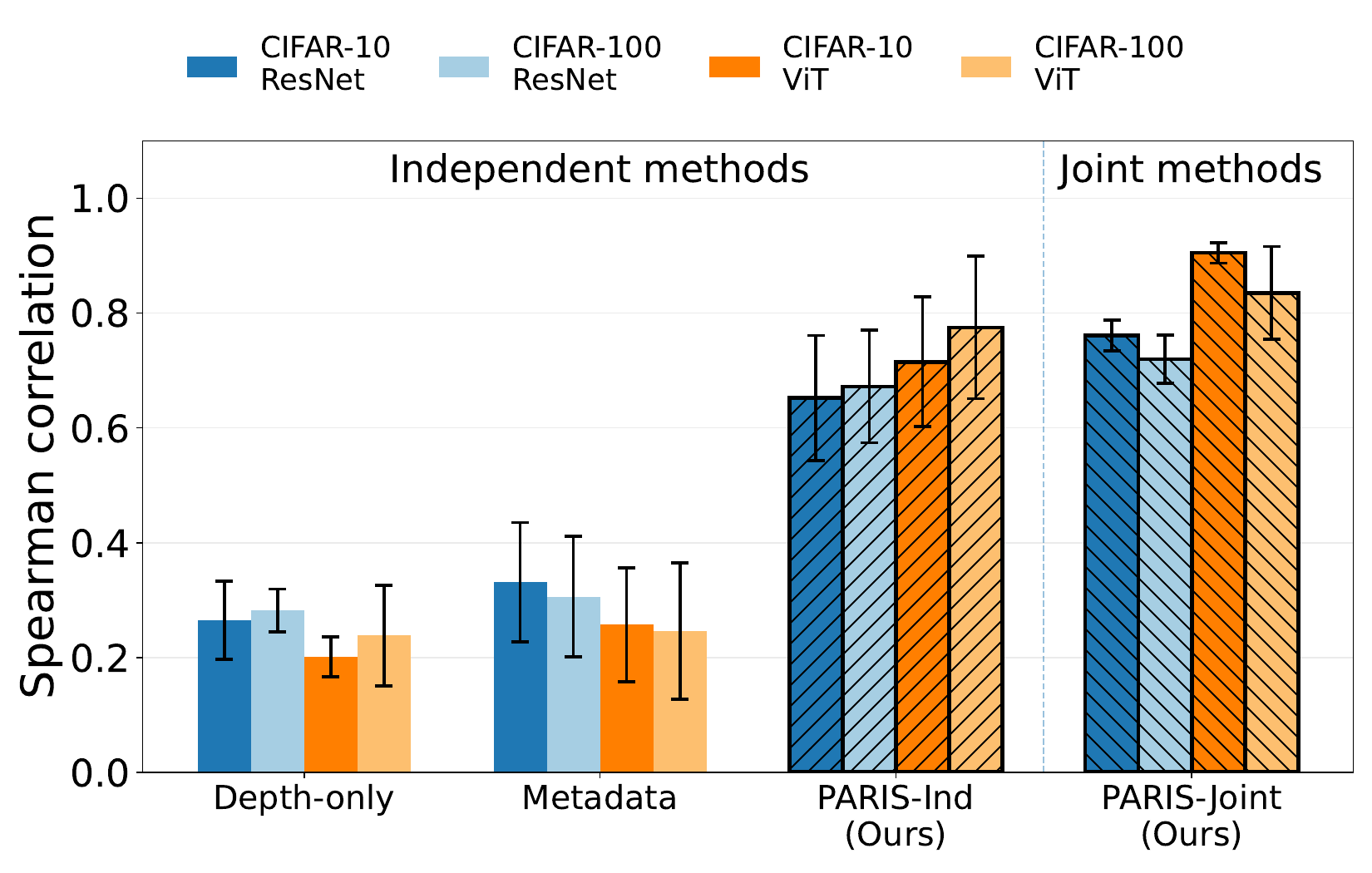}
        \caption{Spearman correlations on a depth-controlled evaluation in which compared sites occupy matched normalized positions across models. }
        \label{fig:depth_only_controlled_dataset}
    \end{minipage}

\end{figure*}

We first compare our methods against depth-only and metadata-based predictors on the original ResNet and ViT populations trained on CIFAR-10 and CIFAR-100. Fig.~\ref{fig:depth_only_original_dataset} shows the resulting Spearman correlations. We see that relative depth and model metadata provide limited predictive information compared with the upstream computation graphs used by $\mathsf{PARIS}$.
To further examine whether $\mathsf{PARIS}$ relies primarily on relative-depth alignment, we construct a controlled evaluation in which the compared sites occupy matched normalized positions across models. This reduces the influence of differences in relative site position and allows us to assess whether the methods capture additional variation in representation similarity.
Fig.~\ref{fig:depth_only_controlled_dataset} demonstrates the results in this setting. We observe that the advantage of $\mathsf{PARIS}$ persists across both ResNet and ViT populations on CIFAR-10 and CIFAR-100.
These results indicate that relative depth provides some information about representation correspondence, but does not adequately explain the variation in similarity scores across the evaluated model populations. The stronger performance of $\mathsf{PARIS}$, including in the depth-controlled setting, supports the benefit of using the upstream computations that produce each representation and they show that the information available to $\mathsf{PARIS}$ goes beyond normalized site positions and the tested metadata features.

\subsection{Representation similarity prediction across modalities}

To investigate whether $\mathsf{PARIS}$ can predict representation relations beyond unimodal vision models, we extend our evaluation to pretrained CLIP checkpoints from \cite{nezhurina2025scaling}. These models consist of image and text encoders jointly trained on paired image-text data, providing a setting for studying representation similarity within and across modalities.
We consider three categories of representation pairs: \textit{(i)} image-image, comparing representations from image encoders,  \textit{(ii)} text-text, comparing representations from text encoders, and \textit{(iii)} image-text, comparing representations across the two modalities. For each category, we evaluate the prediction of activation-derived reference dissimilarities using the trained computations of the corresponding representation sites. We compare $\mathsf{PARIS}$-Ind and $\mathsf{PARIS}$-Joint against an adapted Graph Metanetwork  baseline.

\begin{figure*}[h]
    \centering
     \includegraphics[width=0.55\textwidth]{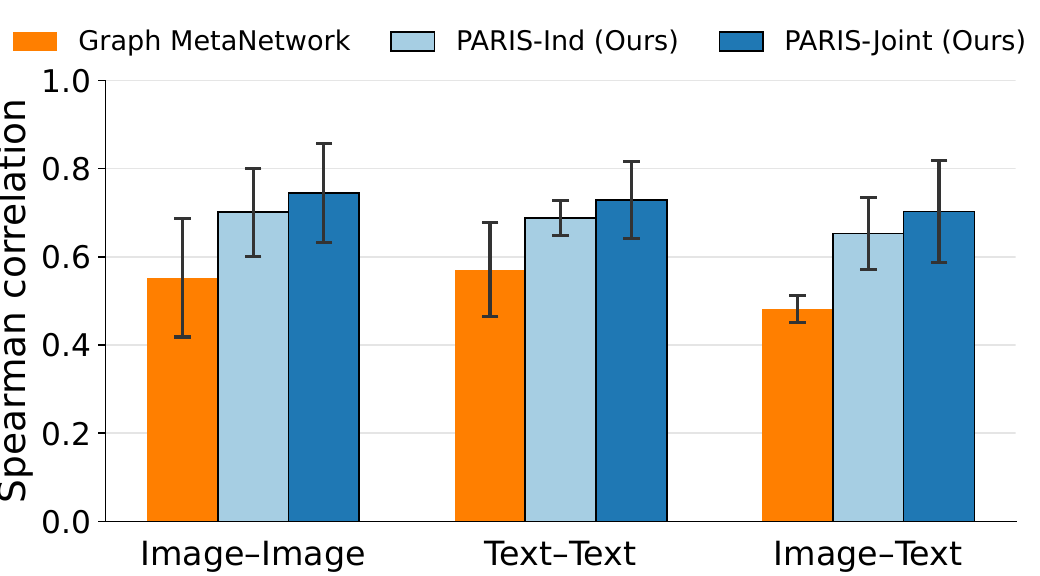}
\caption{ 
Representation similarity prediction across modalities using pretrained CLIP models.
}
\label{fig:multi_modal_pre}
\end{figure*}

Fig.~\ref{fig:multi_modal_pre} presents the Spearman correlations for the three modality-pair categories. For image-image prediction.
We see that both $\mathsf{PARIS}$ variants achieve higher mean correlations than Graph Metanetwork. While image-text prediction is more challenging than the within-modality settings in this evaluation, PARIS maintains meaningful predictive performance. These findings provide initial evidence that computation-based metanetworks can predict representation relationships involving different modalities.
Note that this experiment considers image and text encoders from jointly pretrained CLIP-style models. As such, it establishes an initial extension to multimodal pretrained models, but does not establish generalization to arbitrary independently trained unimodal networks. We plan to investigate in future work about cross-modal representational convergence across unrelated vision and language models, as well as other architectures and training objectives.

\subsection{Stitching-candidate selection across sizes and across diverse architectures}

To investigate the size of stitching-candidate selection, we extend the evaluation in Sec.~\ref{sec:Stitching candidate selection} from a candidate pool of 20 sites to pools of 50, 100, and 200 sites, while keeping the number of selected candidates fixed at five. We evaluate both $\mathsf{PARIS}$-Ind and $\mathsf{PARIS}$-Joint on the four original model/dataset settings: ResNet and ViT populations trained on CIFAR-10 and CIFAR-100.
We additionally compare two meta-training settings (as in App.~\ref{app:Meta-training across diverse architectures}). In Same family, the metanetwork is trained on models from the same architecture family as the evaluation population. In Diverse, a metanetwork is trained on both ResNet and ViT populations. The stitching evaluation continues to consider same-family source and destination models, allowing us to isolate the effect of architecture diversity during meta-training.

\begin{figure*}[h]
    \centering
     \includegraphics[width=0.95\textwidth]{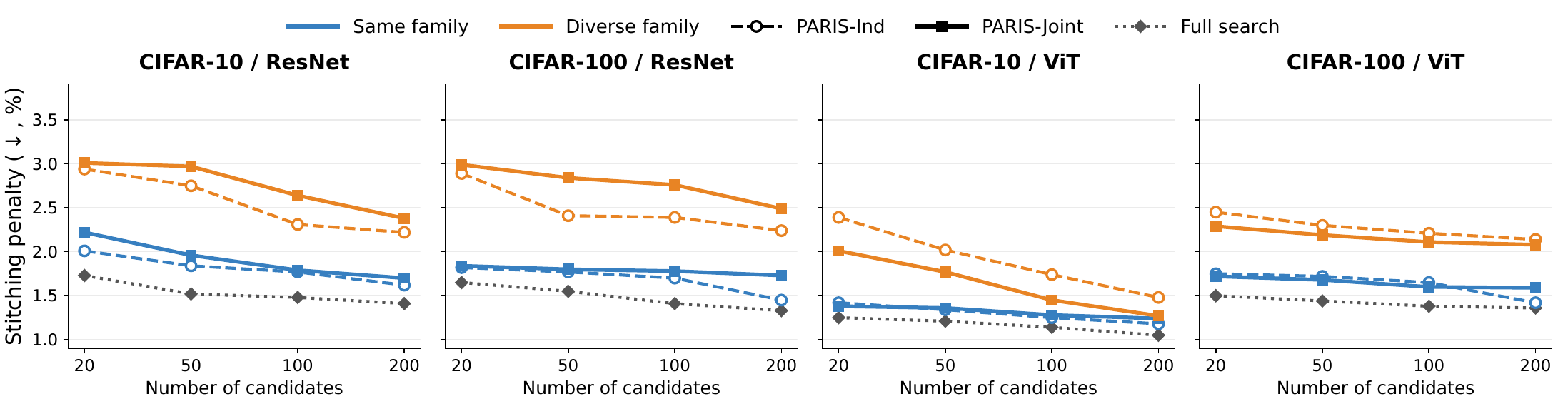}
\caption{ 
Scalability of stitching-candidate selection under same-family and diverse-family meta-training.}
\label{fig:stitching_scale}
\end{figure*}

Fig.~\ref{fig:stitching_scale} shows the performance of the stitching candidate selection. Across all four settings, increasing the number of candidates generally improves the performance of both $\mathsf{PARIS}$ variants under both meta-training regimes. Under same-family meta-training, the stitching penalties remain relatively close to full search, particularly for the ViT populations. These results indicate that our method can benefit from larger candidate pools.

\subsection{Transfer to pretrained vision encoders on Flowers-102}

\begin{wrapfigure}[16]{r}{0.42\textwidth}  
\vspace{-4.5 mm}
           \captionsetup{skip=3pt}
     \includegraphics[width=0.42\textwidth]{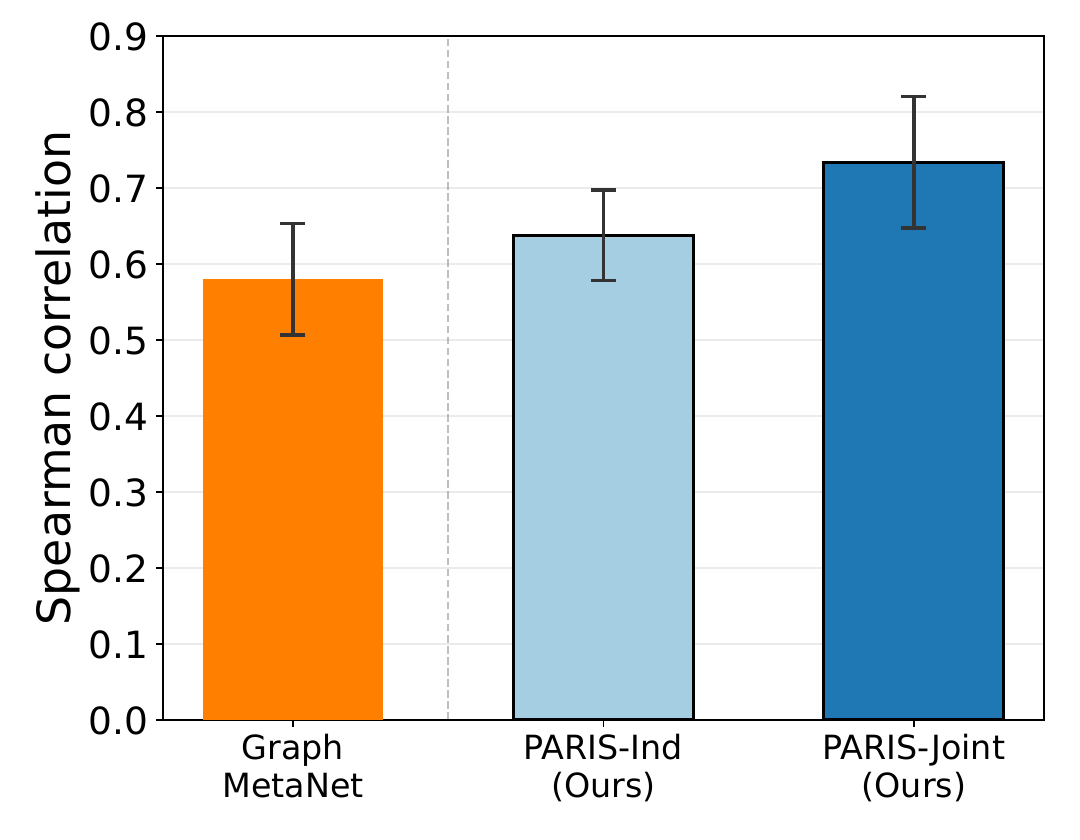}
    \caption{
Representation-score prediction on pretrained CLIP vision encoders evaluated on Flowers-102.
    }
    \label{fig:clip_flower}
\end{wrapfigure}
To evaluate the applicability of $\mathsf{PARIS}$ beyond the CIFAR-trained model populations, we consider a collection of  CLIP ViT-B/32 vision-encoder checkpoints from \cite{wortsman2022model}. These checkpoints originate from a common pretrained CLIP initialization and correspond to distinct fine-tuning runs on ImageNet with different hyperparameter configurations. We use the vision encoders from these checkpoints and this setting allows us to investigate transfer to an existing population of pretrained and subsequently fine-tuned models without training a new base-model population.

We evaluate these vision encoders on Flowers-102 using linear probing. We partition the 72 checkpoints into 48 for adaptation, 12 for validation, and 12 for held-out testing. We investigate whether the learned graph representations of $\mathsf{PARIS}$ can support representation-score prediction in this new setting, without retraining the underlying vision encoders. We compare $\mathsf{PARIS}$-Ind and $\mathsf{PARIS}$-Joint against an adapted Graph Metanetwork baseline, using the same evaluation protocol.

Fig.~\ref{fig:clip_flower} reports the Spearman correlations, where we see that $\mathsf{PARIS}$'s learned computation representations can provide useful information for predicting representation relations in a pretrained-model population and downstream dataset beyond those considered in our original benchmarks.
This evaluation extends our method to a population of pretrained vision encoders and a new dataset as evidence of transfer with adaptation.

\end{document}